%% file: main.tex
\documentclass{article}

\usepackage[preprint]{neurips_data_2023}

\usepackage{adjustbox}

\usepackage[utf8]{inputenc} % allow utf-8 input
\usepackage[T1]{fontenc}    % use 8-bit T1 fonts
\usepackage{url}            % simple URL typesetting
\usepackage{booktabs}       % professional-quality tables
\usepackage{amsfonts}       % blackboard math symbols
\usepackage{nicefrac}       % compact symbols for 1/2, etc.
\usepackage{microtype}      % microtypography
\usepackage{hyperref}
\usepackage[table,xcdraw,usenames,dvipsnames]{xcolor}
\hypersetup{
    colorlinks=true,
    linkcolor=blue,    
    urlcolor=purple,
    citecolor=green,
}

\usepackage{times}
\usepackage{epsfig}
\usepackage{graphicx}
\usepackage{amsmath}
\usepackage{amssymb}

\usepackage{multicol}
\usepackage{microtype}
\usepackage{graphicx}
\usepackage{subfigure}
\usepackage{booktabs} %
\usepackage{multirow}
\usepackage{makecell}
\usepackage{capt-of}
\usepackage{bm,bbm}
\usepackage[makeroom]{cancel}
\usepackage[table,xcdraw,dvipsnames]{xcolor}
\usepackage{wrapfig}

\usepackage{pythonhighlight}

\newcommand{\superclass}[0]{Synonym Consistency score}
\newcommand{\superclassSymbl}[0]{$\gamma_\text{syn}$}
\newcommand{\interSim}[0]{Class Dispersion score}
\newcommand{\interSimSymbl}[0]{$\rho_{\text{disp}}$}
\newcommand{\intraSim}[0]{Silhouette score}
\newcommand{\intraSimSymbl}[0]{$\varphi_\text{sil}$}

\newcommand{\intraClassClose}[0]{Fisher criterion}
\newcommand{\intraClassCloseSymbl}[0]{$\phi_\text{fisher}$}

\newcommand{\numdatasets}[0]{23}
\newcommand{\nummodels}[0]{35}

\definecolor{myyellow}{HTML}{FFFFCC}
\definecolor{myblack}{HTML}{000000}

\newcommand{\myboxa}[1]{%
    \setlength{\fboxsep}{2pt}%
    \colorbox{myyellow}{\raisebox{0ex}{\strut\textcolor{myblack}{\rule{3pt}{3pt}}}}\hspace{0.0em}#1%
}

\definecolor{lightgray}{RGB}{240,240,240} % Adjust the RGB values for a lighter gray
\definecolor{orange}{HTML}{FFA500}

\newcommand{\myboxb}[1]{%
    \setlength{\fboxsep}{5pt}%
    \colorbox{lightgray}{\makebox[1pt][c]{\textcolor{orange}{$\times$}}}\hspace{0.0em}#1%
}

\usepackage[toc,page,header]{appendix}
\usepackage{minitoc}

\title{LOVM: Language-Only Vision Model Selection}

\author{Orr Zohar \\ Stanford Univeristy \\ \texttt{orrzohar@stanford.edu} 
\And
Shih-Cheng Huang  \\ Stanford Univeristy \\ \texttt{mschuanmschuang@stanford.edu}
\And
Kuan-Chieh Wang \\ Stanford Univeristy \\ \texttt{wangkua1@stanford.edu} 
\And
Serena Yeung  \\ Stanford Univeristy \\ \texttt{syyeung@stanford.edu} 
}

\begin{document}
\doparttoc 
\faketableofcontents 

\maketitle

\begin{abstract}
Pre-trained multi-modal vision-language models (VLMs) are becoming increasingly popular due to their exceptional performance on downstream vision applications, particularly in the few- and zero-shot settings. 
However, selecting the best-performing VLM for some downstream applications is non-trivial, as it is dataset and task-dependent. Meanwhile, the exhaustive evaluation of all available VLMs on a novel application is not only time and  computationally demanding but also necessitates the collection of a labeled dataset for evaluation. 
As the number of open-source VLM variants increases, there is a need for an efficient model selection strategy that does not require access to a curated evaluation dataset. 
This paper proposes a novel task and benchmark for efficiently evaluating VLMs' zero-shot performance on downstream applications without access to the downstream task dataset. 
Specifically, we introduce a new task LOVM: \underline{L}anguage- \underline{O}nly  \underline{V}ision  \underline{M}odel Selection, where methods are expected to perform both model selection and performance prediction based solely on a text description of the desired downstream application.
We then introduced an extensive LOVM benchmark consisting of ground-truth evaluations of \nummodels\ pre-trained VLMs and \numdatasets\ datasets, where methods are expected to rank the pre-trained VLMs and predict their zero-shot performance. 

\end{abstract}

\input{sections/introduction}

\input{sections/task}

\input{sections/baselines}

\input{sections/results}

\input{sections/related_works}

\input{sections/conclusions}

\newpage
\bibliographystyle{plainnat}
\bibliography{main}
\newpage

\appendix

\begin{center}
    \textbf{\Large LOVM: Language-Only Vision Model Selection
    \\
    Appendix}
\end{center}
\vspace{0.0in}

\input{sections/appendix}

\newpage
\section*{Checklist}

\begin{enumerate}

\item For all authors...
\begin{enumerate}
  \item Do the main claims made in the abstract and introduction accurately reflect the paper's contributions and scope?
    \answerYes{}
  \item Did you describe the limitations of your work?
    \answerYes{See Sec.~\ref{sup:sec:lim}}
  \item Did you discuss any potential negative societal impacts of your work?
    \answerYes{See Sec.~\ref{sup:sec:neg_impact}}
  \item Have you read the ethics review guidelines and ensured that your paper conforms to them?
    \answerYes{}
\end{enumerate}

\item If you are including theoretical results...
\begin{enumerate}
  \item Did you state the full set of assumptions of all theoretical results?
    \answerNA{}{}
	\item Did you include complete proofs of all theoretical results?
    \answerNA{}
\end{enumerate}

\item If you ran experiments (e.g. for benchmarks)...
\begin{enumerate}
  \item Did you include the code, data, and instructions needed to reproduce the main experimental results (either in the supplemental material or as a URL)?
    \answerYes{See Supplemental Material for code repository to reproduce all of our results, has the dataset, and even has an evaulation script for future works.}
  \item Did you specify all the training details (e.g., data splits, hyperparameters, how they were chosen)?
    \answerYes{See Sec.~\ref{sec:meth} and Sec.~\ref{sup:sec:method_details}}
	\item Did you report error bars (e.g., with respect to the random seed after running experiments multiple times)?
    \answerNA{}
	\item Did you include the total amount of compute and the type of resources used (e.g., type of GPUs, internal cluster, or cloud provider)?
    \answerYes{see Sec.~\ref{sup:sec:benchmark_details}}
\end{enumerate}

\item If you are using existing assets (e.g., code, data, models) or curating/releasing new assets...
\begin{enumerate}
  \item If your work uses existing assets, did you cite the creators?
    \answerYes{open-clip, and yes, see Sec.~\ref{sec:task:eval} and throughout the manuscript}
  \item Did you mention the license of the assets?
    \answerNA{it is open-source and free to use}
  \item Did you include any new assets either in the supplemental material or as a URL?
    \answerYes{See codebase in supplemental, will be made open-source}
  \item Did you discuss whether and how consent was obtained from people whose data you're using/curating?
    \answerNA{}
  \item Did you discuss whether the data you are using/curating contains personally identifiable information or offensive content?
    \answerNA{}
\end{enumerate}

\item If you used crowdsourcing or conducted research with human subjects...
\begin{enumerate}
  \item Did you include the full text of instructions given to participants and screenshots, if applicable?
    \answerNA{}
  \item Did you describe any potential participant risks, with links to Institutional Review Board (IRB) approvals, if applicable?
    \answerNA{}
  \item Did you include the estimated hourly wage paid to participants and the total amount spent on participant compensation?
    \answerNA{}
\end{enumerate}

\end{enumerate}

\end{document}

%% file: sections/introduction.tex
\section{Introduction}

Advancements in artificial intelligence (AI) have permeated diverse sectors, but applications in areas such as medicine or those with long-tail distributions often struggle to collect the sizable training datasets required for the standard supervised learning framework.
Pre-trained vision-language models (VLMs) offer a promising solution, demonstrating robust performance on diverse downstream vision tasks without the necessity of large-scale labeled datasets~\citep{CLIP, ALIGN}. 
However, the performance of VLMs can vary substantially across different tasks and domains, which undermines the reliance solely on benchmark dataset performance for effective VLM selection. Consequently, users aiming to \emph{select a VLM} for custom downstream applications frequently face a predicament: the lack of established performance rankings for these specific, non-conventional tasks.

\begin{wrapfigure}[12]{t}{0.4\textwidth}\vspace{-0.09in}
\begin{center}
\includegraphics[width=0.4\textwidth]{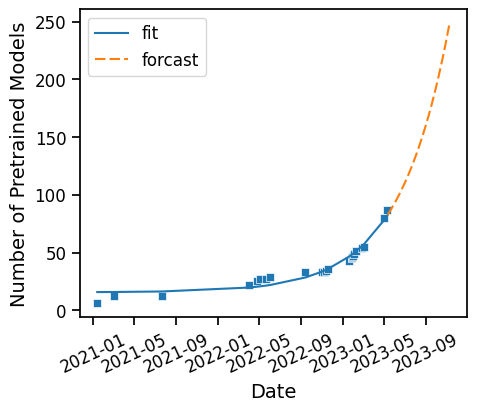}
\end{center}
\vspace{-0.18in}
\caption{\textbf{LOVM Motivation.} Number of pre-trained VLMs released on open-clip over time.\label{fig:motivation}}
\end{wrapfigure}

As the number of pre-trained VLMs increases (see Fig.~\ref{fig:motivation}, \citep{open_clip} and App. Tab.~\ref{tab:modeldataset}),  the challenge of model selection escalates. Exhaustive evaluation of all available VLMs on a novel application is not only time and computationally demanding but also necessitates the collection of a labeled dataset for evaluation. 
However, many users lack the resources or technical proficiency to collect  
and label an evaluation dataset and subsequently evaluate all available VLMs. 
Consequently, the development of methods that efficiently select the most suitable model for a given task without relying on access to the downstream task dataset has become critically important. 

Recent studies have demonstrated that text embeddings from VLMs can be used as a proxy for their corresponding image embeddings in various downstream tasks, including classification and error slice discovery~\citep{drML, DOMINO, MMFailure}. Specifically, although \citet{ModalityGap} has shown that there exists a modality gap between text and image embeddings generated from VLMs, the geometry of this modality gap permits cross-modality transferability. This phenomenon allows text to serve as a proxy to corresponding images and vice versa.
Therefore we aim to explore the utilization of cross-modality transferability to estimate VLM performance on a novel vision task using text alone.

\begin{wrapfigure}[20]{t}{0.4\textwidth}\vspace{0.2in}
\begin{center}
\includegraphics[width=0.4\textwidth]{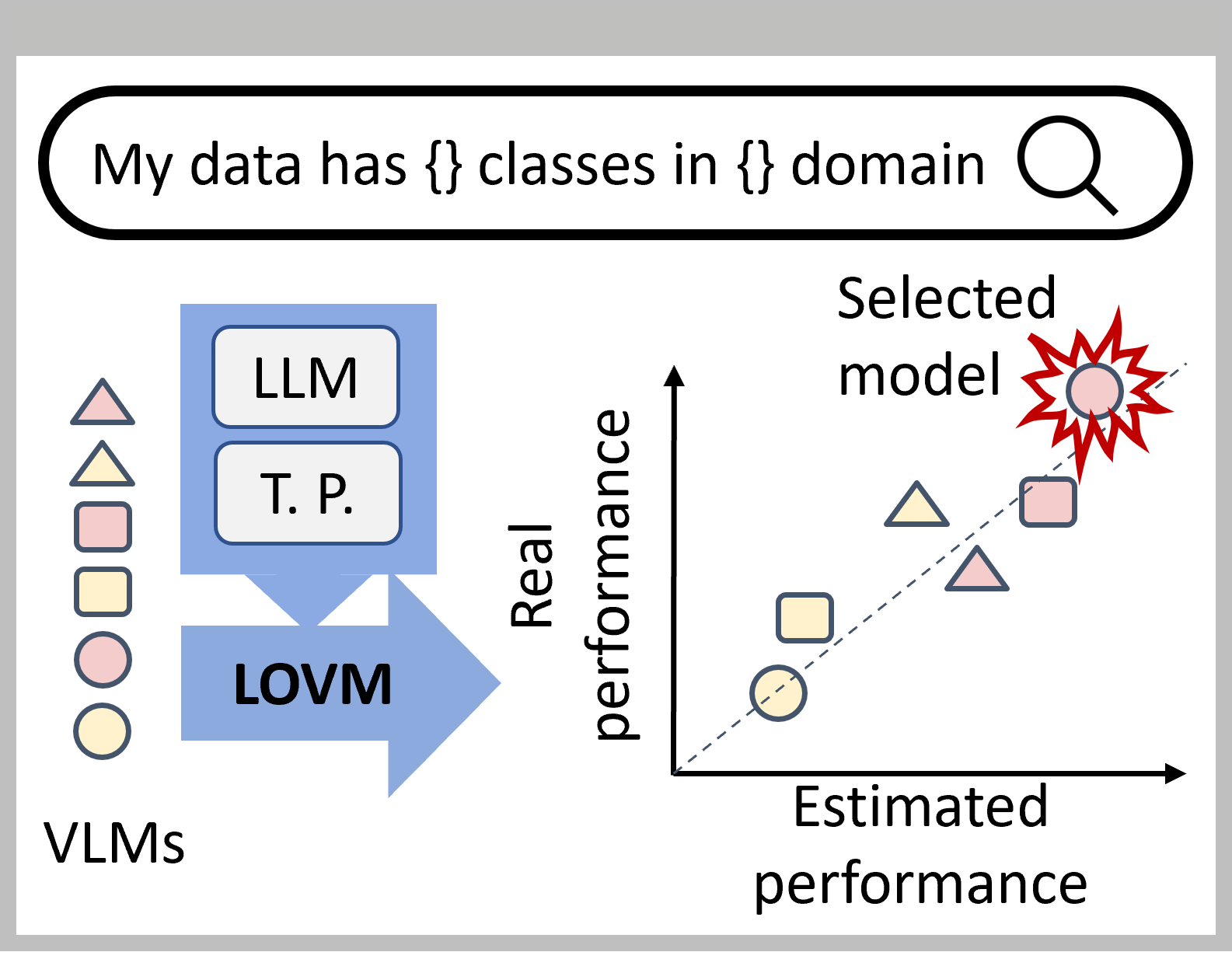}
\end{center}
\vspace{-0.15in}
\caption{\textbf{An overview of an application for LOVM methods.} A user can type into a search bar the details of the desired task, and then LOVM methods evaluate and rank the available models.
\label{fig:application1}}
\end{wrapfigure}

Herein, we propose a novel problem setting - Language-Only VLM selection (LOVM) as a novel model selection task. 
In the LOVM task, methods are expected to select the optimal VLM and predict its expected performance given only a text description of a 
downstream vision task/application,  (see Fig.~\ref{fig:application1}).
Importantly, LOVM eliminate the need to gather, organize, and annotate custom datasets, thereby greatly simplifying the model selection process for downstream users. 
To facilitate the development of LOVM methods in the future, we collected a large dataset of ground-truth evaluations of \nummodels\ pre-trained VLMs on \numdatasets\ datasets. We then introduce the appropriate evaluation protocol and method quality metrics to allow the evaluation and comparison of LOVM methods in the future. 

To show that such a challenging task is possible, we provide simple baselines that utilize readily-available large language models to generate `text datasets' for a given vision task. By utilizing the cross-modality transferability phenomenon, we show how simple baselines can be derived by utilizing the cross-modality transferability phenomenon. 
Our results show that text prompting may be an effective means of estimating zero-shot performance, showing that such a challenging task is possible while providing a baseline for future research. 

The contributions of this study can be summarized as follows:
\begin{itemize}
    \item We propose a novel problem setting, \textbf{LOVM}: Language-Only VLM selection and performance prediction. LOVM methods are expected to perform both model selection and performance prediction using only a text description of the desired zero-shot application. 
    
    \item We provide a benchmark consisting of \nummodels\ pre-trained VLMs and \numdatasets\ datasets. We evaluated all dataset-VLM combinations and reported their corresponding performance, which is used as the ground truth when training and evaluating LOVM methods. We also introduce the corresponding evaluation metrics and protocols. 
    
    \item In developing the LOVM baselines, we introduce several novel methodological contributions, such as using LLM models to generate text proxies for images. Our text-based methods outperform simple baselines - such as ImageNet benchmarking, showing the promise of the direction of LOVM.
    
    \item By analyzing text-based score trends, we draw insights into VLM behavior and shed light on why ResNet-based models perform better on datasets with low visual diversity.
\end{itemize}

Our code and dataset are available at \url{https://github.com/orrzohar/LOVM}

%% file: sections/task.tex
 \begin{figure*}

    \centering
  %%%PREPRINT%%%\vspace{-0.1in}

  \resizebox{1\textwidth}{!}{
  \includegraphics{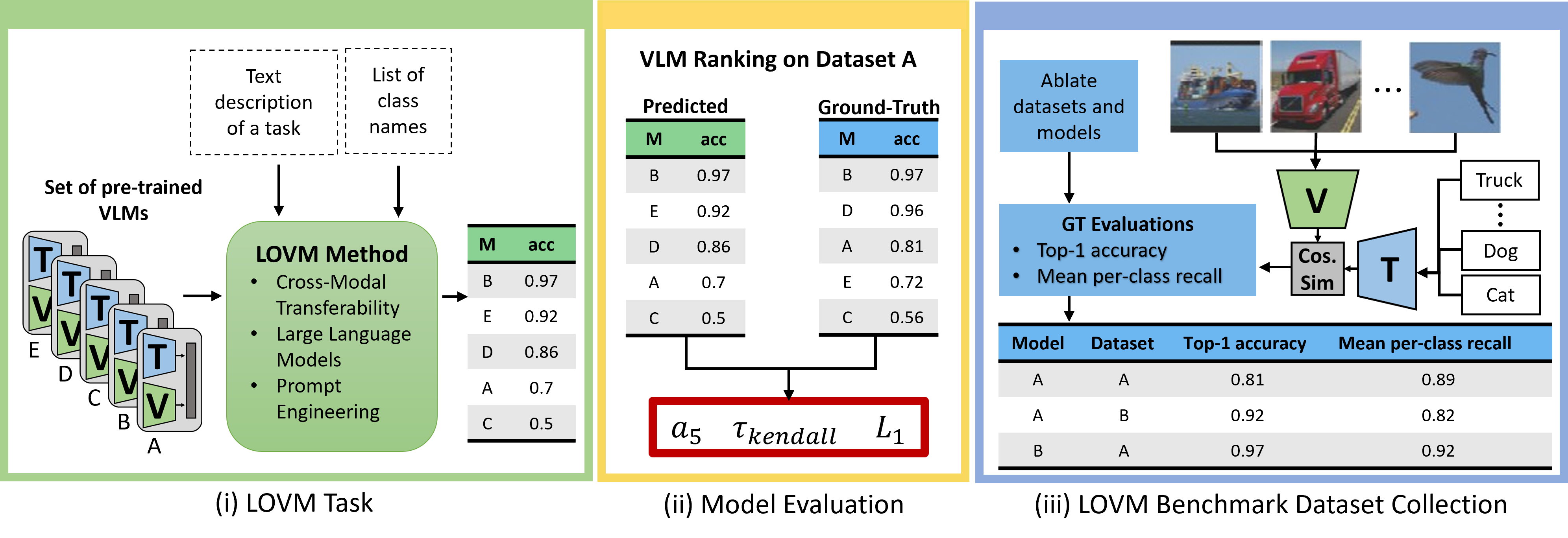}
  }
  %%%PREPRINT%%%\vspace{-0.3in}

    \caption{\textbf{Language-Only Vision Model Selection Overview.} (i) \textbf{Task.} a LOVM method is given a set of pre-trained VLMs, a text description of the desired task, and the list of the classes of interest. Given these, LOVM methods are expected to \textbf{rank} and \textbf{predict the performance} of all the available models on the downstream task. (ii) \textbf{Evaluation.} Given the 
    {\color{green}\textit{predicted (green)}} and {\color{blue} \textit{ground-truth (blue)}} VLM ranking and performance, we evaluate the LOVM method's performance by accepted list ranking and accuracy metrics.
    (iii) \textbf{Data Collection.} We exhaustively evaluated the selected \nummodels\ VLMs on the selected \numdatasets\ datasets to produce the ground-truth (image-based) evaluations.
    }
  \label{fig:task}
\end{figure*}

\section{Language-Only Vision Model Selection}
In order to train and evaluate LOVM methods, we need the ground-truth (GT) zero-shot performance, i.e., image-based evaluation of many VLMs (differing by architecture and pre-training) on many tasks and datasets.
Once collected, we can develop and evaluate LOVM methods.
An ideal LOVM method should be able to select the best performing VLM for a downstream vision task and estimate the performance directly from text embeddings, eliminating the cost of image-based model selection. 
The VLM, dataset selection criterion, and dataset collection procedure are detailed in Sec.~\ref{sec:task:dataset}. Finally, the evaluation protocol of LOVM methods is described in Sec.~\ref{sec:task:eval}. For a discussion on why we only evaluate zero-shot performace, see App. Sec.~\ref{sup:sec:lim}.

\paragraph{Background.}
We first recap how VLMs are used as in zero-shot vision tasks.  Given a pre-trained VLM $v$, along with an image $X\in \mathcal{X}$ or text $Y\in \mathcal{Y}$ input, we can obtain their $L_2$-normalized embeddings $\bm{x}$ or $\bm{y}$ from the image encoder $f_x:\mathcal{X} \mapsto\ \mathbb{R}^n$ or the text encoder $f_y:\mathcal{Y} \mapsto\ \mathbb{R}^n$, where $n$ is the dimension of the shared multi-modal embedding space. 
To use a model $v$ on a particular task, one encodes the class prompts, $Y^c$ for class $c$ using the model's text encoder, producing the class embeddings $\bm{y}^c = f_y(Y^c)$. 
To produce the final class prediction, one calculates the cosine similarity of an image embedding with all the corresponding text embeddings to 
predict the class logits.

\paragraph{Task Definition \label{sec:task:def}}
In the LOVM task, for any downstream application/dataset  $d$, methods are given a set of pre-trained VLMs, $\bm{V}=\{v_0, v_1,..\}\in\mathcal{V}$, a text description of the downstream task $Y_d$ (e.g., classification) and a list of the desired class names $Y^c_d, \forall c \in C_d$ where $C_d$ is the number of classes in task $d$. Given this, LOVM methods are expected to \textbf{rank} and \textbf{predict the accuracy} of the set of models (see Fig.~\ref{fig:task}, i):
\begin{equation}
    p_{v,d} = f_{\text{LOVM}}(v, \{Y^c_d\}_{c=1}^{C_d}, Y_d) , \;\; \forall \; v \in \bm{V},
\end{equation}
where $p_{v,d} \in \mathbb{R}$ is the relative/absolute performance of model $v$ on dataset $d$.

\subsection{Data Collection\label{sec:task:dataset}}
To train and evaluate LOVM methods, we need the \textbf{zero-shot} ground-truth performance of many VLM models on many downstream datasets.
We, therefore, selected \nummodels\ VLMs and \numdatasets\ Datasets and then performed image-based evaluations of each model on all the datasets - a total of 805 evaluations using the same prompting strategies discussed by~\citet{CLIP}, See Fig.~\ref{fig:task}, iii. 
\textbf{These ground truth zero-shot image-based model rankings and accuracies constitute the bulk of our benchmark.}
The proposed LOVM benchmark consists of the aforementioned evaluation tables as well as the per-dataset prompting templates, class names, and domain descriptions.  

\input{tables/dataset_info_small}
\paragraph{Selected Datasets.} The proposed LOVM benchmark utilizes a heterogeneous assortment of \numdatasets\ datasets. These datasets exhibit variability in the number of classes, their target tasks, and corresponding domains. 
The benchmark encompasses a comprehensive range of tasks such as classification, scene understanding, geolocalization, and object counting, rendering it extensively applicable across many applications. Further, the datasets span diverse domains, including natural, satellite, text, and medical images (See Tab.~\ref{tab:dataset_info_small}). 
To ensure maximal compatibility, we have opted for tasks that permit the utilization of the same VLM architecture, precluding any requisite alterations or additional training.
This approach necessitated the exclusion of tasks such as segmentation and object detection, which mandate additional training modules, introducing extraneous noise during the evaluation of VLM performance.

\paragraph{VLM Model Candidates. } 
We utilize the open-clip library~\citep{open_clip}, a diverse collection of pre-trained VLMs spanning various architectures, including but not limited to CLIP and CoCa models, and utilizing encoders such as ResNet, ConvNext, and ViT. These models have undergone pre-training on various datasets, such as WIT~\citep{CLIP}, LAION 400m, and LAION 2b~\citep{LAION}, with different hyperparameters. From the 87 models currently available, we have carefully selected \nummodels\ for our study. A comprehensive list of all models used in this benchmark can be found in the App. Tab.~\ref{tab:modeldataset}. We avoided incorporating additional multi-modal models, such as BEIT\citep{beit3} and VLMO~\citep{vlmo}, as these models utilize a shared text-image encoder and, therefore, cannot be evaluated on the same datasets as CoCa and CLIP. Utilizing models from the open-clip library ensures maximum compatibility and reproducibility in our work. Currently, CLIP models comprise a significant portion of VLMs employed in practice.

\subsection{LOVM Evaluation Protocol \label{sec:task:eval}}
On our benchmark, methods are expected to rank \nummodels\ pre-trained multi-modal models that differ in architecture and pre-training datasets on \numdatasets\ target datasets, and compare these rankings to the ground-truth rankings (see Fig.~\ref{fig:task} (ii)) and report the performance for each of the \numdatasets\ datasets as well as their averaged values.
\paragraph{Model Ranking.} 
When evaluating model ranking on a particular dataset, one has access to the performance of all the models on all the datasets besides the one being evaluated. We use the following metrics:
\begin{itemize}
    \item \textit{Top-5 Recall} ($R_5$) -- We used $R_5$ to evaluate a LOVM method's model ranking 
    capability. It is defined as the ratio of correctly identified models.  \\

     \item \textit{Kendall's Rank Correlation} ($\tau$) -- We used $\tau$ to evaluate a LOVM method's model selection capability and give s fine-grained picture of how well the method ranked the high-performing models and is defined as Kendall's rank over the top-5 selected models.

\end{itemize}

\paragraph{Performance Prediction.}
When evaluating a model's prediction on a dataset, the GT performance of that model on all datasets and the performance of all models on that dataset are held out. 

\begin{itemize}
    \item
     \textit{Mean Absolute Error} ($L_1$) -- We used $L_1$ to evaluate a LOVM method's performance prediction capability. 
     Specifically, we compute the $L_1$ loss of all models' predicted vs. actual mean per-class recall/top-1 accuracy. 

\end{itemize}

%% file: tables/dataset_info_small.tex
\begin{wraptable}[21]{r}{77mm}\vspace{-0.1in}
\caption{Details on the different datasets used, including the number of classes, tasks, and domain.}
\label{tab:dataset_info_small}
\vspace{0.6em}
\centering
\resizebox{0.53\textwidth}{!}{
\begin{tabular}{c|ccc}
\toprule
Dataset & Classes & Task & Domain \\
\midrule
Imagenet & 1000 & classification & natural image \\
Stanford Cars& 196 & classification & natural image \\
Flowers102 & 102 & classification & natural image \\
CIFAR100 & 100 & classification & natural image \\
GTSRB & 43 & classification &  natural image  \\
VOC2007 & 20 & classification & natural image \\
Oxford Pets & 37 & classification & natural image \\
STL10 & 10 & classification & natural image \\

DTD & 46 & classification & textural image \\
RESISC4 & 45 & classification& satellite images  \\
EuroSAT & 10 &  classification & satellite images  \\
MNIST & 10 & classification & hand-writing \\
Retinopathy & 5 & classification  &  retina scan\\
PCam & 2 & classification & histopathology \\
SUN397 & 397 & scene und. & natural image \\
Country211 & 211 & geolocation & natural image \\
SVHN & 10 & OCR  &  natural image\\
CLEVR-C & 8 & object counting & natural image \\
CLEVR-D & 8 & distance est.  & natural image \\
DMLab & 6 &  distance est. & synthetic \\
FER2013 & 7 & fac. exp. rec.  &  natural image\\
KITTI & 4 & distance est. & natural image \\
Rendered SST2 & 2 & OCR & text image \\

\bottomrule
\end{tabular}}
\end{wraptable}

%% file: sections/baselines.tex
\section{LOVM Baselines}
\label{sec:meth}

The assessment of model performance in traditional supervised methods often relies on benchmark dataset performance. Given that most pre-trained vision-language models (VLMs) are evaluated on ImageNet, it is convenient to utilize it as a baseline for comparison (This is our ImageNet Benchmark baseline).
Alternatively, a large language model could generate many probable image captions, which could be encoded using the different VLMs text encoder, producing the corresponding text embeddings.
Treating these embeddings as image-proxies, one can calculate different widely-accepted scores (see Sec.~\ref{sec:meth:features}) and fit a linear regression model to predict performance or rank VLMs. Specifically, from every VLM-dataset combination, one extracts these scores and then fits the model: 
\begin{align}
p_{v, d}&=\bm{w}\cdot \bm{s}_{v, d}+b,   \\
s^i_{v, d} &= f_{\text{feat}}^i  (v, \texttt{TextGen}( \{Y^c_d\}_{c=1}^{C_d}, Y_d)),
\end{align}
where $p_{v,d} \in \mathbb{R}$ is the relative/absolute performance of model $v$ on dataset $d$, $\bm{w}$, $b$ are the weights and bias of the linear model. 
$s^i_{\text{v, d}}$ is the $i$-th element in the score vector, $\bm{s}_{v,t} = [s^1_{\text{v, d}}, s^2_{\text{v, d}}, ... ]^T$, produced by the corresponding feature/score function $f^i_{\text{feat}}$. The function $\texttt{TextGen}$ is a function that generates text given the class names, $\{Y^c_d\}_{c=1}^{C_d}$ and task description $Y_d$ of the desired task/dataset $d$. 
We discuss the different scores, $s^i_{\text{v, d}}$, in Sec.~\ref{sec:meth:features} and the $\texttt{TextGen}$ function in Sec.~\ref{sec:meth:gen_datasets}.  
To evaluate model rankings on a dataset, we hold out the data for that particular dataset and fit a linear model on all the other datasets. Meanwhile, to evaluate the performance prediction of some model on a particular dataset, we hold out the data for that dataset and model and fit a linear model on the remaining combinations. 
We refer to the baselines by the combination of scores used in the linear model.

\subsection{Text Data Generation}
\label{sec:meth:gen_datasets}
The impressive progress in large language models (LLMs)~\citep{GPT4, LLAMA} has rendered the generation of potential - and realistic - `image captions' practically limitless, thus rendering text data generation remarkably attainable. In our study, we employ GPT-3.5, tasked to produce two distinct text-based datasets, each corresponding to a given vision task. 
These generated datasets serve as the foundation for extracting essential features for our task.

\paragraph{Captions Dataset.}  
To generate the captions dataset, $\bm{D^{\text{cap}}}$, we prompt an LLM to generate realistic - but confusing - captions for images containing the user-provided classes in the user-provided domain. We extracted the dataset description and class names from each dataset and prompted the LLM: 
\begin{quote}
Generate long and confusing image captions for the \{domain\} domain, which will be used to evaluate a Vision-Language Model's \{task\} performance. 

Generate 50 captions for \{classname\}:
\end{quote}
Where we assume the user supplies the target domain and task description. For examples of different dataset's domain and task, see Tab.~\ref{tab:benchmark_info}. %%%PREPRINT%%%

\paragraph{Synonyms Dataset.} 
Prior studies have already leveraged synonyms to evaluate LLMs~\citep{syn_metric1}. 
For example, if an LVM has seen many instances of the class `chair' referenced as a `chair', `seat', etc., we expect these embeddings to be closely located in the shared embedding space. 
To evaluate this aspect of the VLM using text, we  prompt an LLM to generate a  list of semantically similar/synonyms for every object class. For example, for the class ``chair'', we get: ``seat'', ``bench'', ``armchair'', and ``furniture''. We prompt the LLM with the following: 
\begin{quote}
Please list the superclasses/synonyms for \{classname\}. For example: 

chair: [furniture, seat, bench, armchair, sofa]

\{classname\}:
\end{quote}

We collect the results from this prompt to form the synonyms dataset, $\bm{D^{\text{syn}}}$.
\newpage %%%PREPRINT%%%

%%%PREPRINT%%%\vspace{-0.7in}
\subsection{Text-Derived Scores} \label{sec:meth:features}
There are many widely reported metrics for model transferability, dataset difficulty, and dataset granularity scores developed on image embeddings. We extract different commonly used features/metrics from the text dataset embeddings and calculate their text-only counterparts.

\paragraph{Text Classification Scores (C).}
We use the generated captions dataset as image proxies and evaluate the resulting model performance. 
Specifically, we replace the images with the generated image captions and evaluate each model's text top-1 accuracy, \textbf{text-acc1} and f1-score \textbf{text-f1}.

\paragraph{Dataset Granularity Scores (G).} 
~\citet{gran} introduced the use of two widely used dataset granularity measures for image classification, \intraClassClose~\citep{fisher}, \intraClassCloseSymbl\ and \intraSim~\citep{silBase}, \intraSimSymbl, and their normalization constant, \interSim, \interSimSymbl. The \intraClassClose\ measures the degree of similarity of the classes or the extent of their separation. The \intraSim\ is a well-established metric used to quantify the tightness of the same-class samples to the separation of different-class samples. The \interSim\ quantifies the degree of same-class tightness or data cone radius. 

Recently, ~\citet{syn_metric1} has shown that synonym consistency can be used in large language models to correlate the degree of familiarity of the model with a particular concept. Using the Synonym dataset, we compare the cosine similarity between the text embedding of each class and its corresponding synonyms. A high \superclass, \superclassSymbl, between the class and its corresponding synonyms indicates that the model is aware of the semantic meaning of the class.

For detailed definitions of these metrics, see App. Sec.~\ref{sup:sec:method_details}.

\vspace{0.1in} %%%PREPRINT%%%

%% file: sections/results.tex
\section{Experiments and Results}
In Sec.~\ref{sec:res:modelselection}, we evaluate the model selection capabilities of the proposed baselins on the LOVM benchmark. In Sec.~\ref{sec:res:pred}, we evaluate the proposed baselines performance prediction cababilities. We then analyse score trends and draw insights in Sec.~\ref{sec:res:trend}.

\input{tables/lovm_main}

\subsection{Model Selection\label{sec:res:modelselection}}
A core aspect of this benchmark is model selection, as it allows the user to quickly and easily select the optimal model for the desired downstream task. From Tab.~\ref{tab:main_table_1_sum}, we can see that, when predicting/ranking by the models mean per-class recall, the (C+G)-baseline can achieve a top-5 recall of $0.261$, indicating that, on average, more than one model is correctly ranked as a top-5 performing model. Meanwhile, the INB-baseline had a $R_5$ of $0.505$. Combining the text and ImageNet scores, the (INB+G)-baseline achieves the highest accuracy of $0.539$, a $\sim 15\%$ improvement over the INB-baseline. When studying Kendall's rank correlation, the (G+C)-, INB-, and (INB+G)-baselines achieve a $\tau$ of $0.101, 0.186$, and $0.214$, respectively. The (INB+G)-baseline had the highest rank correlation with $\sim 6\%$ improvement over the INB-baseline. Similar results can be seen when predicting the top-1 accuracy. The consistent improvement of the baselines over the INB-baseline indicates the utility of both text-based and benchmark features. Interestingly, C-score (or the text-acc1) appears to be more influential in predicting/ranking model's the top-1 accuracy than the mean per-class recall.

\subsection{Performance Prediction\label{sec:res:pred}}

\begin{wrapfigure}[17]{t}{0.55\textwidth}\vspace{-0.35in}
\begin{center}
\includegraphics[width=0.55\textwidth]{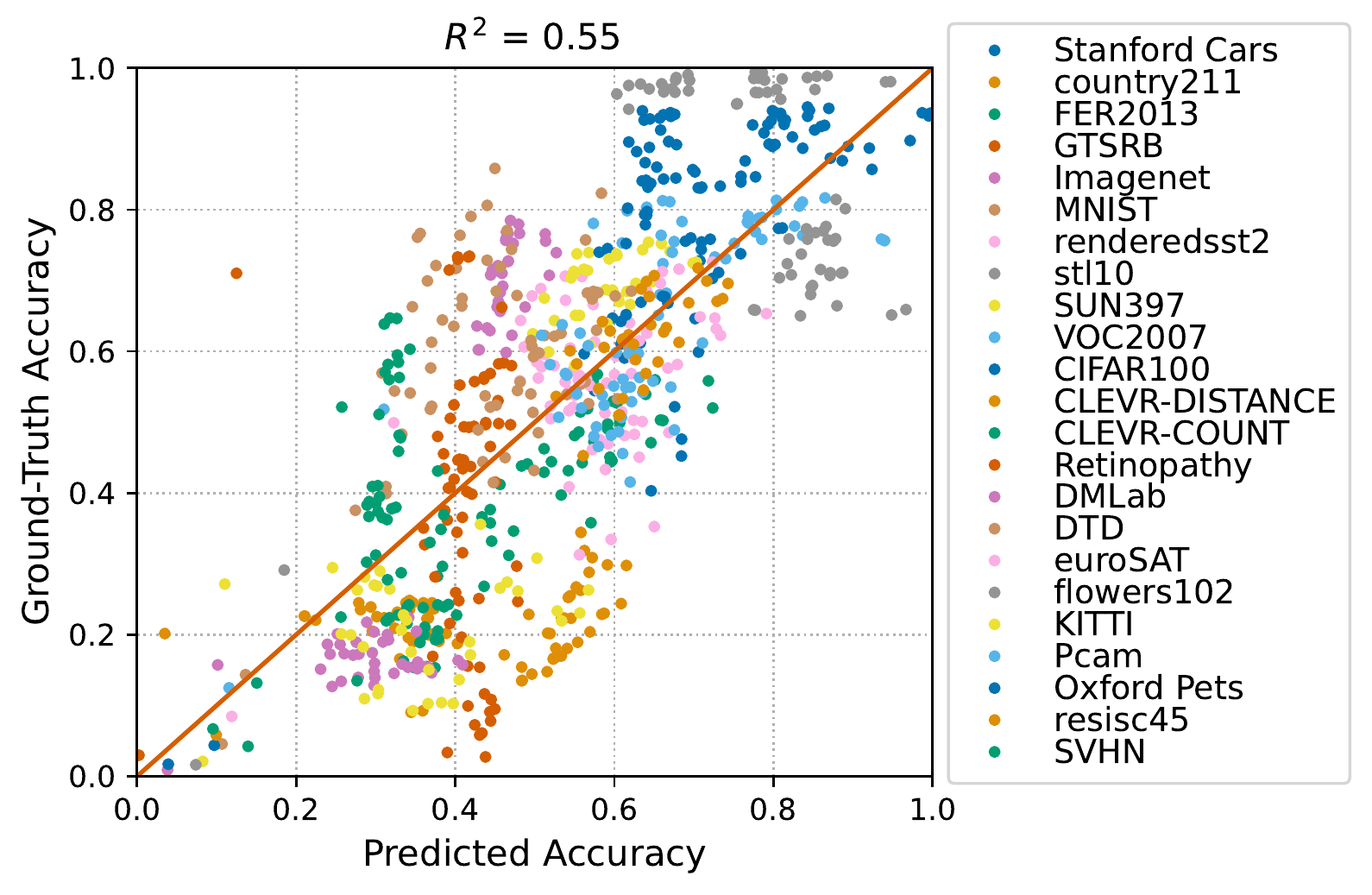}
\end{center} \vspace{-0.07in} 
   \caption{\textbf{Predicted vs. Ground-Truth Accuracy}. Predicted vs. actual top-1 accuracy on the proposed LOVM benchmark. 
   }
   \label{fig:predVSacutal}
\end{wrapfigure}
Based on Tab.~\ref{tab:main_table_1_sum}, it is clear that the granularity scores (G) are instrumental to predicting a model's top-1 accuracy and mean per-class recall. The G-baseline approach can achieve an average $L_1$ error of 0.145 and 0.141 for predicting the mean-per-class recall and top-1 accuracy, respectively. Adding any other scores does not lead to an improvement in performance prediction. 
The INB-baseline, which uses Imagenet performance as prediction, leads to a much higher $L_1$ error of 0.228 and 0.220 compared to the text-base baselines (text-based performance estimation outperformed INB-baselines by $\sim36\%$). 
Finally, adding the ImageNet benchmark score to the text features in the Unified baseline did not improve the $L_1$ compared to the text-only baseline. This is expected as the imagenet performance cannot be used to predict the performance on a different dataset. 
Fig.~\ref{fig:predVSacutal} shows the predicted vs. ground-truth accuracy. Our approach had a $R^2$ 
score (or coefficient of determination) of 
$=0.55$, showing significant room for improvement in accuracy prediction.

\vspace{0.1in}%%%PREPRINT%%%

\subsection{Insights into VLM Behavior\label{sec:res:trend}}
\begin{figure*}

    \centering

  \resizebox{1\textwidth}{!}{
  \includegraphics{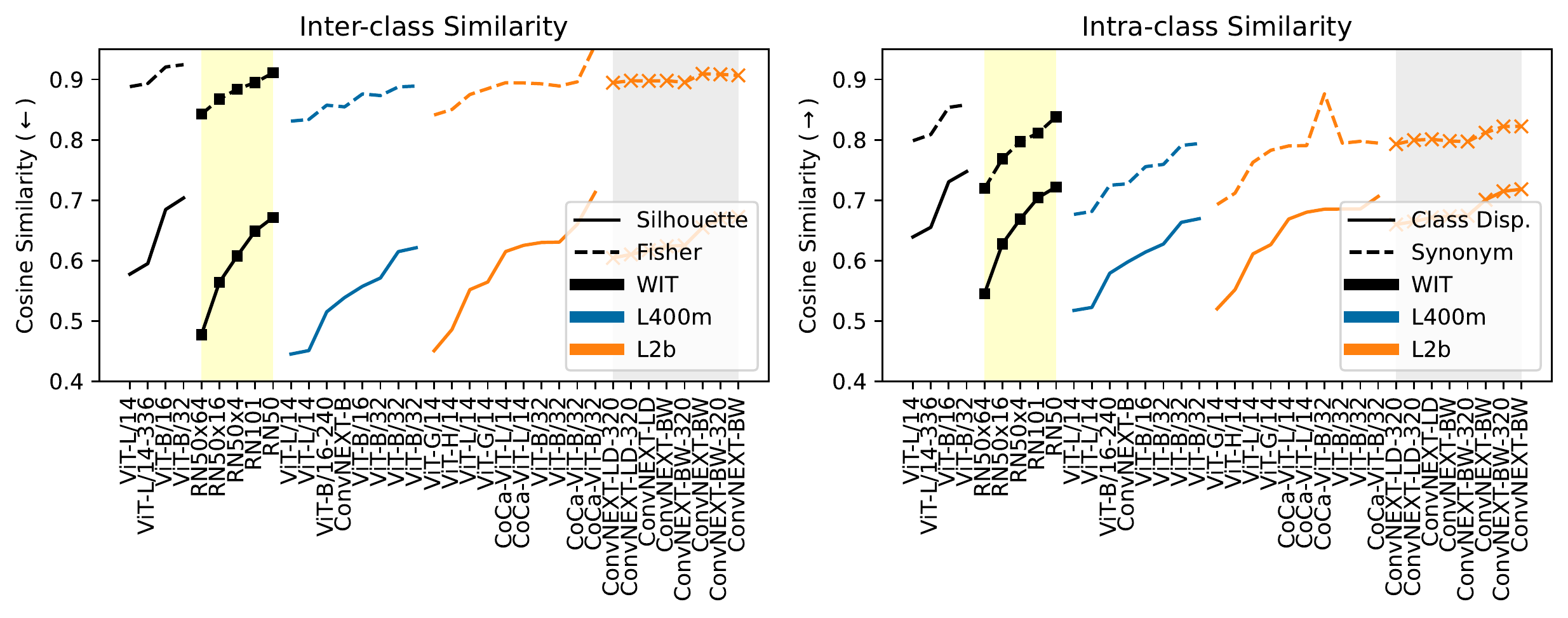}
  }

    \caption{\textbf{Analyzing Score Trends.}  Average text scores dependence on pre-training datasets and  model architecture on our text-derived scores.
     %%%PREPRINT%%%
    (left) scores quantifying inter-class similarity (right) scores quantifying intra-class similarity.
    ResNet (\myboxa) and ConvNext (\myboxb) based models are grouped separately to evaluate their effect on the score trends.
  }%%%PREPRINT%%%\vspace{-0.1in}
  \label{fig:featureAnalysis}
\end{figure*}

In this section, we visualize the dependence of the text-derived features on the pre-training datasets and model architectures while averaging them across the different datasets (see Fig.~\ref{fig:featureAnalysis}). 

\paragraph{Model Size.} From studying Fig.~\ref{fig:featureAnalysis}, we can we can identify a clear trend of \intraClassClose\ and \intraSim\ improving with model size, while \interSim\ and \superclass\ degrade with model size. \intraSim\ quantifies the degree of inter-class overlap or the degree of overlap between different classes in the embedding space. As the model size of the visual encoder increases, the embeddings from different classes become more and more orthogonal, decreasing the inter-class overlap. \intraClassClose\ quantifies the degree of granularity a model perceives the target datasets to be. As model size decreases, \intraClassClose\ decreases, or the degree of perceived granularity increases.  
\interSim\ quantifies the degree of intra-class dispersion, or how similar embeddings of the same class are. Specifically, as we increase model size, \interSim\ decreases, and therefore the class embeddings become more varied, effectively expanding the class cone radius. \superclass\ quantified the closeness of a class to its synonyms and behaved similarly to \intraClassClose.

\paragraph{Pre-training Dataset.}
When studying the effect of pre-training dataset size, it is clear that there is a positive correlation between pre-training dataset size and all of the metrics when comparing models of the same size. 
As the pre-training dataset increases, the intra-class simularity increases more rapidly than the inter-class simularity, hence effectively different classes are more seperated. Specifically, \intraClassClose\ and \intraSim\ increase, or the degree of perceived granularity decreases, and 
embeddings from different classes become less orthogonal, increasing the inter-class overlap. 
As the pre-training dataset size increases, \interSim\ increases and the intra-class dispersion is more condensed, leading to a smaller effective radius of a class dataset cone. Interestingly, larger models are more affected by the increase in dataset size (as seen by the large slope of ViT-L compared to ViT-B) - which could explain previous works' observation that larger models benefit more when trained on larger datasets~\citep{robustness}.

\paragraph{Model Architecture.}
Pre-training datasets and model architectures significantly influence each other. ResNets and ViTs, for instance, consistently demonstrated differing behaviors and appeared to reside at distinct points on the class separation-dispersion trade-off curve. In particular, ResNets displayed lower \interSim\ and \intraSim, indicating challenges in encoding instances of the same class within the feature space compared to ViTs. This may account for ResNets' superior performance on datasets with low visual variation, like MNIST; as the visual variation is relatively low, we would not expect the \interSim\ to be the limiting factor in model performance, making them less affected by this aspect of the dataset. Intriguingly, ConvNEXT models exhibited characteristics more in line with ViT-base models than ResNet-based ones. What leads to variation between WIT and L400m remains unclear, necessitating further investigation.

%% file: tables/lovm_main.tex
\begin{table}[t]
\caption{\textbf{LOVM Benchmark}. We evaluate our method's performance over \numdatasets\ datasets and \nummodels\ pre-trained models, and when predicting the top-1 accuracy and mean per-class recall (averaged over all datasets, for the per-dataset breakdown, see App. Tab.~\ref{tab:main_table_acc} and~\ref{tab:main_table_mpcr} ).  INB - ImageNet Baseline, C - Text Classification scores, G - Granularity scores. As can be seen, mixed aproaches achieve the best VLM ranking and performance prediction.
}
\label{tab:main_table_1_sum}
\centering
\adjustbox{width=0.7\textwidth}{
\begin{tabular}{c|cc|c||cc|c}
\toprule
used & \multicolumn{3}{c||}{mean per-class recall} & \multicolumn{3}{c}{top-1 accuracy} \\

scores& $R_5 (\uparrow)$ & $\tau(\uparrow)$ & $L_1(\downarrow)$ & $R_5(\uparrow)$ & $\tau(\uparrow)$ & $L_1(\downarrow)$\\
\midrule

INB     &  0.504  & 0.186   & 0.228  & 0.452  & 0.177   & 0.220\\ 
C       &  0.252  & 0.058   & 0.182  & 0.226  & 0.058   & 0.176\\ 
G       &  0.270  &  -0.014 & \textbf{0.145}  & 0.252  & -0.014  & 0.143\\ 
G+C     &   0.261 & 0.101   &  \textbf{0.145} & 0.252  &  -0.014 & 0.143\\ 
INB+C   &  0.522  & \textbf{0.214}   & 0.182  & \textbf{0.461}  & \textbf{0.223}   & 0.176\\ 
INB+G   &  \textbf{0.539}  & 0.197   & \textbf{0.145}  & \textbf{0.461}  & 0.096   & \textbf{0.141}\\
INB+G+C &  \textbf{0.539}  &  0.197  &  \textbf{0.145} & \textbf{0.461}  & \textbf{0.223} &  \textbf{0.141}\\ 

    \bottomrule
\end{tabular}}

%%%PREPRINT%%%\vspace{-0.1in}

\end{table}

%% file: sections/related_works.tex
\section{Related Work} 
\vspace{-0.07in}

\paragraph{Vision-Language Models.} The field of vision-language models (VLMs) has witnessed significant progress in recent years, particularly with the introduction of contrastive pre-trained VLMs such as CLIP~\citep{CLIP}. These models leverage large-scale datasets of aligned image-caption pairs to obtain shared embedding spaces that capture rich visual and textual features.
The learned image and text encoders from these VLMs have demonstrated impressive feature extraction capabilities and even set state-of-the-art zero-shot performances. 
However, the performance of VLMs can vary significantly across different datasets, especially when there exists a domain, content, or distribution shift~\citep{robustness}. 
As the number of model architectures \& pre-training datasets~\citep{open_clip, LAION} increase, it is challenging to select a pre-trained VLM, as good performance on existing benchmarks does not always translate to the downstream task. 
Therefore, there is a need to develop strategies that can estimate VLM performance on a new task without requiring an exhaustive evaluation of these models using the target dataset.  

\paragraph{Text as a Proxy For Images.}
While these VLMs aim to project representations from different modalities into a shared embedding space,~\citet{ModalityGap} found that corresponding image and text pairs don't completely overlap in the embedding space. Instead, a ``modality gap'' exists between the image embeddings and text embeddings sub-space. 
Subsequently,~\citet{drML} has found that this gap can be approximated as an orthogonal constant between true pairs of image and text and is, therefore, parallel to the decision boundaries for a given modality. This suggests that cross-modality transferability - using one modality as input to the other's classifier - is possible for these contrastively pre-trained VLMs. 
Severak studies have demonstrated the utility of the cross-modality transferability phenomenon in different tasks. For instance, Domino leveraged the cross-modal embeddings to identify error slices and generate natural language descriptions of the error slices~\citep{DOMINO}. Similarly, ~\citet{MMFailure} used these embeddings to discover model failure directions in the multi-modal embedding space. Meanwhile, ~\citet{drML} proposed the DrML, which diagnoses and rectifies vision classifiers using natural language inputs. In this study, we also use text as a proxy for images, but for the novel task of ranking and estimating VLM performance.

\paragraph{Unsupervised Model Selection.} Unsupervised model selection was recently introduced by \citet{modelRanking1}, to select the best model for a new target domain without utilizing labeled data. Their work only considered domain (and not content) shifts and proposed constructing a proxy dataset that captures/closely approximates this shift. This proxy dataset is constructed by minimizing different dataset statistics using several labeled datasets. Evaluating models on this proxy set performs well for model selection/ranking. However, such a strategy is limiting in the setting of evaluating VLMs - the size of these models and their pre-training datasets makes it too computationally expensive to achieve the desired goal of evaluating model performance on \textit{any} downstream task.

\paragraph{Unsupervised Accuracy Estimation.} 
Unsupervised or label-free accuracy estimation aims to estimate classifier model performance with only access to the unlabeled test set of a new task. ~\citet{UnlabeledAccuracyLogic, UnlabeledAccuracyBayesian} proposed strategies to apply probabilistic modeling approaches, such as the probabilistic logic or Bayesian modeling, to analyze and aggregate predictions from multiple classifiers. Other works approach this task by fitting models on feature statistics of the target dataset~\citep{UnlabeledAccuracyLearn}. Some studies evaluated model agreement, where many classifiers are used on the target dataset, and the degree of agreement was correlated with model performance~\citep{UnlabeledAccuracyAgree, UnlabeledAccuracyDisAgree}
Other approaches for unsupervised accuracy estimation include training a neural network on the weight distribution statistics~\citep{UnlabeledAccuracyWeights} or composing a meta-dataset with available datasets, such that the meta-dataset matched some target dataset statistics~\citep{UnlabeledAccuracyDatsets}. 
Some have attempted to craft embedding-based scores, trying to quantify the separability of clusters in the embeddings spaces~\citep{Transferability2, Transferability1}.
All these methods assume access to the unlabeled dataset of the target task. Instead, our method only requires text descriptions of the novel task to estimate the model's performance.

%% file: sections/conclusions.tex
\section{Conclusion}
%%%PREPRINT%%%\vspace{-0.1in}
In this work, we introduce a new problem setting and task LOVM, which aims to select the best-performing VLMs for a downstream vision task by only using its textual description. 
To demonstrate the feasibility of such a task, we show how large language models, in combination with the cross-modal transferability phenomenon, can be leveraged for such a task.
We exhaustively test these methods on the proposed LOVM benchmark, consisting of  \nummodels\ VLMs and \numdatasets\ benchmark datasets.
Our findings validate the viability of our proposed LOVM task, with unified (both text scores and ImageNet benchmarking) baselines outperforming the ImageNet benchmarking baseline. 
This suggests that text-based model selection methods (i.e., LOVM methods) provide additional benefits to baseline selection based on a model's performance on ImageNet.
Furthermore, we found that the granularity-based scores influence performance prediction and modal ranking more greatly. 
These findings bolster the research direction of developing methods for VLM selections using only text.

Our proposed LOVM benchmark aims to foster this research direction.
We see two promising avenues for future research: (i) improving text-based classification correlation with ground-truth accuracy by either text generation, evaluation metrics, or cross-modal transferability, and (ii) introducing new granularity and transferability scores to the text-only paradigm. Namely, we anticipate the development of methods improving over our proposed baselines presented in Tab.~\ref{tab:main_table_1_sum}.
 Our work aims to facilitate future research in this area and provide a more accurate and reliable means of comparing pre-trained VLMs, accelerating their utilization in downstream applications.

For a discussion about broader and potential negative societal impacts please see App. Sec~\ref{sup:sec:neg_impact}.

\paragraph{Acknowledgments.} We gratefully acknowledge the computational credits provided by Google Cloud Platform through Stanford's HAI Institute for Human-Centered Artificial Intelligence. We also thank the Knight-Hennessy Scholars Foundation for generously funding Orr Zohar.

%% file: sections/appendix.tex
\part{} 
\vspace{-0.5in}
{
\hypersetup{linkcolor=black} 
\parttoc 
}

\section{LOVM Benchmark Details\label{sup:sec:benchmark_details}}
We evaluated \nummodels\ on \numdatasets, a total of 805 evaluations. This constituted the bulk of our compute with a total of 4 days on an nvidia V100 instance. Evaluations were carried out using the CLIP\_benchmark repository (\url{https://github.com/LAION-AI/CLIP_benchmark}).

\subsection{LOVM Benchmark - Datasets\label{sup:sec:benchmark_details:dataset}}

The proposed LOVM benchmark comprises of \numdatasets\ datasets, which were selected to maximize diversity. Specifically, these datasets vary in the number of classes (2 to 1000), their target tasks, and domains. 
The benchmark encompasses a comprehensive range of tasks such as classification, scene understanding, geolocalization, object counting, and more, with the goal of rendering it extensively applicable across many applications. 
Further, the datasets span diverse domains, including natural, satellite, text, and medical images (See Tab.~\ref{tab:benchmark_info} for a comprehensive account of the datasets and their source). 
To ensure maximal compatibility, we have opted for tasks that permit the utilization of the same VLM architecture, precluding any requisite alterations or additional training. 
This approach necessitated the exclusion of tasks such as segmentation and object detection, which mandate additional training modules, introducing extraneous noise while evaluating VLM performance. However, it is worth noting that previous transferability works have shown that these approaches may generalize to more complex applications such as semantic segmentation \citep{bhattacharyya, segTransfer}.

\subsection{LOVM Benchmark - Vision-Language Models\label{sup:sec:benchmark_details:models}}
Tab.~\ref{tab:modeldataset} presents a list of models and their corresponding pre-training datasets used in the LOVM benchmark. 
We utilize the open-clip library~\citep{open_clip}, a diverse collection of pre-trained VLMs spanning various architectures, including but not limited to CLIP and CoCa models, and utilizing encoders such as ResNet, ConvNext, and ViT. These models have undergone pre-training on various datasets, such as WIT~\citep{CLIP}, LAION 400m, and LAION 2b~\citep{LAION}, with different hyperparameters. From the 87 models currently available, we have carefully selected \nummodels\ for our study. A comprehensive list of all models used in this benchmark can be found in Tab.~\ref{tab:modeldataset}. We avoided incorporating additional multi-modal models, such as BEIT\citep{beit3} and VLMO~\citep{vlmo}, as these models utilize a shared text-image encoder and, therefore, cannot be evaluated on the same datasets as CoCa and CLIP. Using models from the open-clip library ensures maximum compatibility and reproducibility in our work. Currently, CLIP models comprise a significant portion of VLMs employed in practice.
Tab.~\ref{tab:modeldataset} includes \nummodels\ entries, each identified by an ID number. The first four columns indicate the ID number, model name, model abbreviation, and pre-training dataset name. The fifth column shows the abbreviation of the pre-training dataset name. The models listed in the table include ResNet (RN50, RN101, etc.) and Vision Transformer (ViT) with different sizes (B/32, B/16, L/14, etc.), and the pre-training datasets include OpenAI's WIT dataset and two variants of LAION (L400m and L2b) datasets.

\subsection{LOVM Benchmark - Ground-Truth Model Ranking\label{sup:sec:benchmark_details:gt_rank}}
\begin{figure*}

    \centering
  \vspace{-0.1in}

  \resizebox{1\textwidth}{!}{
  \includegraphics{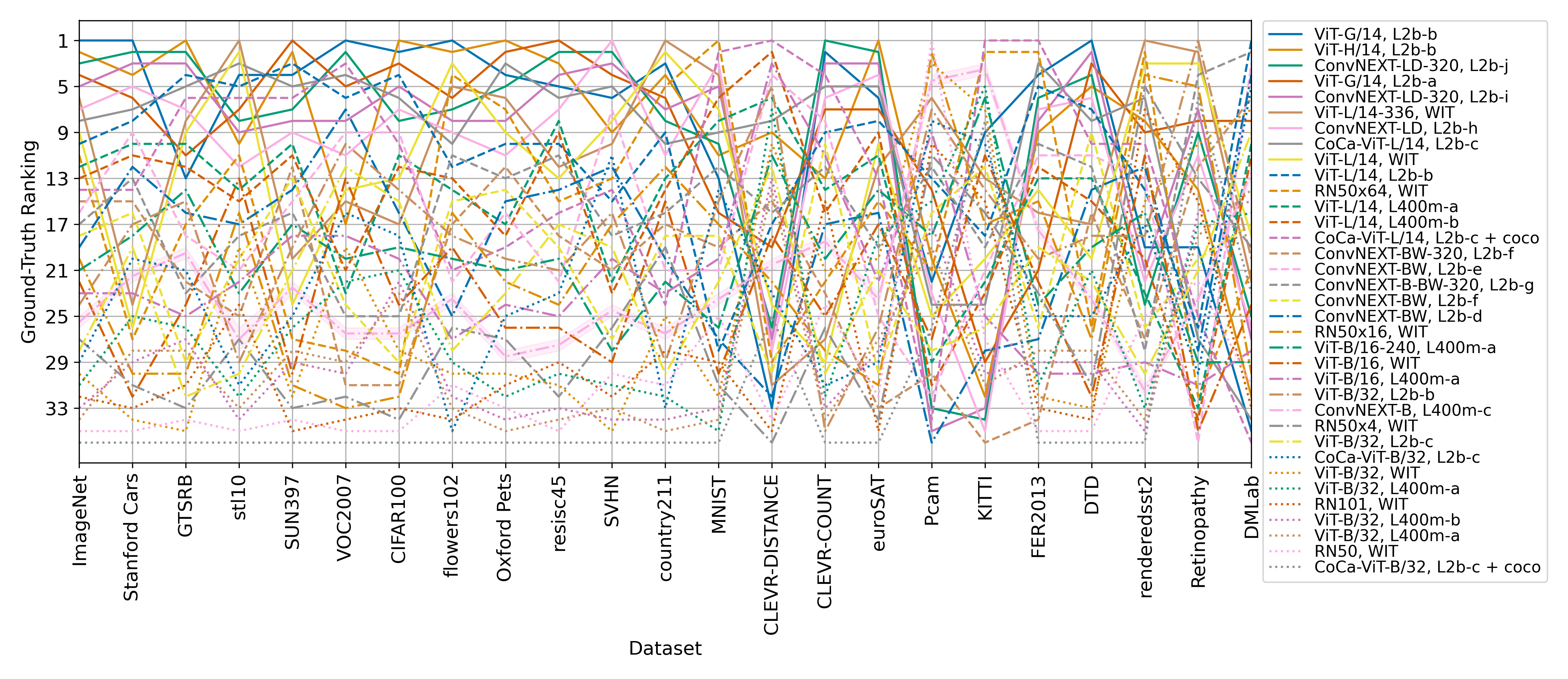}
  }

    \caption{\textbf{Ground-Truth VLM Ranking.} As can be seen, there is a lot of variation in the ground-truth model ranking across both the natural image (left) and other (right) benchmarks. 
  }
  \label{fig:gt_rank}
\end{figure*}

To evaluate the validity and generalizability of the LOVM benchmark, we first present the ground-truth model ranking over all datasets to show that the model order is not constant across the datasets. We organized the benchmarks from natural image classification (Fig.~\ref{fig:gt_rank}, left) to non-natural image / non-classification benchmarks (Fig.~\ref{fig:gt_rank}, right). As depicted in Fig.~\ref{fig:gt_rank}, the distribution exhibits a non-uniform pattern, indicating the utility of LOVM methods and the importance of VLM selection methods in general. Interestingly, ranking variations are more significant on the non-natural image / non-classification benchmarks. This exemplifies the need for LOVM methods to contend with content shift (i.e., changing what classes are in the target domain) and domain/task shift.

\input{tables/dataset_info}
\input{tables/dataset_nameing}

\section{Baseline Details\label{sup:sec:method_details}}
\input{sections/sup_method}

\section{Additional Results \label{sup:sec:res}}

\subsection{LOVM Per-Dataset Breakdown\label{sup:sec:res:quant}}
Here, we show the per-dataset breakdown of our main results. In Tab.~\ref{tab:main_table_acc}, we show our model ranking and performance prediction results for top-1 accuracy.  In Tab.~\ref{tab:main_table_mpcr}, we show our model ranking and performance prediction results for mean per-class recall.

\input{tables/lovm_acc}
\input{tables/lovm_mpcr}

\subsection{Ablation Experiments\label{sup:sec:res:ablations}}
To understand the utility of each of our extracted scores, we exhaustively ablated their effect on top-1 accuracy model ranking and performance prediction (Tab.~\ref{tab:ablation:rank_acc} and Tab.~\ref{tab:ablation:pred_acc}), and
mean per-class recall model ranking and performance prediction (Tab.~\ref{tab:ablation:rank_mpcr} and Tab.~\ref{tab:ablation:pred_mpcr}). Specifically, we ablated each score's impact on the resulting model's performance. As can be seen, using more than $\sim 3$ features at a time seldom improves performance. Future work can investigate the use of more sophisticated models that may be able to utilize more scores in predicting model ranking and performance. Specifically, for ranking models, text classification and scores quantifying intra-class similarity (\interSim\ \& \superclass) were the most dominant, while for performance prediction, granularity scores quantifying both inter- and inta- class similarity was the most important. This

We ablate the model ranking performance to understand each extracted score's effect on ranking models. The text classification scores and scores quantifying intra-class similarity were the most consequential in predicting model ranking. Specifically, in ranking models, the text-f1 score, \interSim\ (\interSimSymbl), and \superclass\ (\superclassSymbl) where the most dominant (Tab.~\ref{tab:ablation:rank_acc} rows 8 \& 11, Tab.~\ref{tab:ablation:rank_mpcr} row 38). Overall, it seems like the text classification excelled at fine-grained ranking (as quantified by $\tau$), while the inter-class granularity scores improved the coarse ranking prediction (as quantified by $R_5$. Meanwhile, granularity scores quantifying inter- and intra- class similarity were the most dominant for performance prediction. Specifically, \interSim\ (\interSimSymbl), \superclass\ (\superclassSymbl), and \intraSim\ (\intraSimSymbl) were the most influential (Tab.~\ref{tab:ablation:pred_acc} rows 26 \& 41, Tab.~\ref{tab:ablation:pred_mpcr} row 60). INB does not aid performance prediction, indicating that getting a course estimation of dataset difficulty dominates performance prediction.

\input{tables/ablation_acc1}
\input{tables/ablation_mpcr}

\subsection{Raw Model Ranking Details}
\begin{figure*}
    \centering
  \vspace{-0.1in}
    \resizebox{1\textwidth}{!}{
  \includegraphics{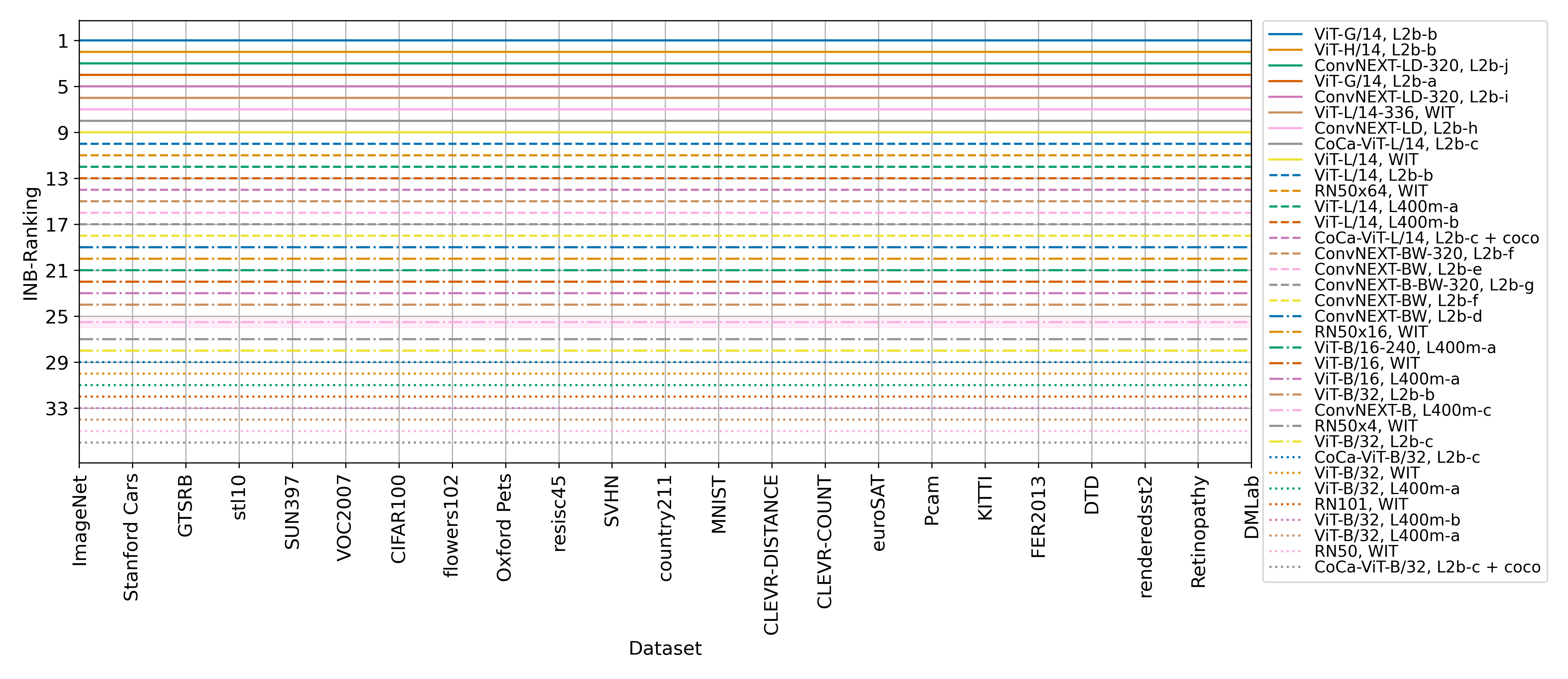}
  }
    \vspace{-0.1in}
  \resizebox{1\textwidth}{!}{
  \includegraphics{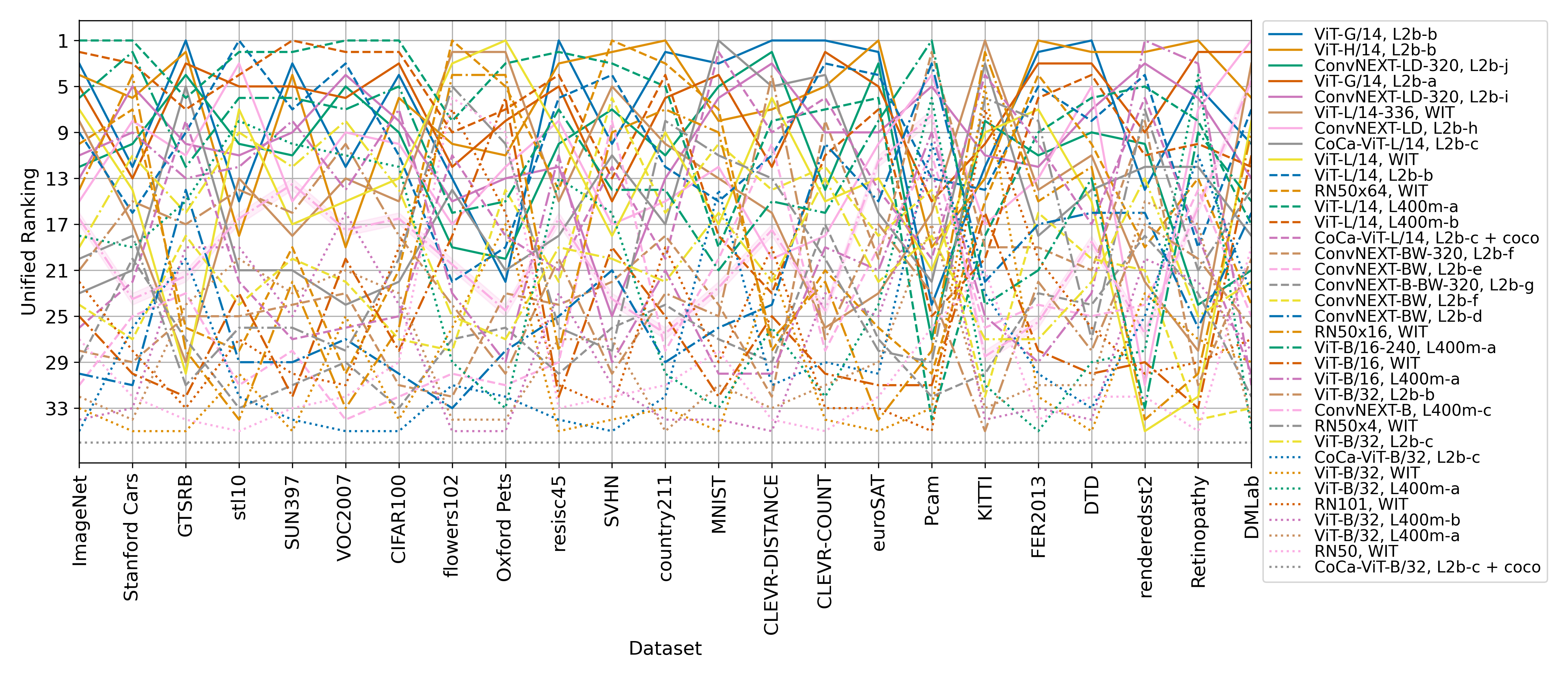}
  }
    \caption{\textbf{Raw Model Ranking.} (top) ImageNet benchmark approach assumes the same model ranking for all datasets and cannot predict fine-grained model ranking. (bottom) The unified approach can adjust the coarse ImageNet rankings for a more realistic model ranking.
  }
  \label{fig:modelrank}
\end{figure*}

To illustrate the model ranking of the naive (ImageNet Benchmark) baseline to some text-based approaches, we visualize the raw ranking prediction of each method. We sort the datasets from natural image classification (Fig.~\ref{fig:modelrank} left) to non-natural image / non-classification benchmarks (Fig.~\ref{fig:modelrank} right). 
Here, the evident failure of the ImageNet Benchmark baseline to capture dataset-specific changes in ranking is apparent. As the benchmark approach ranks models by their ImageNet performance, the model ranking is constant for all datasets. Meanwhile, integrating the text features produces a ranking distribution with a discernible positive correlation between the ground truth and the predicted model ranking. The unified approach also captures more significant ranking variation in the non-natural image / non-classification benchmarks. 

\subsection{Domain Shift Experiment}\label{sec:dshift}
An obvious obstacle to text-based model performance prediction methods is the difficulty in describing distribution shifts. For example, VLMs evaluated on ImageNet and ImageNet-v2 will get the same text-predicted accuracy while the actual performance differs. Meanwhile, for some domain shifts - like ImageNet and ImageNet sketch -  this shift can be described via text. We want to evaluate how capable text-only methods are at estimating the dataset difficulty under such shifts and compare them to well-accepted image-based approaches. 

\paragraph{Dataset Description Similarity.} We extract each dataset's description from either the abstract or introduction section of the original manuscript. Subsequently, we extract the text embeddings for all dataset descriptions using a pre-trained CLIP model. We then compute the cosine similarity between the descriptions of the downstream datasets to the original pre-training dataset to quantify how different the two datasets are. 

\paragraph{Prompt Embedding Similarity.} We use the cosine similarity between dataset-specific and generic class prompts to evaluate domain shift. Specifically, based on the original list of class prompts from CLIP, we pick the ones that best describe our target dataset. For instance, for the ImageNet-sketch dataset, we selected prompts such as ``A sketch of a \{c\}'' or ``A doodle of a \{c\}''. Then, we use the text encoder from a pre-trained CLIP model to extract embeddings from dataset-specific and generic class prompts and compute the cosine similarity between each pair. We use the mean cosine similarity to measure how similar the target dataset is to the pre-training dataset. The dataset-specific prompts can be found in the following subsection. 

\paragraph{Image-Text Embedding.} We wanted to compare with widely used dataset difficulty approaches. One common approach is to use the confidence of a model's prediction to determine a dataset's difficulty. To do so, we first $n$ images from the target dataset and extract image embeddings for each of the $n$ images. This simulates the scenario where we only have access to $n$ images from the target dataset to estimate model performance, where $n$ is much smaller than the dataset size. Then, we embed the class prompt into text embeddings and compute the prediction logits between each image embedding and class embeddings. Lastly, we compute the \textit{entropy score} \citet{v-usable_info}
and \textit{max prediction logit} \citet{datadiff} to determine dataset difficulty. 

\paragraph{Image Embedding Distance.} Another common approach is estimating the difference in distribution between the test and train sets~\citet{silClass}. We, therefore, use the distance between the target and pre-training image embeddings to quantify dataset difficulty. Similar to the image-text embedding approach, we first sample $n$ images from the target dataset to extract image embedding using a pre-trained CLIP image encoder. Additionally, we sample $m$ images from the pre-training dataset to extract image embeddings. We only sample $m$ examples since these VLMs are typically pre-trained on an internet-scale dataset, which makes it challenging to embed and compute distance measures on the entire pre-training dataset. We then compute the $L_2$ distances between the target and pre-training datasets. We use \textit{max, min \& mean} $L_2$ to quantify dataset difficulty.

\paragraph{Datasets.} To evaluate the feasibility and effectiveness of each, we use the following variants of ImageNet. Each dataset captures a different distribution shift from the original ImageNet: 
\begin{itemize}
  \item \textbf{ImageNet}: The original ImageNet dataset. 
  \item \textbf{ImageNet Version 2 (ImageNet-v2)}:  A new test set for ImageNet sampled a decade later. 
  \item \textbf{ImageNet Sketch (ImageNet-s)}: A ImageNet test set dataset with sketch-like iamges. 
  \item \textbf{ImageNet Rendition (ImageNet-r)}: contains art, cartoons, deviantart, graffiti, embroidery, graphics, origami, paintings, patterns, plastic objects, plush objects, sculptures, sketches, tattoos, toys, and video game renditions of ImageNet classes.
  \item \textbf{ImageNet Adversarial (ImageNet-a)}: A real-world distribution shift ImageNet dataset with changes in image style, blurriness, camera operation, and geographic location. 
\end{itemize}

We extract descriptions of each dataset from either the abstract or induction section of their original manuscript. The description used for each dataset is as shown here: 
\begin{itemize}
  \item \textbf{LAION400m}: \textit{``a dataset with CLIP-filtered 400 million image-text pairs.''}
  \item \textbf{ImageNet}: \textit{``a benchmark in object category classification and detection on hundreds of object categories.''}
  \item \textbf{ImageNet Version 2}: \textit{``three test sets with 10,000 new images each. Importantly, these test sets were sampled after a decade of progress on the original ImageNet dataset.''}
  \item \textbf{ImageNet Adversarial}: \textit{``real-world distribution shift datasets consisting of changes in image style, image blurriness, geographic locations.''}
  \item \textbf{ImageNet Rendition}: \textit{``art, cartoons, DeviantArt, graffiti, embroidery, graphics, origami, paintings, patterns, plastic objects, plush objects, sculptures, sketches, tattoos, toys, and video game renditions of ImageNet classes.''}
  \item \textbf{ImageNet Sketch}: \textit{``a new dataset consisting of sketch-like images, that matches the ImageNet classification validation set in categories and scale''}
\end{itemize}

The dataset-specific prompts used for the prompt embedding distance metric are listed in Fig.~\ref{sup:fig:templates}.

\input{tables/templates}

\paragraph{Evaluation.} We use Kendall's rank correlation ($\tau$) to evaluate our method's ability to rank the datasets in terms of their difficulties. Since \textit{image-text embedding} and \textit{ImageNet embedding distance} require sampling from the target dataset, we run our evaluation 1,000 times with different samples and compute the average metric. We also compute the standard deviations of the 1,000 run to estimate the variability of random samples. 

We show the results of using our strategies to estimate domain shift in Tab~\ref{tab:sup_dataset_difficulty}. Based on the results, it is clear that none of the current methods can capture the dataset difficulty. Furthermore, the variability based on the standard deviation makes our results heavily dependent on the samples drawn from the target dataset, again suggesting these approaches' limitations. 

\input{tables/domain_shift_rank}

\section{Limitations\label{sup:sec:lim}}

Our study, while extensive, is not without limitations. Primarily, our focus rests on zero-shot tasks due to the nature of the LOVM's design. The framework's primary aim is to determine the best model for a given task when there is \textbf{no} access to the downstream task dataset. Under these circumstances, fine-tuning or linear probing is not viable, as they require access to labeled or unlabeled images from the downstream task dataset. If such data were available, the more straightforward approach would be to address the conventional transferability problem as detailed in prior works.
The ideal scenario we envision for using LOVM is one where a user with minimal technical expertise seeks to conduct a vision task. In this situation, the user can utilize a LOVM method to discern the most suitable model and the relevant classes, enabling them to deploy the model without needing to delve into technical nuances.
However, if one possesses data for fine-tuning, conducting a direct evaluation on this small dataset is likely the most accurate course of action. This constraint stems from the fact that LOVM methods cannot make differential predictions without access to the fine-tuning data. Predicting the performance after fine-tuning or linear probing would heavily depend on the correlation between the results pre and post- fine-tuning/linear probing, a scenario we aim to avoid in the design of LOVM. However, previous work has shown some correlation exists, so there may be some transferability to fine-tuned/linear probed models~\citep{fewShotTransfer}.

Secondly, as discussed in Sec.~\ref{sec:dshift}, even datasets bearing identical content may encounter a domain shift. Such shifts can be clearly explained in some cases, such as when comparing ImageNet-regular/rendition/sketch, but in others, the shift may be more elusive. For instance, when comparing ImageNet to ImageNet-a, or when class distribution shifts occur, identifying the source of the shift becomes challenging. In these scenarios, LOVM methods might struggle to accurately predict the performance of a VLM, though model selection might be marginally affected.

Finally, while the utility of text-only solutions described in Sec.~\ref{sec:dshift} warrants continued investigation, it may be necessary to incorporate unlabeled test images to gauge domain shifts. Combining LOVM methods with these image-based evaluations remains a promising area of ongoing research.

\section{Broader Impacts\label{sup:sec:neg_impact}}
Our work simplifies selecting vision-language models (VLMs) for specific tasks, increasing the accessibility of artificial intelligence (AI) applications. However, this accessibility may be a double-edged sword. On the one hand, it could democratize AI applications, allowing smaller entities or independent researchers to utilize AI technologies more effectively. On the other hand, this easy access might also enable malicious entities to deploy harmful applications more readily, posing risks to sectors such as information security and personal privacy.

Moreover, despite our methodology's efficiencies, it carries the risk of sub-optimal model selection due to inherent limitations. Inaccuracies could lead to inefficient resource allocation or inferior performance in real-world applications, particularly in high-stakes fields such as healthcare or autonomous driving. Overall, while our work contributes to the efficiency and accessibility of AI applications, it highlights the need for vigilance and continuous refinement to mitigate potential negative impacts.

%% file: tables/dataset_info.tex
\begin{table}[t]
\centering
\caption{Details on the different benchmarks used in the study, including the number of classes, tasks, and target domain.\label{tab:benchmark_info}
}
\resizebox{0.9\textwidth}{!}{
\begin{tabular}{c|ccc}
\toprule
Dataset & Classes & Task & Domain \\
\midrule

Imagenet~\citep{deng2009Imagenet} & 1000 & classification & natural image \\
Stanford Cars~\citep{KrauseStarkDengFei-Fei_3DRR2013} & 196 & classification & natural image \\
Flowers102~\citep{Nilsback08} & 102 & classification & natural image \\
CIFAR100~\citep{krizhevsky2009learning} & 100 & classification & natural image \\
GTSRB~\citep{stallkamp2011german} & 43 & classification &  natural image  \\
VOC2007~\citep{pascal-voc-2007} & 20 & classification & natural image \\
Oxford Pets~\citep{parkhi12a} & 37 & classification & natural image \\
STL10~\citep{coates2011analysis} & 10 & classification & natural image \\
DTD~\citep{cimpoi14describing} & 46 & classification & textural image \\
RESISC45~\citep{Cheng_2017} & 45 & classification& satellite images  \\
EuroSAT~\citep{helber2019eurosat} & 10 &  classification & satellite images  \\
MNIST~\citep{lecun2010mnist} & 10 & classification & hand-writing \\
Retinopathy~\citep{retinopathy} & 5 & classification  &  retina scan\\
PCam~\citep{Veeling2018-qh} & 2 & classification & histopathology \\
SUN397~\citep{Xiao:2010} & 397 & scene und. & natural image \\
Country211~\citep{CLIP} & 211 & geolocation & natural image \\
SVHN~\citep{svhn} & 10 & OCR  &  natural image\\
Rendered SST2~\citep{CLIP} & 2 & OCR & text image \\
FER2013~\citep{fer2013} & 7 & fac. exp. rec.  &  natural image\\
CLEVR-C~\citep{johnson2017clevr} & 8 & object counting & natural image \\
CLEVR-D~\citep{johnson2017clevr}& 8 & distance est.  & natural image \\
DMLab~\citep{DMLab} & 6 &  distance est. & synthetic \\
KITTI~\citep{Geiger2013IJRR} & 4 & distance est. & natural image \\
\bottomrule
\end{tabular}}

\end{table}

%% file: tables/dataset_nameing.tex
\begin{table}
\caption{\textbf{Translation of open clip to model/pre-training dataset names used in paper.} When renaming the datasets we tried to group models with similar optimization schemes to minimize the number of pre-training datasets without causing undo overlap.
\label{tab:modeldataset}}
\centering
\resizebox{1\textwidth}{!}{
\begin{tabular}{c|cc|cc}
\toprule
ID & Model & Name & Dataset & Name \\  
\midrule
1 & RN50 & RN50 & openai & WIT \\ 
2 & RN101 & RN101 & openai & WIT \\ 
3 & RN50x4 & RN50x4 & openai & WIT \\ 
4 & RN50-16 & RN50x16 & openai & WIT \\ 
5 & RN50x64 & RN50x64 & openai & WIT \\ 
6 & ViT-B-32 & ViT-B/32 & laion400m\_e31 & L400m \\ 
7 & ViT-B-32 & ViT-B/32 & laion400m\_e32 & L400m \\ 
8 & ViT-B-32-quickgelu & ViT-B/32 & laion400m\_e32 & L400m \\ 
9 & ViT-B-32 & ViT-B/32 & openai & WIT \\ 
10 & ViT-B-32 & ViT-B/32 & laion2b\_s34b\_b79k & L2b-b \\ 
11 & ViT-B-32 & ViT-B/32 & laion2b\_e16 &  L2b-c\\ 
12 & ViT-B-16 & ViT-B/16 & laion400m\_e32 & L400m \\ 
13 & ViT-B-16 & ViT-B/16 & openai & WIT \\ 
14 & ViT-B-16-240 & ViT-B/16-240 & laion400m\_e32 & L400m \\ 
15 & ViT-L-14 & ViT-L/14 & laion400m\_e31 & L400m \\ 
16 & ViT-L-14 & ViT-L/14 & laion400m\_e32 & L400m \\ 
17 & ViT-L-14 & ViT-L/14 & laion2b\_s32b\_b82k & L2b-b \\ 
18 & ViT-L-14 & ViT-L/14 & openai & WIT \\ 
19 & ViT-L-14-336 & ViT-L/14-336 & openai & WIT \\ 
20 & ViT-G-14 & ViT-G/14 & laion2b\_s12b\_b42k & L2b-a \\ 
21 & ViT-G-14 & ViT-G/14 & laion2b\_s34b\_b88k & L2b-a \\ 
22 & ViT-H-14 & ViT-H/14 & laion2b\_s32b\_b79k & L2b-b \\ 
23 & coca\_ViT-B-32 & CoCa-ViT-B/32 & laion2b\_s13b\_b90k  & L2b-c\\
24 & coca\_ViT-B-32 & CoCa-ViT-B/32 & mscoco\_finetuned\_laion2b\_s13b\_b90k & L2b-c + coco\\
25 & coca\_ViT-L-14 & CoCa-ViT-L/14 & laion2b\_s13b\_b90k & L2b-c\\
26 & coca\_ViT-L-14 & CoCa-ViT-L/14 & mscoco\_finetuned\_laion2b\_s13b\_b90k & L2b-c + coco\\

27 & convnext\_base & ConvNEXT-B & laion400m\_s13b\_b51k & L400m-c\\
28 & convnext\_base\_w & ConvNEXT-BW & laion2b\_s13b\_b82k &  L2b-d\\
29 & convnext\_base\_w & ConvNEXT-BW & laion2b\_s13b\_b82k\_augreg & L2b-e\\

30 & convnext\_base\_w & ConvNEXT-BW & laion\_aesthetic\_s13b\_b82k & L2b-f\\
31 & convnext\_base\_w\_320 & ConvNEXT-BW-320 & laion\_aesthetic\_s13b\_b82k & L2b-f\\
32 & convnext\_base\_w\_320 & ConvNEXT-BW-320 & laion\_aesthetic\_s13b\_b82k\_augreg & L2b-g\\
33 & convnext\_large\_d & ConvNEXT-LD & laion2b\_s26b\_b102k\_augreg  &  L2b-h\\
34 & convnext\_large\_d\_320 & ConvNEXT-LD-320 & laion2b\_s29b\_b131k\_ft  & L2b-i\\
35 & convnext\_large\_d\_320 & ConvNEXT-LD-320 & laion2b\_s29b\_b131k\_ft\_soup  &  L2b-j\\
    \bottomrule
\end{tabular}}

\end{table}

%% file: sections/sup_method.tex
\begin{figure*}
\begin{center}
\fbox{
\includegraphics[width=0.95\textwidth]{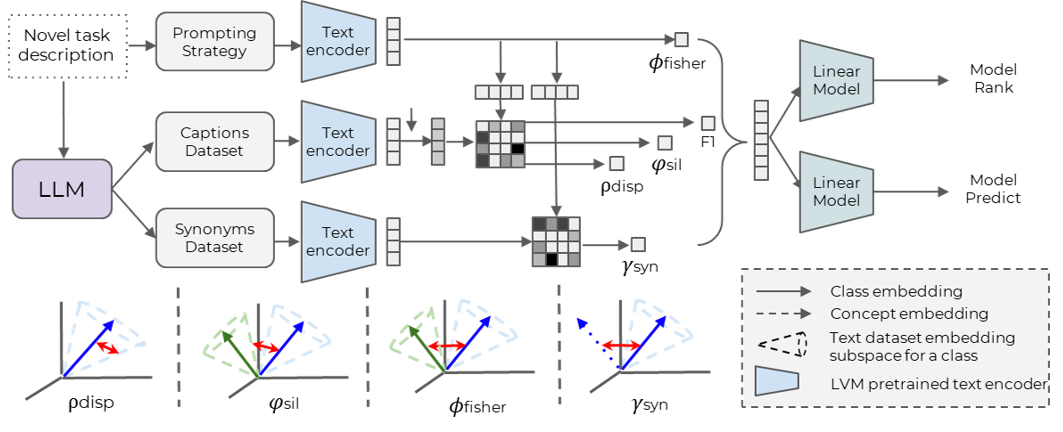}
}
\end{center}
   \caption{\textbf{Overview of the baselines}. (top left) Using a text description of a new task, we use a large language model to generate the Image Caption and Class-Synonym datasets. We feed these text datasets into a VLMs text encoder, which generates the text-derived multi-modal embeddings. Using these embeddings, as well as the user-defined prompting strategies, we extract different scores. Finally, we fit a linear model on the extracted scores to predict model ranking and accuracy. (bottom) Schematic drawings of our different proposed scores.} 
\label{sup:fig:method}
\end{figure*}

Fig.~\ref{sup:fig:method} shows an overview of our baselines. We first describe the prompting protocol used in Sec~\ref{sup:sec:benchmark_details:prompt}. We then give an in-detail description of the scores used in the study in Sec.~\ref{sup:sec:meth:features}.
In Sec.~\ref{sup:sec:meth:datagen}, we give additional details for the text dataset generation. Finally, in Sec.~\ref{sup:sec:meth:acc1_noise_corruption}, we describe how we use noise to corrupt the text caption dataset when calculating text-acc1 and text-f1 scores.

\subsection{Prompting Templates\label{sup:sec:benchmark_details:prompt}}

We use the same prompting strategy introduced by \citet{CLIP} to generate the model zero-shot weights (see Fig.~\ref{sup:fig:templates} for examples of templates from different datasets). Specifically, for every class $c$, we used the reported templates to produce the text prompts $\bm{Y^c}$ and encoded these prompts using the VLM text encoder, $f_y$, to produce the text embeddings for class $c$, $\hat{\bm{y}}^c$:
\begin{equation}
    \bm{\hat{y}}^c = f_y(\bm{Y}^c).\nonumber
\end{equation}
We then normalized each separate prompt by its $L_2$ norm and averaged the resulting vector to produce $\bm{\bar{y}^c}$, or the unnormalized zero-shot weight of class $c$:
\begin{align}
\bar{\bm{y}}^c&=\frac{1}{N}\sum_{\text{j=1}}^{\text{N}}\frac{\bm{y}_j^c}{||\bm{y}_j^c||_2},\nonumber
\end{align}
where $\bar{\bm{y}}^c$ is then normalized again to produce the final zero-shot classification weight of class $c$,
\begin{align}
\bm{y}^c&=\frac{\bar{\bm{y}}^c}{||\bar{\bm{y}}^c||_2}.
\end{align}

\subsection{Text-Derived Scores}
\label{sup:sec:meth:features}

We define the six scores we derived for model selection and performance prediction. The \textit{Text top-1 accuracy score} and  \textit{Text f1-score} is used to estimate the VLMs' performance on a vision task using text as a proxy, while the \textit{\intraClassClose} and \textit{\intraSim} are used to understand the VLM's capability to separate samples from different classes in the target task (inter-class similarity. To estimate dataset granularity, we use \textit{\interSim}.
Finally, the \textit{\superclass} allows us to evaluate the degree of content shift between the VLMs' pre-training and target dataset (intra-class similarity). 

\paragraph{Text Classification scores.} We use the generated captions dataset (see Sec.~\ref{sec:meth:gen_datasets}) as a proxy for images and evaluate the resulting model performance. Specifically, we use the VLM text encoder to generate text-derived multi-modal embeddings. We then corrupt these embeddings with Gaussian noise to approximate image-instance variation (see Sec.~\ref{sup:sec:meth:acc1_noise_corruption})and calculate their cosine similarity with the class prompt embeddings - derived using the same prompt ensembling strategies proposed by~\cite{CLIP} (see Fig.~\ref{sup:fig:method}). We then calculate the text top-1 accuracy (\textbf{text-acc1}) and text f1-score (\textbf{text-f1}).

\paragraph{\intraClassClose,} \intraClassCloseSymbl\textbf{.} The Fisher criterion~\citep{fisher} has been widely used as a dataset granularity measure and has recently been shown to be effective for classification by ~\citet{gran}. The Fisher score measures the degree of similarity of the classes or the extent of their separation. 
In VLMs, The quality of the class separation can be evaluated using text by assessing how close the different (text-derived) class prompt embeddings are. 
We introduce the concept of \textit{\intraClassClose}, a score that quantifies how close the class prompt embeddings are to each other (see Fig.~\ref{sup:fig:method}):
\begin{equation}
\phi_{\text{fisher}} = \frac{1}{C}\sum_{j=1}^{C}\text{max}_{c,c \neq j}\left[\theta(\bm{y}^{j},\bm{y}^{c})\right],
\end{equation}
where $\bm{y}^c$ is the class prompt embedding derived using the prompt ensembling strategies proposed in~\cite{CLIP} for class $c$ (see Sec.~\ref{sup:sec:benchmark_details:prompt}), $\theta(\cdot,\cdot)$ is a function that calculates the cosine similarity between two vectors, and $C$ is the number of classes.

\paragraph{\intraSim,} \intraSimSymbl\textbf{.} The silhouette score~\citep{silBase} is a well-established score that has been used to quantify the tightness of the same-class samples to the separation of different-class samples~\citep{silClass, gran}. Inspired by this score, we introduce the text-based \textit{\intraSim}, \intraSimSymbl, which measures the separation of different-class samples in the caption dataset $\bm{D^{\text{cap}}}$. To do so, we evaluate the average cosine similarity of captions to the nearest other class by: 
\begin{equation}
\varphi_\text{sil} = \frac{1}{C}
    \sum_{j=1}^C \text{max}_{c, c \neq j}\left[\frac{1}{N}\sum_{k=1}^N \theta(\bm{D^{\text{cap}}}[j]_{k}, \bm{y}^{c})\right],
\end{equation}
where $\bm{y}^c$ is the class prompt embedding derived using the prompt ensembling strategies proposed in~\cite{CLIP} for class $c$ (see Sec.~\ref{sup:sec:benchmark_details:prompt}), $\theta(\cdot,\cdot)$ is a function that calculates the cosine similarity between two vectors, and $C$ is the number of classes. $\bm{D^{\text{cap}}}[j]_{k}$ representing sample $k$ of class $j$ in the caption dataset $\bm{D^{\text{cap}}}$, and there is a total of $N$ such samples in for each class.

.

\paragraph{\interSim,} \interSimSymbl\textbf{.} The \textit{\interSim} is used as the normalization constant to generate the Fisher and Silhouette scores, and it quantifies the degree of same-class tightness or data cone radius (see Fig.~\ref{sup:fig:method}). 
\begin{equation}
    \rho_{\text{disp}} = \frac{1}{CN}
    \sum_{c=1}^C \sum_{k=1}^N \theta(\bm{D^{\text{cap}}}[c]_{k}, \bm{y}^{c}),
\end{equation}
where $\bm{y}^c$ is the class prompt embedding derived using the prompt ensembling strategies proposed in~\citet{CLIP} for class $c$ (see Sec.~\ref{sup:sec:benchmark_details:prompt}), $\theta(\cdot,\cdot)$ is a function that calculates the cosine similarity between two vectors, and $C$ is the number of classes. $\bm{D^{\text{cap}}}[c]_{k}$ representing sample $k$ of class $c$ in the caption dataset $\bm{D^{\text{cap}}}$, and there is a total of $N$ such samples in for each class.

 \paragraph{\superclass,} \superclassSymbl\textbf{.} Synonym consistency has been shown in large language models to correlate with the degree of familiarity of the model with a particular concept~\citep{syn_metric1}. Using the Synonym dataset, we compare the cosine similarity between the text embedding of each class and its corresponding synonyms. A high cosine similarity between the class and its corresponding synonyms/supercategories indicates that the model is aware of the semantic meaning of the class and is defined as: 
\begin{equation}
\gamma_{\text{syn}} = \frac{1}{CN}
    \sum_{c=1}^C \sum_{k=1}^N \theta(\bm{D^{\text{syn}}}[c]_{k}, \bm{y}^{c}),
\end{equation}
where $\bm{y}^c$ is the class prompt embedding derived using the prompt ensembling strategies proposed in~\cite{CLIP} for class $c$ (see Sec.~\ref{sup:sec:benchmark_details:prompt}), $\theta(\cdot,\cdot)$ is a function that calculates the cosine similarity between two vectors, and $C$ is the number of classes. $\bm{D^{\text{syn}}}[c]_{k}$ representing sample $k$ of class $c$ in the synonym dataset $\bm{D^{\text{syn}}}$, and there is a total of $N$ such samples in for each class.

\subsection{Text Dataset Generation\label{sup:sec:meth:datagen}}
To generate the Captions dataset, we used a large language model to generate realistic (but confusing) image captions. It was necessary to request confusing image captions to get sufficient variation in the image captions. We used OpenAI's `gpt-3.5-turbo-0301' model with a temperature of 1. 
For the synonym dataset, we reduced the temperature to 0.1 and only requested the synonyms themselves. We then used the prompting templates with the synonym in place of the original class name to generate $\bm{D^{\text{syn}}}$.

\subsection{Text Classification and Noise Corruption\label{sup:sec:meth:acc1_noise_corruption}}

In this work, we introduce the use of Gaussian noise to corrupt text-derived multi-modal embeddings to approximate image-instance variation. The corrupted embeddings are then used to calculate the text top-1 accuracy and text-f1 score, which serves as a proxy for evaluating the performance of a vision model. 
The scores are derived from the Captions dataset, a collection of complex but probable image captions generated using a large language model for images containing the user-provided classes in the user-provided domain and for the user-provided task. For more, see Sec.~\ref{sec:meth:gen_datasets}.

\begin{figure*}[t]
\begin{center}
  \resizebox{1\textwidth}{!}{
\includegraphics[width=1\textwidth]{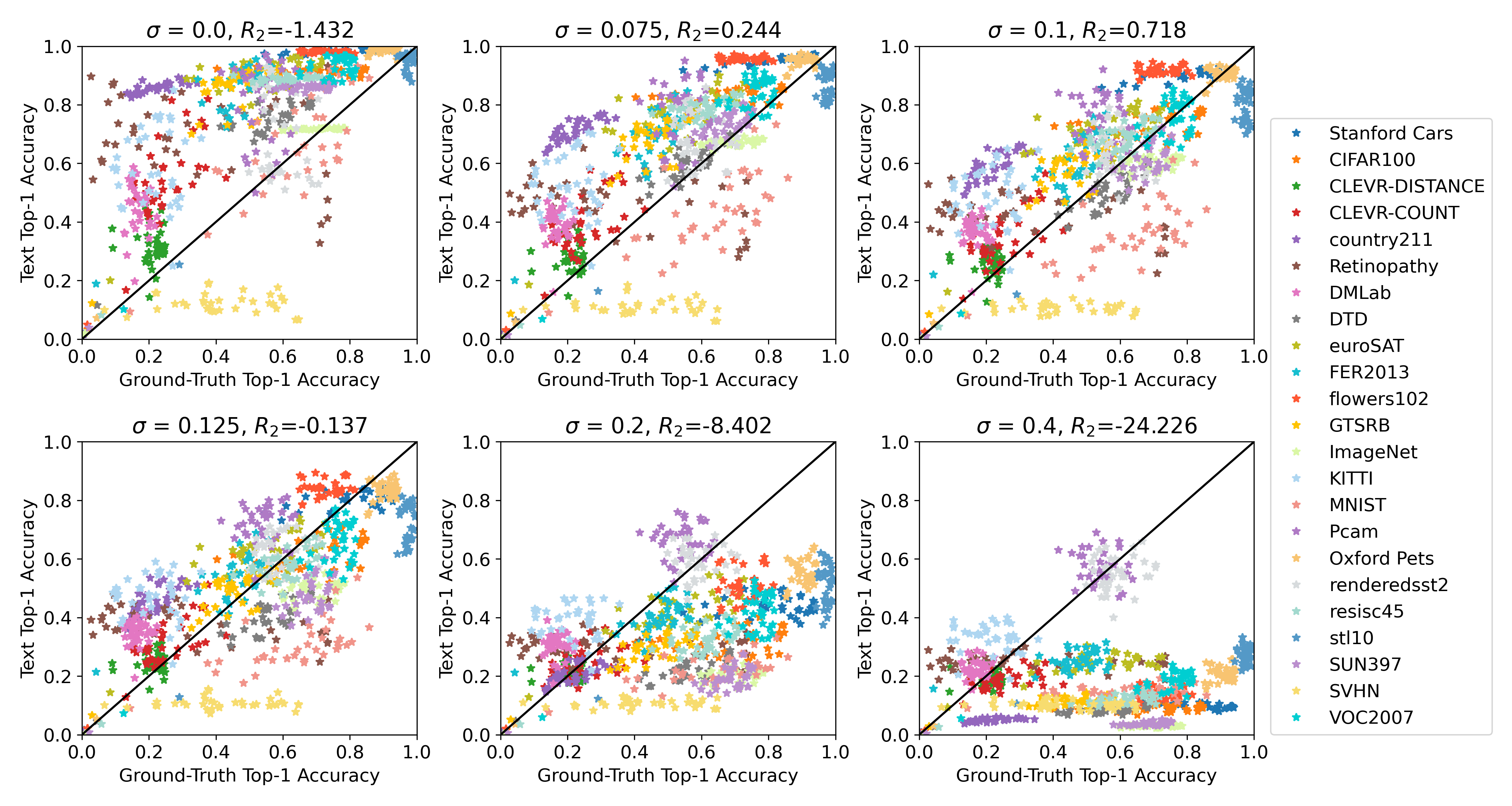}
}
\vspace{-0.3in}
\end{center}
       \caption{\textbf{Ablating Noise Injection Effect on Text Top-1 Accuracy}. Without noise (sigma=0), the text top-1 accuracy saturates on many datasets and models, with extremely high top-1 accuracy, making the correlation between the ground-truth top-1 accuracy and text top-1 accuracy quite poor. By corrupting the text embeddings with noise, we notice an improvement in correlation up to sigma=0.1, after which the correlation is steadily corrupted. 
   }
   \label{fig:AddingNoise}
\end{figure*}

To evaluate the effectiveness of the text top-1 accuracy, we systematically increase the level of noise corruption and plot the text top-1 accuracy against the ground-truth top-1 accuracy  (See Fig.~\ref{fig:AddingNoise}). We quantify this correlation via the $R^2$ score or the degree of explained variance. As we do not fit a linear model to the predicted vs. ground-truth predictions, $R^2$ ranges from 1 (perfect linear fit) to $-\infty$, where non of the variance is explained. 
The results show that, without noise corruption, the text top-1 accuracy is too high and frequently saturates without any corruption.
However, as the noise level increases to 0.1, the text top-1 accuracy progressively improves until a better linear correlation can be seen. 
This indicates that increasing noise corruption can better approximate image-instance variation and improve the correlation of the text top-1 accuracy. Beyond 0.1, however, the correlation between the text top-1 accuracy and  top-1 accuracy progressively worsens. This shows that while noise corruption helps improve the correlation of the text top-1 accuracy, there is a limit beyond which further noise corruption degrades its effectiveness.

%% file: tables/lovm_acc.tex
\begin{table*}[t]
\caption{\textbf{LOVM Benchmark (top-1 accuracy)}.We evaluate our method's performance over \numdatasets\ datasets and \nummodels\ pre-trained models.
(top) Model Ranking results. (bottom) Performance Prediction results. INB - ImageNet Benchmark score, C - Text Classification scores G - Granularity Scores.
}

\vspace{0.8em}
\adjustbox{width=1\textwidth}{
\begin{tabular}{cc|ccccccccccccccccccccccc|c}

    & 
    & \rotatebox[origin=lb]{90}{\smash{Stanford Cars}} 
    & \rotatebox[origin=lb]{90}{\smash{CIFAR100}} 
    & \rotatebox[origin=lb]{90}{\smash{CLEVR-DIST.}} 
    & \rotatebox[origin=lb]{90}{\smash{CLEVR-COUNT}} 
    & \rotatebox[origin=lb]{90}{\smash{Country211}} 
    & \rotatebox[origin=lb]{90}{\smash{Retinopathy}} 
    & \rotatebox[origin=lb]{90}{\smash{DMLab}} 
    & \rotatebox[origin=lb]{90}{\smash{DTD}} 
    & \rotatebox[origin=lb]{90}{\smash{EuroSAT}} 
    & \rotatebox[origin=lb]{90}{\smash{FER2013}} 
    & \rotatebox[origin=lb]{90}{\smash{Flowers102}} 
    & \rotatebox[origin=lb]{90}{\smash{GTSRB}} 
    & \rotatebox[origin=lb]{90}{\smash{ImageNet}} 
    & \rotatebox[origin=lb]{90}{\smash{KITTI}} 
    & \rotatebox[origin=lb]{90}{\smash{MNIST}} 
    & \rotatebox[origin=lb]{90}{\smash{PCam}}
    & \rotatebox[origin=lb]{90}{\smash{Oxford Pets}} 
    &  \rotatebox[origin=lb]{90}{\smash{Rendered SST2}}
    & \rotatebox[origin=lb]{90}{\smash{RESISC45}}
    & \rotatebox[origin=lb]{90}{\smash{STL10}} 
    & \rotatebox[origin=lb]{90}{\smash{SUN397}} 
    & \rotatebox[origin=lb]{90}{\smash{SVHN}} 
    & \rotatebox[origin=lb]{90}{\smash{VOC2007}} 
    %& \rotatebox[origin=lb]{90}{\smash{UCF101}} 
    %& \rotatebox[origin=lb]{90}{\smash{Kinetics700}} 
& \cellcolor[HTML]{FFFFED} \rotatebox[origin=lb]{90}{\smash{Mean}} 
    \\
    \midrule
    
\multirow{7}{*}{\rotatebox[origin=c]{90}{$R_5$}}

\hspace{-0.4em} &  \hspace{-0.9em} INB &
0.80 \hspace{-0.4em} & \hspace{-0.9em} 0.80 \hspace{-0.4em} & \hspace{-0.9em} 0.00 \hspace{-0.4em} & \hspace{-0.9em} 0.60 \hspace{-0.4em} & \hspace{-0.9em} 0.40 \hspace{-0.4em} & \hspace{-0.9em} 0.00 \hspace{-0.4em} & \hspace{-0.9em} 0.00 \hspace{-0.4em} & \hspace{-0.9em} 1.00 \hspace{-0.4em} & \hspace{-0.9em} 0.60 \hspace{-0.4em} & \hspace{-0.9em} 0.20 \hspace{-0.4em} & \hspace{-0.9em} 0.40 \hspace{-0.4em} & \hspace{-0.9em} 0.60 \hspace{-0.4em} & \hspace{-0.9em} 1.00 \hspace{-0.4em} & \hspace{-0.9em} 0.00 \hspace{-0.4em} & \hspace{-0.9em} 0.20 \hspace{-0.4em} & \hspace{-0.9em} 0.00 \hspace{-0.4em} & \hspace{-0.9em} 0.80 \hspace{-0.4em} & \hspace{-0.9em} 0.00 \hspace{-0.4em} & \hspace{-0.9em} 1.00 \hspace{-0.4em} & \hspace{-0.9em} 0.20 \hspace{-0.4em} & \hspace{-0.9em} 0.60 \hspace{-0.4em} & \hspace{-0.9em} 0.60 \hspace{-0.4em} & \hspace{-0.9em} 0.60 \hspace{-0.4em} &  \cellcolor[HTML]{FFFFED} \hspace{-0.9em} 0.452      \\

\hspace{-0.4em} &  \hspace{-0.9em} C &
0.00 \hspace{-0.4em} & \hspace{-0.9em} 0.20 \hspace{-0.4em} & \hspace{-0.9em} 0.20 \hspace{-0.4em} & \hspace{-0.9em} 0.60 \hspace{-0.4em} & \hspace{-0.9em} 0.40 \hspace{-0.4em} & \hspace{-0.9em} 0.00 \hspace{-0.4em} & \hspace{-0.9em} 0.00 \hspace{-0.4em} & \hspace{-0.9em} 0.60 \hspace{-0.4em} & \hspace{-0.9em} 0.40 \hspace{-0.4em} & \hspace{-0.9em} 0.40 \hspace{-0.4em} & \hspace{-0.9em} 0.20 \hspace{-0.4em} & \hspace{-0.9em} 0.40 \hspace{-0.4em} & \hspace{-0.9em} 0.00 \hspace{-0.4em} & \hspace{-0.9em} 0.20 \hspace{-0.4em} & \hspace{-0.9em} 0.20 \hspace{-0.4em} & \hspace{-0.9em} 0.20 \hspace{-0.4em} & \hspace{-0.9em} 0.00 \hspace{-0.4em} & \hspace{-0.9em} 0.20 \hspace{-0.4em} & \hspace{-0.9em} 0.40 \hspace{-0.4em} & \hspace{-0.9em} 0.20 \hspace{-0.4em} & \hspace{-0.9em} 0.40 \hspace{-0.4em} & \hspace{-0.9em} 0.00 \hspace{-0.4em} & \hspace{-0.9em} 0.00 \hspace{-0.4em} &  \cellcolor[HTML]{FFFFED} \hspace{-0.9em} 0.226       \\

\hspace{-0.4em} &  \hspace{-0.9em} G &
0.40 \hspace{-0.4em} & \hspace{-0.9em} 0.40 \hspace{-0.4em} & \hspace{-0.9em} 0.20 \hspace{-0.4em} & \hspace{-0.9em} 0.20 \hspace{-0.4em} & \hspace{-0.9em} 0.40 \hspace{-0.4em} & \hspace{-0.9em} 0.00 \hspace{-0.4em} & \hspace{-0.9em} 0.00 \hspace{-0.4em} & \hspace{-0.9em} 0.40 \hspace{-0.4em} & \hspace{-0.9em} 0.20 \hspace{-0.4em} & \hspace{-0.9em} 0.20 \hspace{-0.4em} & \hspace{-0.9em} 0.80 \hspace{-0.4em} & \hspace{-0.9em} 0.20 \hspace{-0.4em} & \hspace{-0.9em} 0.40 \hspace{-0.4em} & \hspace{-0.9em} 0.20 \hspace{-0.4em} & \hspace{-0.9em} 0.00 \hspace{-0.4em} & \hspace{-0.9em} 0.20 \hspace{-0.4em} & \hspace{-0.9em} 0.40 \hspace{-0.4em} & \hspace{-0.9em} 0.00 \hspace{-0.4em} & \hspace{-0.9em} 0.40 \hspace{-0.4em} & \hspace{-0.9em} 0.20 \hspace{-0.4em} & \hspace{-0.9em} 0.40 \hspace{-0.4em} & \hspace{-0.9em} 0.00 \hspace{-0.4em} & \hspace{-0.9em} 0.20 \hspace{-0.4em} &  \cellcolor[HTML]{FFFFED} \hspace{-0.9em} 0.252      \\ 

\hspace{-0.4em} &  \hspace{-0.9em} G+C &
0.40 \hspace{-0.4em} & \hspace{-0.9em} 0.40 \hspace{-0.4em} & \hspace{-0.9em} 0.20 \hspace{-0.4em} & \hspace{-0.9em} 0.20 \hspace{-0.4em} & \hspace{-0.9em} 0.40 \hspace{-0.4em} & \hspace{-0.9em} 0.00 \hspace{-0.4em} & \hspace{-0.9em} 0.00 \hspace{-0.4em} & \hspace{-0.9em} 0.40 \hspace{-0.4em} & \hspace{-0.9em} 0.20 \hspace{-0.4em} & \hspace{-0.9em} 0.20 \hspace{-0.4em} & \hspace{-0.9em} 0.80 \hspace{-0.4em} & \hspace{-0.9em} 0.20 \hspace{-0.4em} & \hspace{-0.9em} 0.40 \hspace{-0.4em} & \hspace{-0.9em} 0.20 \hspace{-0.4em} & \hspace{-0.9em} 0.00 \hspace{-0.4em} & \hspace{-0.9em} 0.20 \hspace{-0.4em} & \hspace{-0.9em} 0.40 \hspace{-0.4em} & \hspace{-0.9em} 0.00 \hspace{-0.4em} & \hspace{-0.9em} 0.40 \hspace{-0.4em} & \hspace{-0.9em} 0.20 \hspace{-0.4em} & \hspace{-0.9em} 0.40 \hspace{-0.4em} & \hspace{-0.9em} 0.00 \hspace{-0.4em} & \hspace{-0.9em} 0.20 \hspace{-0.4em} &  \cellcolor[HTML]{FFFFED} \hspace{-0.9em} 0.252   \\

\hspace{-0.4em} &  \hspace{-0.9em} INB+C &
0.80 \hspace{-0.4em} & \hspace{-0.9em} 0.80 \hspace{-0.4em} & \hspace{-0.9em} 0.00 \hspace{-0.4em} & \hspace{-0.9em} 0.60 \hspace{-0.4em} & \hspace{-0.9em} 0.60 \hspace{-0.4em} & \hspace{-0.9em} 0.00 \hspace{-0.4em} & \hspace{-0.9em} 0.00 \hspace{-0.4em} & \hspace{-0.9em} 1.00 \hspace{-0.4em} & \hspace{-0.9em} 0.60 \hspace{-0.4em} & \hspace{-0.9em} 0.20 \hspace{-0.4em} & \hspace{-0.9em} 0.60 \hspace{-0.4em} & \hspace{-0.9em} 0.60 \hspace{-0.4em} & \hspace{-0.9em} 0.80 \hspace{-0.4em} & \hspace{-0.9em} 0.00 \hspace{-0.4em} & \hspace{-0.9em} 0.20 \hspace{-0.4em} & \hspace{-0.9em} 0.20 \hspace{-0.4em} & \hspace{-0.9em} 0.80 \hspace{-0.4em} & \hspace{-0.9em} 0.00 \hspace{-0.4em} & \hspace{-0.9em} 1.00 \hspace{-0.4em} & \hspace{-0.9em} 0.20 \hspace{-0.4em} & \hspace{-0.9em} 0.60 \hspace{-0.4em} & \hspace{-0.9em} 0.40 \hspace{-0.4em} & \hspace{-0.9em} 0.60 \hspace{-0.4em} &  \cellcolor[HTML]{FFFFED} \hspace{-0.9em} \textbf{0.461} \\

\hspace{-0.4em} &  \hspace{-0.9em} INB+G &
0.80 \hspace{-0.4em} & \hspace{-0.9em} 0.80 \hspace{-0.4em} & \hspace{-0.9em} 0.00 \hspace{-0.4em} & \hspace{-0.9em} 0.60 \hspace{-0.4em} & \hspace{-0.9em} 0.40 \hspace{-0.4em} & \hspace{-0.9em} 0.00 \hspace{-0.4em} & \hspace{-0.9em} 0.00 \hspace{-0.4em} & \hspace{-0.9em} 1.00 \hspace{-0.4em} & \hspace{-0.9em} 0.60 \hspace{-0.4em} & \hspace{-0.9em} 0.20 \hspace{-0.4em} & \hspace{-0.9em} 0.40 \hspace{-0.4em} & \hspace{-0.9em} 0.60 \hspace{-0.4em} & \hspace{-0.9em} 1.00 \hspace{-0.4em} & \hspace{-0.9em} 0.00 \hspace{-0.4em} & \hspace{-0.9em} 0.40 \hspace{-0.4em} & \hspace{-0.9em} 0.00 \hspace{-0.4em} & \hspace{-0.9em} 0.80 \hspace{-0.4em} & \hspace{-0.9em} 0.20 \hspace{-0.4em} & \hspace{-0.9em} 1.00 \hspace{-0.4em} & \hspace{-0.9em} 0.40 \hspace{-0.4em} & \hspace{-0.9em} 0.60 \hspace{-0.4em} & \hspace{-0.9em} 0.40 \hspace{-0.4em} & \hspace{-0.9em} 0.40 \hspace{-0.4em} & \cellcolor[HTML]{FFFFED} \hspace{-0.9em} \textbf{0.461}   \\

\hspace{-0.4em} &  \hspace{-0.9em} INB+C+G &
0.80 \hspace{-0.4em} & \hspace{-0.9em} 0.80 \hspace{-0.4em} & \hspace{-0.9em} 0.00 \hspace{-0.4em} & \hspace{-0.9em} 0.60 \hspace{-0.4em} & \hspace{-0.9em} 0.40 \hspace{-0.4em} & \hspace{-0.9em} 0.00 \hspace{-0.4em} & \hspace{-0.9em} 0.00 \hspace{-0.4em} & \hspace{-0.9em} 1.00 \hspace{-0.4em} & \hspace{-0.9em} 0.60 \hspace{-0.4em} & \hspace{-0.9em} 0.20 \hspace{-0.4em} & \hspace{-0.9em} 0.40 \hspace{-0.4em} & \hspace{-0.9em} 0.60 \hspace{-0.4em} & \hspace{-0.9em} 1.00 \hspace{-0.4em} & \hspace{-0.9em} 0.00 \hspace{-0.4em} & \hspace{-0.9em} 0.40 \hspace{-0.4em} & \hspace{-0.9em} 0.00 \hspace{-0.4em} & \hspace{-0.9em} 0.80 \hspace{-0.4em} & \hspace{-0.9em} 0.20 \hspace{-0.4em} & \hspace{-0.9em} 1.00 \hspace{-0.4em} & \hspace{-0.9em} 0.40 \hspace{-0.4em} & \hspace{-0.9em} 0.60 \hspace{-0.4em} & \hspace{-0.9em} 0.40 \hspace{-0.4em} & \hspace{-0.9em} 0.40 \hspace{-0.4em} &  \cellcolor[HTML]{FFFFED} \hspace{-0.9em} \textbf{0.461}  \\

\midrule

%%%%%%%%%%%%%%%%%%%%%%%%%%%

\multirow{7}{*}{\rotatebox[origin=c]{90}{$\tau$}}

\hspace{-0.4em} &  \hspace{-0.9em} INB&
0.33 \hspace{-0.4em} & \hspace{-0.9em} 0.67 \hspace{-0.4em} & \hspace{-0.9em} 0.00 \hspace{-0.4em} & \hspace{-0.9em} 0.33 \hspace{-0.4em} & \hspace{-0.9em} 0.00 \hspace{-0.4em} & \hspace{-0.9em} 0.00 \hspace{-0.4em} & \hspace{-0.9em} 0.00 \hspace{-0.4em} & \hspace{-0.9em} -0.20 \hspace{-0.4em} & \hspace{-0.9em} 1.00 \hspace{-0.4em} & \hspace{-0.9em} 0.00 \hspace{-0.4em} & \hspace{-0.9em} 0.00 \hspace{-0.4em} & \hspace{-0.9em} 1.00 \hspace{-0.4em} & \hspace{-0.9em} 1.00 \hspace{-0.4em} & \hspace{-0.9em} 0.00 \hspace{-0.4em} & \hspace{-0.9em} 0.00 \hspace{-0.4em} & \hspace{-0.9em} 0.00 \hspace{-0.4em} & \hspace{-0.9em} 0.00 \hspace{-0.4em} & \hspace{-0.9em} 0.00 \hspace{-0.4em} & \hspace{-0.9em} -0.40 \hspace{-0.4em} & \hspace{-0.9em} 0.00 \hspace{-0.4em} & \hspace{-0.9em} -1.00 \hspace{-0.4em} & \hspace{-0.9em} 0.33 \hspace{-0.4em} & \hspace{-0.9em} 1.00 \hspace{-0.4em} &  \cellcolor[HTML]{FFFFED} \hspace{-0.9em} 0.177    \\

\hspace{-0.4em} & \hspace{-0.9em} C &
0.00 \hspace{-0.4em} & \hspace{-0.9em} 0.00 \hspace{-0.4em} & \hspace{-0.9em} 0.00 \hspace{-0.4em} & \hspace{-0.9em} 1.00 \hspace{-0.4em} & \hspace{-0.9em} 0.00 \hspace{-0.4em} & \hspace{-0.9em} 0.00 \hspace{-0.4em} & \hspace{-0.9em} 0.00 \hspace{-0.4em} & \hspace{-0.9em} 0.33 \hspace{-0.4em} & \hspace{-0.9em} 0.00 \hspace{-0.4em} & \hspace{-0.9em} 0.00 \hspace{-0.4em} & \hspace{-0.9em} 0.00 \hspace{-0.4em} & \hspace{-0.9em} 0.00 \hspace{-0.4em} & \hspace{-0.9em} 0.00 \hspace{-0.4em} & \hspace{-0.9em} 0.00 \hspace{-0.4em} & \hspace{-0.9em} 0.00 \hspace{-0.4em} & \hspace{-0.9em} 0.00 \hspace{-0.4em} & \hspace{-0.9em} 0.00 \hspace{-0.4em} & \hspace{-0.9em} 0.00 \hspace{-0.4em} & \hspace{-0.9em} 0.00 \hspace{-0.4em} & \hspace{-0.9em} 0.00 \hspace{-0.4em} & \hspace{-0.9em} 0.00 \hspace{-0.4em} & \hspace{-0.9em} 0.00 \hspace{-0.4em} & \hspace{-0.9em} 0.00 \hspace{-0.4em} &  \cellcolor[HTML]{FFFFED} \hspace{-0.9em} 0.058    \\

\hspace{-0.4em} & \hspace{-0.9em} G &
0.00 \hspace{-0.4em} & \hspace{-0.9em} 0.00 \hspace{-0.4em} & \hspace{-0.9em} 0.00 \hspace{-0.4em} & \hspace{-0.9em} 0.00 \hspace{-0.4em} & \hspace{-0.9em} 0.00 \hspace{-0.4em} & \hspace{-0.9em} 0.00 \hspace{-0.4em} & \hspace{-0.9em} 0.00 \hspace{-0.4em} & \hspace{-0.9em} 0.00 \hspace{-0.4em} & \hspace{-0.9em} 0.00 \hspace{-0.4em} & \hspace{-0.9em} 0.00 \hspace{-0.4em} & \hspace{-0.9em} -0.33 \hspace{-0.4em} & \hspace{-0.9em} 0.00 \hspace{-0.4em} & \hspace{-0.9em} 0.00 \hspace{-0.4em} & \hspace{-0.9em} 0.00 \hspace{-0.4em} & \hspace{-0.9em} 0.00 \hspace{-0.4em} & \hspace{-0.9em} 0.00 \hspace{-0.4em} & \hspace{-0.9em} 0.00 \hspace{-0.4em} & \hspace{-0.9em} 0.00 \hspace{-0.4em} & \hspace{-0.9em} 0.00 \hspace{-0.4em} & \hspace{-0.9em} 0.00 \hspace{-0.4em} & \hspace{-0.9em} 0.00 \hspace{-0.4em} & \hspace{-0.9em} 0.00 \hspace{-0.4em} & \hspace{-0.9em} 0.00 \hspace{-0.4em} &  \cellcolor[HTML]{FFFFED}\hspace{-0.9em} -0.014            \\

\hspace{-0.4em} & \hspace{-0.9em} C+G &
0.00 \hspace{-0.4em} & \hspace{-0.9em} 0.00 \hspace{-0.4em} & \hspace{-0.9em} 0.00 \hspace{-0.4em} & \hspace{-0.9em} 0.00 \hspace{-0.4em} & \hspace{-0.9em} 0.00 \hspace{-0.4em} & \hspace{-0.9em} 0.00 \hspace{-0.4em} & \hspace{-0.9em} 0.00 \hspace{-0.4em} & \hspace{-0.9em} 0.00 \hspace{-0.4em} & \hspace{-0.9em} 0.00 \hspace{-0.4em} & \hspace{-0.9em} 0.00 \hspace{-0.4em} & \hspace{-0.9em} -0.33 \hspace{-0.4em} & \hspace{-0.9em} 0.00 \hspace{-0.4em} & \hspace{-0.9em} 0.00 \hspace{-0.4em} & \hspace{-0.9em} 0.00 \hspace{-0.4em} & \hspace{-0.9em} 0.00 \hspace{-0.4em} & \hspace{-0.9em} 0.00 \hspace{-0.4em} & \hspace{-0.9em} 0.00 \hspace{-0.4em} & \hspace{-0.9em} 0.00 \hspace{-0.4em} & \hspace{-0.9em} 0.00 \hspace{-0.4em} & \hspace{-0.9em} 0.00 \hspace{-0.4em} & \hspace{-0.9em} 0.00 \hspace{-0.4em} & \hspace{-0.9em} 0.00 \hspace{-0.4em} & \hspace{-0.9em} 0.00 \hspace{-0.4em} &  \cellcolor[HTML]{FFFFED}\hspace{-0.9em} -0.014         \\

\hspace{-0.4em} & \hspace{-0.9em} INB+C &
0.33 \hspace{-0.4em} & \hspace{-0.9em} 0.67 \hspace{-0.4em} & \hspace{-0.9em} 0.00 \hspace{-0.4em} & \hspace{-0.9em} 0.33 \hspace{-0.4em} & \hspace{-0.9em} -0.33 \hspace{-0.4em} & \hspace{-0.9em} 0.00 \hspace{-0.4em} & \hspace{-0.9em} 0.00 \hspace{-0.4em} & \hspace{-0.9em} 0.00 \hspace{-0.4em} & \hspace{-0.9em} 1.00 \hspace{-0.4em} & \hspace{-0.9em} 0.00 \hspace{-0.4em} & \hspace{-0.9em} 1.00 \hspace{-0.4em} & \hspace{-0.9em} 1.00 \hspace{-0.4em} & \hspace{-0.9em} 1.00 \hspace{-0.4em} & \hspace{-0.9em} 0.00 \hspace{-0.4em} & \hspace{-0.9em} 0.00 \hspace{-0.4em} & \hspace{-0.9em} 0.00 \hspace{-0.4em} & \hspace{-0.9em} 0.33 \hspace{-0.4em} & \hspace{-0.9em} 0.00 \hspace{-0.4em} & \hspace{-0.9em} -0.20 \hspace{-0.4em} & \hspace{-0.9em} 0.00 \hspace{-0.4em} & \hspace{-0.9em} -1.00 \hspace{-0.4em} & \hspace{-0.9em} 0.00 \hspace{-0.4em} & \hspace{-0.9em} 1.00 \hspace{-0.4em} &  \cellcolor[HTML]{FFFFED} \hspace{-0.9em} \textbf{0.223}   \\

\hspace{-0.4em} & \hspace{-0.9em} INB+G &
0.33 \hspace{-0.4em} & \hspace{-0.9em} 0.33 \hspace{-0.4em} & \hspace{-0.9em} 0.00 \hspace{-0.4em} & \hspace{-0.9em} 0.33 \hspace{-0.4em} & \hspace{-0.9em} 0.00 \hspace{-0.4em} & \hspace{-0.9em} 0.00 \hspace{-0.4em} & \hspace{-0.9em} 0.00 \hspace{-0.4em} & \hspace{-0.9em} 0.00 \hspace{-0.4em} & \hspace{-0.9em} 1.00 \hspace{-0.4em} & \hspace{-0.9em} 0.00 \hspace{-0.4em} & \hspace{-0.9em} 0.00 \hspace{-0.4em} & \hspace{-0.9em} 1.00 \hspace{-0.4em} & \hspace{-0.9em} 0.80 \hspace{-0.4em} & \hspace{-0.9em} 0.00 \hspace{-0.4em} & \hspace{-0.9em} 0.00 \hspace{-0.4em} & \hspace{-0.9em} 0.00 \hspace{-0.4em} & \hspace{-0.9em} 0.00 \hspace{-0.4em} & \hspace{-0.9em} 0.00 \hspace{-0.4em} & \hspace{-0.9em} -0.60 \hspace{-0.4em} & \hspace{-0.9em} 0.00 \hspace{-0.4em} & \hspace{-0.9em} -1.00 \hspace{-0.4em} & \hspace{-0.9em} 0.00 \hspace{-0.4em} & \hspace{-0.9em} 0.00 \hspace{-0.4em} &  \cellcolor[HTML]{FFFFED} \hspace{-0.9em} 0.096 \\

\hspace{-0.4em} & \hspace{-0.9em} INB+C+G &
0.33 \hspace{-0.4em} & \hspace{-0.9em} 0.67 \hspace{-0.4em} & \hspace{-0.9em} 0.00 \hspace{-0.4em} & \hspace{-0.9em} 0.33 \hspace{-0.4em} & \hspace{-0.9em} -0.33 \hspace{-0.4em} & \hspace{-0.9em} 0.00 \hspace{-0.4em} & \hspace{-0.9em} 0.00 \hspace{-0.4em} & \hspace{-0.9em} 0.00 \hspace{-0.4em} & \hspace{-0.9em} 1.00 \hspace{-0.4em} & \hspace{-0.9em} 0.00 \hspace{-0.4em} & \hspace{-0.9em} 1.00 \hspace{-0.4em} & \hspace{-0.9em} 1.00 \hspace{-0.4em} & \hspace{-0.9em} 1.00 \hspace{-0.4em} & \hspace{-0.9em} 0.00 \hspace{-0.4em} & \hspace{-0.9em} 0.00 \hspace{-0.4em} & \hspace{-0.9em} 0.00 \hspace{-0.4em} & \hspace{-0.9em} 0.33 \hspace{-0.4em} & \hspace{-0.9em} 0.00 \hspace{-0.4em} & \hspace{-0.9em} -0.20 \hspace{-0.4em} & \hspace{-0.9em} 0.00 \hspace{-0.4em} & \hspace{-0.9em} -1.00 \hspace{-0.4em} & \hspace{-0.9em} 0.00 \hspace{-0.4em} & \hspace{-0.9em} 1.00 \hspace{-0.4em} &  \cellcolor[HTML]{FFFFED} \hspace{-0.9em} \textbf{0.223} \\

\midrule 
\midrule 

%%%%%%%%%%%%%%%%%%%%%%%%%%%

\multirow{7}{*}{\rotatebox[origin=c]{90}{$L_1$}}
\hspace{-0.4em} & \hspace{-0.9em} INB&
0.16 \hspace{-0.4em} & \hspace{-0.9em} 0.07 \hspace{-0.4em} & \hspace{-0.9em} 0.58 \hspace{-0.4em} & \hspace{-0.9em} 0.42 \hspace{-0.4em} & \hspace{-0.9em} 0.49 \hspace{-0.4em} & \hspace{-0.9em} 0.52 \hspace{-0.4em} & \hspace{-0.9em} 0.61 \hspace{-0.4em} & \hspace{-0.9em} 0.09 \hspace{-0.4em} & \hspace{-0.9em} 0.08 \hspace{-0.4em} & \hspace{-0.9em} 0.26 \hspace{-0.4em} & \hspace{-0.9em} 0.01 \hspace{-0.4em} & \hspace{-0.9em} 0.26 \hspace{-0.4em} & \hspace{-0.9em} 0.00 \hspace{-0.4em} & \hspace{-0.9em} 0.50 \hspace{-0.4em} & \hspace{-0.9em} 0.03 \hspace{-0.4em} & \hspace{-0.9em} 0.26 \hspace{-0.4em} & \hspace{-0.9em} 0.17 \hspace{-0.4em} & \hspace{-0.9em} 0.18 \hspace{-0.4em} & \hspace{-0.9em} 0.06 \hspace{-0.4em} & \hspace{-0.9em} 0.21 \hspace{-0.4em} & \hspace{-0.9em} 0.03 \hspace{-0.4em} & \hspace{-0.9em} 0.17 \hspace{-0.4em} & \hspace{-0.9em} 0.08 \hspace{-0.4em} &  \cellcolor[HTML]{FFFFED} \hspace{-0.9em}  0.228 \\

\hspace{-0.4em} & \hspace{-0.9em} C&
0.08 \hspace{-0.4em} & \hspace{-0.9em} 0.13 \hspace{-0.4em} & \hspace{-0.9em} 0.19 \hspace{-0.4em} & \hspace{-0.9em} 0.15 \hspace{-0.4em} & \hspace{-0.9em} 0.25 \hspace{-0.4em} & \hspace{-0.9em} 0.25 \hspace{-0.4em} & \hspace{-0.9em} 0.32 \hspace{-0.4em} & \hspace{-0.9em} 0.17 \hspace{-0.4em} & \hspace{-0.9em} 0.05 \hspace{-0.4em} & \hspace{-0.9em} 0.13 \hspace{-0.4em} & \hspace{-0.9em} 0.15 \hspace{-0.4em} & \hspace{-0.9em} 0.03 \hspace{-0.4em} & \hspace{-0.9em} 0.19 \hspace{-0.4em} & \hspace{-0.9em} 0.37 \hspace{-0.4em} & \hspace{-0.9em} 0.28 \hspace{-0.4em} & \hspace{-0.9em} 0.34 \hspace{-0.4em} & \hspace{-0.9em} 0.10 \hspace{-0.4em} & \hspace{-0.9em} 0.14 \hspace{-0.4em} & \hspace{-0.9em} 0.05 \hspace{-0.4em} & \hspace{-0.9em} 0.23 \hspace{-0.4em} & \hspace{-0.9em} 0.21 \hspace{-0.4em} & \hspace{-0.9em} 0.16 \hspace{-0.4em} & \hspace{-0.9em} 0.07 \hspace{-0.4em} & \cellcolor[HTML]{FFFFED} \hspace{-0.9em} 0.176  \\

\hspace{-0.4em} & \hspace{-0.9em} G&
0.03 \hspace{-0.4em} & \hspace{-0.9em} 0.03 \hspace{-0.4em} & \hspace{-0.9em} 0.19 \hspace{-0.4em} & \hspace{-0.9em} 0.22 \hspace{-0.4em} & \hspace{-0.9em} 0.44 \hspace{-0.4em} & \hspace{-0.9em} 0.16 \hspace{-0.4em} & \hspace{-0.9em} 0.23 \hspace{-0.4em} & \hspace{-0.9em} 0.01 \hspace{-0.4em} & \hspace{-0.9em} 0.03 \hspace{-0.4em} & \hspace{-0.9em} 0.07 \hspace{-0.4em} & \hspace{-0.9em} 0.45 \hspace{-0.4em} & \hspace{-0.9em} 0.01 \hspace{-0.4em} & \hspace{-0.9em} 0.22 \hspace{-0.4em} & \hspace{-0.9em} 0.29 \hspace{-0.4em} & \hspace{-0.9em} 0.29 \hspace{-0.4em} & \hspace{-0.9em} 0.04 \hspace{-0.4em} & \hspace{-0.9em} 0.09 \hspace{-0.4em} & \hspace{-0.9em} 0.11 \hspace{-0.4em} & \hspace{-0.9em} 0.02 \hspace{-0.4em} & \hspace{-0.9em} 0.21 \hspace{-0.4em} & \hspace{-0.9em} 0.03 \hspace{-0.4em} & \hspace{-0.9em} 0.10 \hspace{-0.4em} & \hspace{-0.9em} 0.02 \hspace{-0.4em} & \cellcolor[HTML]{FFFFED} \hspace{-0.9em} 0.143   \\ 

\hspace{-0.4em} & \hspace{-0.9em} C+G&
0.03 \hspace{-0.4em} & \hspace{-0.9em} 0.03 \hspace{-0.4em} & \hspace{-0.9em} 0.19 \hspace{-0.4em} & \hspace{-0.9em} 0.22 \hspace{-0.4em} & \hspace{-0.9em} 0.44 \hspace{-0.4em} & \hspace{-0.9em} 0.16 \hspace{-0.4em} & \hspace{-0.9em} 0.23 \hspace{-0.4em} & \hspace{-0.9em} 0.01 \hspace{-0.4em} & \hspace{-0.9em} 0.03 \hspace{-0.4em} & \hspace{-0.9em} 0.07 \hspace{-0.4em} & \hspace{-0.9em} 0.45 \hspace{-0.4em} & \hspace{-0.9em} 0.01 \hspace{-0.4em} & \hspace{-0.9em} 0.22 \hspace{-0.4em} & \hspace{-0.9em} 0.29 \hspace{-0.4em} & \hspace{-0.9em} 0.29 \hspace{-0.4em} & \hspace{-0.9em} 0.04 \hspace{-0.4em} & \hspace{-0.9em} 0.09 \hspace{-0.4em} & \hspace{-0.9em} 0.11 \hspace{-0.4em} & \hspace{-0.9em} 0.02 \hspace{-0.4em} & \hspace{-0.9em} 0.21 \hspace{-0.4em} & \hspace{-0.9em} 0.03 \hspace{-0.4em} & \hspace{-0.9em} 0.10 \hspace{-0.4em} & \hspace{-0.9em} 0.02 \hspace{-0.4em} & \cellcolor[HTML]{FFFFED} \hspace{-0.9em} 0.143   \\

\hspace{-0.4em} & \hspace{-0.9em} INB+C&
0.08 \hspace{-0.4em} & \hspace{-0.9em} 0.13 \hspace{-0.4em} & \hspace{-0.9em} 0.19 \hspace{-0.4em} & \hspace{-0.9em} 0.15 \hspace{-0.4em} & \hspace{-0.9em} 0.25 \hspace{-0.4em} & \hspace{-0.9em} 0.25 \hspace{-0.4em} & \hspace{-0.9em} 0.32 \hspace{-0.4em} & \hspace{-0.9em} 0.17 \hspace{-0.4em} & \hspace{-0.9em} 0.05 \hspace{-0.4em} & \hspace{-0.9em} 0.13 \hspace{-0.4em} & \hspace{-0.9em} 0.15 \hspace{-0.4em} & \hspace{-0.9em} 0.03 \hspace{-0.4em} & \hspace{-0.9em} 0.19 \hspace{-0.4em} & \hspace{-0.9em} 0.37 \hspace{-0.4em} & \hspace{-0.9em} 0.28 \hspace{-0.4em} & \hspace{-0.9em} 0.34 \hspace{-0.4em} & \hspace{-0.9em} 0.10 \hspace{-0.4em} & \hspace{-0.9em} 0.14 \hspace{-0.4em} & \hspace{-0.9em} 0.05 \hspace{-0.4em} & \hspace{-0.9em} 0.23 \hspace{-0.4em} & \hspace{-0.9em} 0.21 \hspace{-0.4em} & \hspace{-0.9em} 0.16 \hspace{-0.4em} & \hspace{-0.9em} 0.07 \hspace{-0.4em} & \cellcolor[HTML]{FFFFED} \hspace{-0.9em} 0.176  \\

\hspace{-0.4em} & \hspace{-0.9em} INB+G&
0.03 \hspace{-0.4em} & \hspace{-0.9em} 0.02 \hspace{-0.4em} & \hspace{-0.9em} 0.20 \hspace{-0.4em} & \hspace{-0.9em} 0.20 \hspace{-0.4em} & \hspace{-0.9em} 0.44 \hspace{-0.4em} & \hspace{-0.9em} 0.02 \hspace{-0.4em} & \hspace{-0.9em} 0.28 \hspace{-0.4em} & \hspace{-0.9em} 0.01 \hspace{-0.4em} & \hspace{-0.9em} 0.02 \hspace{-0.4em} & \hspace{-0.9em} 0.10 \hspace{-0.4em} & \hspace{-0.9em} 0.43 \hspace{-0.4em} & \hspace{-0.9em} 0.02 \hspace{-0.4em} & \hspace{-0.9em} 0.21 \hspace{-0.4em} & \hspace{-0.9em} 0.33 \hspace{-0.4em} & \hspace{-0.9em} 0.25 \hspace{-0.4em} & \hspace{-0.9em} 0.03 \hspace{-0.4em} & \hspace{-0.9em} 0.06 \hspace{-0.4em} & \hspace{-0.9em} 0.12 \hspace{-0.4em} & \hspace{-0.9em} 0.00 \hspace{-0.4em} & \hspace{-0.9em} 0.23 \hspace{-0.4em} & \hspace{-0.9em} 0.03 \hspace{-0.4em} & \hspace{-0.9em} 0.22 \hspace{-0.4em} & \hspace{-0.9em} 0.00 \hspace{-0.4em} & \cellcolor[HTML]{FFFFED} \hspace{-0.9em} \textbf{0.141} \\ 

\hspace{-0.4em} & \hspace{-0.9em} INB+C+G&
0.03 \hspace{-0.4em} & \hspace{-0.9em} 0.02 \hspace{-0.4em} & \hspace{-0.9em} 0.20 \hspace{-0.4em} & \hspace{-0.9em} 0.20 \hspace{-0.4em} & \hspace{-0.9em} 0.44 \hspace{-0.4em} & \hspace{-0.9em} 0.02 \hspace{-0.4em} & \hspace{-0.9em} 0.28 \hspace{-0.4em} & \hspace{-0.9em} 0.01 \hspace{-0.4em} & \hspace{-0.9em} 0.02 \hspace{-0.4em} & \hspace{-0.9em} 0.10 \hspace{-0.4em} & \hspace{-0.9em} 0.43 \hspace{-0.4em} & \hspace{-0.9em} 0.02 \hspace{-0.4em} & \hspace{-0.9em} 0.21 \hspace{-0.4em} & \hspace{-0.9em} 0.33 \hspace{-0.4em} & \hspace{-0.9em} 0.25 \hspace{-0.4em} & \hspace{-0.9em} 0.03 \hspace{-0.4em} & \hspace{-0.9em} 0.06 \hspace{-0.4em} & \hspace{-0.9em} 0.12 \hspace{-0.4em} & \hspace{-0.9em} 0.00 \hspace{-0.4em} & \hspace{-0.9em} 0.23 \hspace{-0.4em} & \hspace{-0.9em} 0.03 \hspace{-0.4em} & \hspace{-0.9em} 0.22 \hspace{-0.4em} & \hspace{-0.9em} 0.00 \hspace{-0.4em} & \cellcolor[HTML]{FFFFED} \hspace{-0.9em} \textbf{0.141} \\ 

    \bottomrule
\end{tabular}}

\vspace{0.01in}

\label{tab:main_table_acc}
\vspace{-0.1in}
\end{table*}

%% file: tables/lovm_mpcr.tex
\begin{table*}[t]
\caption{\textbf{LOVM Benchmark (mean per-class recall)}. We evaluate our method's performance over \numdatasets\ datasets and \nummodels\ pre-trained models.
(top) Model Ranking results. (bottom) Performance Prediction results. INB - ImageNet Benchmark score, C - Text Classification scores G - Granularity Scores.
}
\vspace{0.8em}
\adjustbox{width=\textwidth}{
\begin{tabular}{cc|ccccccccccccccccccccccc|c}

    & 
    & \rotatebox[origin=lb]{90}{\smash{Stanford Cars}} 
    & \rotatebox[origin=lb]{90}{\smash{CIFAR100}} 
    & \rotatebox[origin=lb]{90}{\smash{CLEVR-DIST.}} 
    & \rotatebox[origin=lb]{90}{\smash{CLEVR-COUNT}} 
    & \rotatebox[origin=lb]{90}{\smash{Country211}} 
    & \rotatebox[origin=lb]{90}{\smash{Retinopathy}} 
    & \rotatebox[origin=lb]{90}{\smash{DMLab}} 
    & \rotatebox[origin=lb]{90}{\smash{DTD}} 
    & \rotatebox[origin=lb]{90}{\smash{EuroSAT}} 
    & \rotatebox[origin=lb]{90}{\smash{FER2013}} 
    & \rotatebox[origin=lb]{90}{\smash{Flowers102}} 
    & \rotatebox[origin=lb]{90}{\smash{GTSRB}} 
    & \rotatebox[origin=lb]{90}{\smash{ImageNet}} 
    & \rotatebox[origin=lb]{90}{\smash{KITTI}} 
    & \rotatebox[origin=lb]{90}{\smash{MNIST}} 
    & \rotatebox[origin=lb]{90}{\smash{PCam}}
    & \rotatebox[origin=lb]{90}{\smash{Oxford Pets}} 
    &  \rotatebox[origin=lb]{90}{\smash{Rendered SST2}}
    & \rotatebox[origin=lb]{90}{\smash{RESISC45}}
    & \rotatebox[origin=lb]{90}{\smash{STL10}} 
    & \rotatebox[origin=lb]{90}{\smash{SUN397}} 
    & \rotatebox[origin=lb]{90}{\smash{SVHN}} 
    & \rotatebox[origin=lb]{90}{\smash{VOC2007}} 
    %& \rotatebox[origin=lb]{90}{\smash{UCF101}} 
    %& \rotatebox[origin=lb]{90}{\smash{Kinetics700}} 
& \cellcolor[HTML]{FFFFED} \rotatebox[origin=lb]{90}{\smash{Mean}} 
    \\
    \midrule
    
\multirow{7}{*}{\rotatebox[origin=c]{90}{$R_5$}}

\hspace{-0.4em} &  \hspace{-0.9em} INB &
0.80 \hspace{-0.4em} & \hspace{-0.9em} 0.80 \hspace{-0.4em} & \hspace{-0.9em} 0.40 \hspace{-0.4em} & \hspace{-0.9em} 0.60 \hspace{-0.4em} & \hspace{-0.9em} 0.40 \hspace{-0.4em} & \hspace{-0.9em} 0.40 \hspace{-0.4em} & \hspace{-0.9em} 0.00 \hspace{-0.4em} & \hspace{-0.9em} 1.00 \hspace{-0.4em} & \hspace{-0.9em} 0.60 \hspace{-0.4em} & \hspace{-0.9em} 0.40 \hspace{-0.4em} & \hspace{-0.9em} 0.60 \hspace{-0.4em} & \hspace{-0.9em} 0.60 \hspace{-0.4em} & \hspace{-0.9em} 1.00 \hspace{-0.4em} & \hspace{-0.9em} 0.20 \hspace{-0.4em} & \hspace{-0.9em} 0.20 \hspace{-0.4em} & \hspace{-0.9em} 0.00 \hspace{-0.4em} & \hspace{-0.9em} 0.60 \hspace{-0.4em} & \hspace{-0.9em} 0.00 \hspace{-0.4em} & \hspace{-0.9em} 1.00 \hspace{-0.4em} & \hspace{-0.9em} 0.20 \hspace{-0.4em} & \hspace{-0.9em} 0.80 \hspace{-0.4em} & \hspace{-0.9em} 0.60 \hspace{-0.4em} & \hspace{-0.9em} 0.40 \hspace{-0.4em} & \cellcolor[HTML]{FFFFED} \hspace{-0.9em} 0.504     \\

\hspace{-0.4em} &  \hspace{-0.9em} C &
0.00 \hspace{-0.4em} & \hspace{-0.9em} 0.20 \hspace{-0.4em} & \hspace{-0.9em} 0.20 \hspace{-0.4em} & \hspace{-0.9em} 0.60 \hspace{-0.4em} & \hspace{-0.9em} 0.40 \hspace{-0.4em} & \hspace{-0.9em} 0.20 \hspace{-0.4em} & \hspace{-0.9em} 0.20 \hspace{-0.4em} & \hspace{-0.9em} 0.60 \hspace{-0.4em} & \hspace{-0.9em} 0.40 \hspace{-0.4em} & \hspace{-0.9em} 0.40 \hspace{-0.4em} & \hspace{-0.9em} 0.00 \hspace{-0.4em} & \hspace{-0.9em} 0.40 \hspace{-0.4em} & \hspace{-0.9em} 0.00 \hspace{-0.4em} & \hspace{-0.9em} 0.40 \hspace{-0.4em} & \hspace{-0.9em} 0.20 \hspace{-0.4em} & \hspace{-0.9em} 0.20 \hspace{-0.4em} & \hspace{-0.9em} 0.20 \hspace{-0.4em} & \hspace{-0.9em} 0.20 \hspace{-0.4em} & \hspace{-0.9em} 0.40 \hspace{-0.4em} & \hspace{-0.9em} 0.20 \hspace{-0.4em} & \hspace{-0.9em} 0.40 \hspace{-0.4em} & \hspace{-0.9em} 0.00 \hspace{-0.4em} & \hspace{-0.9em} 0.00 \hspace{-0.4em} &  \cellcolor[HTML]{FFFFED} \hspace{-0.9em} 0.252   \\

\hspace{-0.4em} &  \hspace{-0.9em} G &
0.40 \hspace{-0.4em} & \hspace{-0.9em} 0.40 \hspace{-0.4em} & \hspace{-0.9em} 0.20 \hspace{-0.4em} & \hspace{-0.9em} 0.20 \hspace{-0.4em} & \hspace{-0.9em} 0.40 \hspace{-0.4em} & \hspace{-0.9em} 0.00 \hspace{-0.4em} & \hspace{-0.9em} 0.40 \hspace{-0.4em} & \hspace{-0.9em} 0.40 \hspace{-0.4em} & \hspace{-0.9em} 0.20 \hspace{-0.4em} & \hspace{-0.9em} 0.20 \hspace{-0.4em} & \hspace{-0.9em} 0.60 \hspace{-0.4em} & \hspace{-0.9em} 0.20 \hspace{-0.4em} & \hspace{-0.9em} 0.40 \hspace{-0.4em} & \hspace{-0.9em} 0.40 \hspace{-0.4em} & \hspace{-0.9em} 0.00 \hspace{-0.4em} & \hspace{-0.9em} 0.20 \hspace{-0.4em} & \hspace{-0.9em} 0.40 \hspace{-0.4em} & \hspace{-0.9em} 0.00 \hspace{-0.4em} & \hspace{-0.9em} 0.40 \hspace{-0.4em} & \hspace{-0.9em} 0.20 \hspace{-0.4em} & \hspace{-0.9em} 0.40 \hspace{-0.4em} & \hspace{-0.9em} 0.20 \hspace{-0.4em} & \hspace{-0.9em} 0.00 \hspace{-0.4em} & \hspace{-0.9em} \cellcolor[HTML]{FFFFED} 0.270 \\ 

\hspace{-0.4em} &  \hspace{-0.9em} G+C &
0.00 \hspace{-0.4em} & \hspace{-0.9em} 0.40 \hspace{-0.4em} & \hspace{-0.9em} 0.20 \hspace{-0.4em} & \hspace{-0.9em} 0.20 \hspace{-0.4em} & \hspace{-0.9em} 0.60 \hspace{-0.4em} & \hspace{-0.9em} 0.40 \hspace{-0.4em} & \hspace{-0.9em} 0.60 \hspace{-0.4em} & \hspace{-0.9em} 0.40 \hspace{-0.4em} & \hspace{-0.9em} 0.20 \hspace{-0.4em} & \hspace{-0.9em} 0.20 \hspace{-0.4em} & \hspace{-0.9em} 0.00 \hspace{-0.4em} & \hspace{-0.9em} 0.60 \hspace{-0.4em} & \hspace{-0.9em} 0.20 \hspace{-0.4em} & \hspace{-0.9em} 0.20 \hspace{-0.4em} & \hspace{-0.9em} 0.40 \hspace{-0.4em} & \hspace{-0.9em} 0.20 \hspace{-0.4em} & \hspace{-0.9em} 0.00 \hspace{-0.4em} & \hspace{-0.9em} 0.00 \hspace{-0.4em} & \hspace{-0.9em} 0.40 \hspace{-0.4em} & \hspace{-0.9em} 0.20 \hspace{-0.4em} & \hspace{-0.9em} 0.40 \hspace{-0.4em} & \hspace{-0.9em} 0.20 \hspace{-0.4em} & \hspace{-0.9em} 0.00 \hspace{-0.4em} & \cellcolor[HTML]{FFFFED} \hspace{-0.9em} 0.261  \\

\hspace{-0.4em} &  \hspace{-0.9em} INB+C &
0.80 \hspace{-0.4em} & \hspace{-0.9em} 0.80 \hspace{-0.4em} & \hspace{-0.9em} 0.40 \hspace{-0.4em} & \hspace{-0.9em} 0.60 \hspace{-0.4em} & \hspace{-0.9em} 0.60 \hspace{-0.4em} & \hspace{-0.9em} 0.40 \hspace{-0.4em} & \hspace{-0.9em} 0.00 \hspace{-0.4em} & \hspace{-0.9em} 1.00 \hspace{-0.4em} & \hspace{-0.9em} 0.60 \hspace{-0.4em} & \hspace{-0.9em} 0.40 \hspace{-0.4em} & \hspace{-0.9em} 0.80 \hspace{-0.4em} & \hspace{-0.9em} 0.60 \hspace{-0.4em} & \hspace{-0.9em} 0.80 \hspace{-0.4em} & \hspace{-0.9em} 0.20 \hspace{-0.4em} & \hspace{-0.9em} 0.20 \hspace{-0.4em} & \hspace{-0.9em} 0.20 \hspace{-0.4em} & \hspace{-0.9em} 0.80 \hspace{-0.4em} & \hspace{-0.9em} 0.00 \hspace{-0.4em} & \hspace{-0.9em} 1.00 \hspace{-0.4em} & \hspace{-0.9em} 0.20 \hspace{-0.4em} & \hspace{-0.9em} 0.80 \hspace{-0.4em} & \hspace{-0.9em} 0.40 \hspace{-0.4em} & \hspace{-0.9em} 0.40 \hspace{-0.4em} & \cellcolor[HTML]{FFFFED} \hspace{-0.9em} 0.522  \\

\hspace{-0.4em} &  \hspace{-0.9em} INB+G &
0.80 \hspace{-0.4em} & \hspace{-0.9em} 0.80 \hspace{-0.4em} & \hspace{-0.9em} 0.40 \hspace{-0.4em} & \hspace{-0.9em} 0.60 \hspace{-0.4em} & \hspace{-0.9em} 0.40 \hspace{-0.4em} & \hspace{-0.9em} 0.40 \hspace{-0.4em} & \hspace{-0.9em} 0.00 \hspace{-0.4em} & \hspace{-0.9em} 1.00 \hspace{-0.4em} & \hspace{-0.9em} 0.60 \hspace{-0.4em} & \hspace{-0.9em} 0.40 \hspace{-0.4em} & \hspace{-0.9em} 0.60 \hspace{-0.4em} & \hspace{-0.9em} 0.60 \hspace{-0.4em} & \hspace{-0.9em} 1.00 \hspace{-0.4em} & \hspace{-0.9em} 0.20 \hspace{-0.4em} & \hspace{-0.9em} 0.40 \hspace{-0.4em} & \hspace{-0.9em} 0.20 \hspace{-0.4em} & \hspace{-0.9em} 0.60 \hspace{-0.4em} & \hspace{-0.9em} 0.00 \hspace{-0.4em} & \hspace{-0.9em} 1.00 \hspace{-0.4em} & \hspace{-0.9em} 0.40 \hspace{-0.4em} & \hspace{-0.9em} 0.80 \hspace{-0.4em} & \hspace{-0.9em} 0.60 \hspace{-0.4em} & \hspace{-0.9em} 0.60 \hspace{-0.4em} & \cellcolor[HTML]{FFFFED} \hspace{-0.9em} \textbf{0.539} \\

\hspace{-0.4em} &  \hspace{-0.9em} INB+C+G &
0.80 \hspace{-0.4em} & \hspace{-0.9em} 0.80 \hspace{-0.4em} & \hspace{-0.9em} 0.40 \hspace{-0.4em} & \hspace{-0.9em} 0.60 \hspace{-0.4em} & \hspace{-0.9em} 0.40 \hspace{-0.4em} & \hspace{-0.9em} 0.40 \hspace{-0.4em} & \hspace{-0.9em} 0.00 \hspace{-0.4em} & \hspace{-0.9em} 1.00 \hspace{-0.4em} & \hspace{-0.9em} 0.60 \hspace{-0.4em} & \hspace{-0.9em} 0.40 \hspace{-0.4em} & \hspace{-0.9em} 0.60 \hspace{-0.4em} & \hspace{-0.9em} 0.60 \hspace{-0.4em} & \hspace{-0.9em} 1.00 \hspace{-0.4em} & \hspace{-0.9em} 0.20 \hspace{-0.4em} & \hspace{-0.9em} 0.40 \hspace{-0.4em} & \hspace{-0.9em} 0.20 \hspace{-0.4em} & \hspace{-0.9em} 0.60 \hspace{-0.4em} & \hspace{-0.9em} 0.00 \hspace{-0.4em} & \hspace{-0.9em} 1.00 \hspace{-0.4em} & \hspace{-0.9em} 0.40 \hspace{-0.4em} & \hspace{-0.9em} 0.80 \hspace{-0.4em} & \hspace{-0.9em} 0.60 \hspace{-0.4em} & \hspace{-0.9em} 0.60 \hspace{-0.4em} & \cellcolor[HTML]{FFFFED} \hspace{-0.9em} \textbf{0.539} \\

\midrule

%%%%%%%%%%%%%%%%%%%%%%%%%%%

\multirow{7}{*}{\rotatebox[origin=c]{90}{$\tau$}}

\hspace{-0.4em} &  \hspace{-0.9em} INB&
0.67 \hspace{-0.4em} & \hspace{-0.9em} 0.67 \hspace{-0.4em} & \hspace{-0.9em} 0.00 \hspace{-0.4em} & \hspace{-0.9em} 0.33 \hspace{-0.4em} & \hspace{-0.9em} 0.00 \hspace{-0.4em} & \hspace{-0.9em} 0.00 \hspace{-0.4em} & \hspace{-0.9em} 0.00 \hspace{-0.4em} & \hspace{-0.9em} 0.00 \hspace{-0.4em} & \hspace{-0.9em} 1.00 \hspace{-0.4em} & \hspace{-0.9em} 0.00 \hspace{-0.4em} & \hspace{-0.9em} 1.00 \hspace{-0.4em} & \hspace{-0.9em} 1.00 \hspace{-0.4em} & \hspace{-0.9em} 1.00 \hspace{-0.4em} & \hspace{-0.9em} 0.00 \hspace{-0.4em} & \hspace{-0.9em} 0.00 \hspace{-0.4em} & \hspace{-0.9em} 0.00 \hspace{-0.4em} & \hspace{-0.9em} -0.33 \hspace{-0.4em} & \hspace{-0.9em} 0.00 \hspace{-0.4em} & \hspace{-0.9em} -0.40 \hspace{-0.4em} & \hspace{-0.9em} 0.00 \hspace{-0.4em} & \hspace{-0.9em} -0.33 \hspace{-0.4em} & \hspace{-0.9em} -0.33 \hspace{-0.4em} & \hspace{-0.9em} 0.00 \hspace{-0.4em} &  \cellcolor[HTML]{FFFFED} \hspace{-0.9em} 0.186  \\

\hspace{-0.4em} & \hspace{-0.9em} C &
0.00 \hspace{-0.4em} & \hspace{-0.9em} 0.00 \hspace{-0.4em} & \hspace{-0.9em} 0.00 \hspace{-0.4em} & \hspace{-0.9em} 1.00 \hspace{-0.4em} & \hspace{-0.9em} 0.00 \hspace{-0.4em} & \hspace{-0.9em} 0.00 \hspace{-0.4em} & \hspace{-0.9em} 0.00 \hspace{-0.4em} & \hspace{-0.9em} 0.33 \hspace{-0.4em} & \hspace{-0.9em} 0.00 \hspace{-0.4em} & \hspace{-0.9em} 0.00 \hspace{-0.4em} & \hspace{-0.9em} 0.00 \hspace{-0.4em} & \hspace{-0.9em} 0.00 \hspace{-0.4em} & \hspace{-0.9em} 0.00 \hspace{-0.4em} & \hspace{-0.9em} 0.00 \hspace{-0.4em} & \hspace{-0.9em} 0.00 \hspace{-0.4em} & \hspace{-0.9em} 0.00 \hspace{-0.4em} & \hspace{-0.9em} 0.00 \hspace{-0.4em} & \hspace{-0.9em} 0.00 \hspace{-0.4em} & \hspace{-0.9em} 0.00 \hspace{-0.4em} & \hspace{-0.9em} 0.00 \hspace{-0.4em} & \hspace{-0.9em} 0.00 \hspace{-0.4em} & \hspace{-0.9em} 0.00 \hspace{-0.4em} & \hspace{-0.9em} 0.00 \hspace{-0.4em} &  \cellcolor[HTML]{FFFFED} \hspace{-0.9em} 0.058 \\

\hspace{-0.4em} & \hspace{-0.9em} G &
0.00 \hspace{-0.4em} & \hspace{-0.9em} 0.00 \hspace{-0.4em} & \hspace{-0.9em} 0.00 \hspace{-0.4em} & \hspace{-0.9em} 0.00 \hspace{-0.4em} & \hspace{-0.9em} 0.00 \hspace{-0.4em} & \hspace{-0.9em} 0.00 \hspace{-0.4em} & \hspace{-0.9em} 0.00 \hspace{-0.4em} & \hspace{-0.9em} 0.00 \hspace{-0.4em} & \hspace{-0.9em} 0.00 \hspace{-0.4em} & \hspace{-0.9em} 0.00 \hspace{-0.4em} & \hspace{-0.9em} -0.33 \hspace{-0.4em} & \hspace{-0.9em} 0.00 \hspace{-0.4em} & \hspace{-0.9em} 0.00 \hspace{-0.4em} & \hspace{-0.9em} 0.00 \hspace{-0.4em} & \hspace{-0.9em} 0.00 \hspace{-0.4em} & \hspace{-0.9em} 0.00 \hspace{-0.4em} & \hspace{-0.9em} 0.00 \hspace{-0.4em} & \hspace{-0.9em} 0.00 \hspace{-0.4em} & \hspace{-0.9em} 0.00 \hspace{-0.4em} & \hspace{-0.9em} 0.00 \hspace{-0.4em} & \hspace{-0.9em} 0.00 \hspace{-0.4em} & \hspace{-0.9em} 0.00 \hspace{-0.4em} & \hspace{-0.9em} 0.00 \hspace{-0.4em} & \cellcolor[HTML]{FFFFED}\hspace{-0.9em} -0.014      \\

\hspace{-0.4em} & \hspace{-0.9em} C+G &
0.00 \hspace{-0.4em} & \hspace{-0.9em} 0.00 \hspace{-0.4em} & \hspace{-0.9em} 0.00 \hspace{-0.4em} & \hspace{-0.9em} 0.00 \hspace{-0.4em} & \hspace{-0.9em} 1.00 \hspace{-0.4em} & \hspace{-0.9em} 0.00 \hspace{-0.4em} & \hspace{-0.9em} 1.00 \hspace{-0.4em} & \hspace{-0.9em} 0.00 \hspace{-0.4em} & \hspace{-0.9em} 0.00 \hspace{-0.4em} & \hspace{-0.9em} 0.00 \hspace{-0.4em} & \hspace{-0.9em} 0.00 \hspace{-0.4em} & \hspace{-0.9em} 0.33 \hspace{-0.4em} & \hspace{-0.9em} 0.00 \hspace{-0.4em} & \hspace{-0.9em} 0.00 \hspace{-0.4em} & \hspace{-0.9em} 0.00 \hspace{-0.4em} & \hspace{-0.9em} 0.00 \hspace{-0.4em} & \hspace{-0.9em} 0.00 \hspace{-0.4em} & \hspace{-0.9em} 0.00 \hspace{-0.4em} & \hspace{-0.9em} 0.00 \hspace{-0.4em} & \hspace{-0.9em} 0.00 \hspace{-0.4em} & \hspace{-0.9em} 0.00 \hspace{-0.4em} & \hspace{-0.9em} 0.00 \hspace{-0.4em} & \hspace{-0.9em} 0.00 \hspace{-0.4em} & \cellcolor[HTML]{FFFFED} \hspace{-0.9em} 0.101 \\

\hspace{-0.4em} & \hspace{-0.9em} INB+C &
0.67 \hspace{-0.4em} & \hspace{-0.9em} 0.67 \hspace{-0.4em} & \hspace{-0.9em} 0.00 \hspace{-0.4em} & \hspace{-0.9em} 0.33 \hspace{-0.4em} & \hspace{-0.9em} -0.33 \hspace{-0.4em} & \hspace{-0.9em} 0.00 \hspace{-0.4em} & \hspace{-0.9em} 0.00 \hspace{-0.4em} & \hspace{-0.9em} -0.20 \hspace{-0.4em} & \hspace{-0.9em} 1.00 \hspace{-0.4em} & \hspace{-0.9em} 0.00 \hspace{-0.4em} & \hspace{-0.9em} 1.00 \hspace{-0.4em} & \hspace{-0.9em} 1.00 \hspace{-0.4em} & \hspace{-0.9em} 1.00 \hspace{-0.4em} & \hspace{-0.9em} 0.00 \hspace{-0.4em} & \hspace{-0.9em} 0.00 \hspace{-0.4em} & \hspace{-0.9em} 0.00 \hspace{-0.4em} & \hspace{-0.9em} 0.00 \hspace{-0.4em} & \hspace{-0.9em} 0.00 \hspace{-0.4em} & \hspace{-0.9em} -0.20 \hspace{-0.4em} & \hspace{-0.9em} 0.00 \hspace{-0.4em} & \hspace{-0.9em} 0.00 \hspace{-0.4em} & \hspace{-0.9em} 0.00 \hspace{-0.4em} & \hspace{-0.9em} 0.00 \hspace{-0.4em} & \cellcolor[HTML]{FFFFED} \hspace{-0.9em} \textbf{0.214}  \\

\hspace{-0.4em} & \hspace{-0.9em} INB+G &
0.67 \hspace{-0.4em} & \hspace{-0.9em} 0.33 \hspace{-0.4em} & \hspace{-0.9em} 0.00 \hspace{-0.4em} & \hspace{-0.9em} 0.33 \hspace{-0.4em} & \hspace{-0.9em} 0.00 \hspace{-0.4em} & \hspace{-0.9em} 0.00 \hspace{-0.4em} & \hspace{-0.9em} 0.00 \hspace{-0.4em} & \hspace{-0.9em} 0.00 \hspace{-0.4em} & \hspace{-0.9em} 1.00 \hspace{-0.4em} & \hspace{-0.9em} 0.00 \hspace{-0.4em} & \hspace{-0.9em} 1.00 \hspace{-0.4em} & \hspace{-0.9em} 1.00 \hspace{-0.4em} & \hspace{-0.9em} 0.80 \hspace{-0.4em} & \hspace{-0.9em} 0.00 \hspace{-0.4em} & \hspace{-0.9em} 0.00 \hspace{-0.4em} & \hspace{-0.9em} 0.00 \hspace{-0.4em} & \hspace{-0.9em} 0.33 \hspace{-0.4em} & \hspace{-0.9em} 0.00 \hspace{-0.4em} & \hspace{-0.9em} -0.60 \hspace{-0.4em} & \hspace{-0.9em} 0.00 \hspace{-0.4em} & \hspace{-0.9em} -0.33 \hspace{-0.4em} & \hspace{-0.9em} -0.33 \hspace{-0.4em} & \hspace{-0.9em} 0.33 \hspace{-0.4em} &\cellcolor[HTML]{FFFFED}  \hspace{-0.9em} 0.197 \\

\hspace{-0.4em} & \hspace{-0.9em} INB+C+G &
0.67 \hspace{-0.4em} & \hspace{-0.9em} 0.33 \hspace{-0.4em} & \hspace{-0.9em} 0.00 \hspace{-0.4em} & \hspace{-0.9em} 0.33 \hspace{-0.4em} & \hspace{-0.9em} 0.00 \hspace{-0.4em} & \hspace{-0.9em} 0.00 \hspace{-0.4em} & \hspace{-0.9em} 0.00 \hspace{-0.4em} & \hspace{-0.9em} 0.00 \hspace{-0.4em} & \hspace{-0.9em} 1.00 \hspace{-0.4em} & \hspace{-0.9em} 0.00 \hspace{-0.4em} & \hspace{-0.9em} 1.00 \hspace{-0.4em} & \hspace{-0.9em} 1.00 \hspace{-0.4em} & \hspace{-0.9em} 0.80 \hspace{-0.4em} & \hspace{-0.9em} 0.00 \hspace{-0.4em} & \hspace{-0.9em} 0.00 \hspace{-0.4em} & \hspace{-0.9em} 0.00 \hspace{-0.4em} & \hspace{-0.9em} 0.33 \hspace{-0.4em} & \hspace{-0.9em} 0.00 \hspace{-0.4em} & \hspace{-0.9em} -0.60 \hspace{-0.4em} & \hspace{-0.9em} 0.00 \hspace{-0.4em} & \hspace{-0.9em} -0.33 \hspace{-0.4em} & \hspace{-0.9em} -0.33 \hspace{-0.4em} & \hspace{-0.9em} 0.33 \hspace{-0.4em} & \cellcolor[HTML]{FFFFED}  \hspace{-0.9em} 0.197 \\

\midrule 
\midrule 

%%%%%%%%%%%%%%%%%%%%%%%%%%%

\multirow{7}{*}{\rotatebox[origin=c]{90}{$L_1$}}
\hspace{-0.4em} & \hspace{-0.9em} INB&
0.16 \hspace{-0.4em} & \hspace{-0.9em} 0.07 \hspace{-0.4em} & \hspace{-0.9em} 0.58 \hspace{-0.4em} & \hspace{-0.9em} 0.42 \hspace{-0.4em} & \hspace{-0.9em} 0.49 \hspace{-0.4em} & \hspace{-0.9em} 0.52 \hspace{-0.4em} & \hspace{-0.9em} 0.61 \hspace{-0.4em} & \hspace{-0.9em} 0.09 \hspace{-0.4em} & \hspace{-0.9em} 0.08 \hspace{-0.4em} & \hspace{-0.9em} 0.26 \hspace{-0.4em} & \hspace{-0.9em} 0.01 \hspace{-0.4em} & \hspace{-0.9em} 0.26 \hspace{-0.4em} & \hspace{-0.9em} 0.00 \hspace{-0.4em} & \hspace{-0.9em} 0.50 \hspace{-0.4em} & \hspace{-0.9em} 0.03 \hspace{-0.4em} & \hspace{-0.9em} 0.26 \hspace{-0.4em} & \hspace{-0.9em} 0.17 \hspace{-0.4em} & \hspace{-0.9em} 0.18 \hspace{-0.4em} & \hspace{-0.9em} 0.06 \hspace{-0.4em} & \hspace{-0.9em} 0.21 \hspace{-0.4em} & \hspace{-0.9em} 0.03 \hspace{-0.4em} & \hspace{-0.9em} 0.17 \hspace{-0.4em} & \hspace{-0.9em} 0.08 \hspace{-0.4em} &  \cellcolor[HTML]{FFFFED} \hspace{-0.9em}  0.228 \\

\hspace{-0.4em} & \hspace{-0.9em} C&
0.07 \hspace{-0.4em} & \hspace{-0.9em} 0.09 \hspace{-0.4em} & \hspace{-0.9em} 0.22 \hspace{-0.4em} & \hspace{-0.9em} 0.14 \hspace{-0.4em} & \hspace{-0.9em} 0.26 \hspace{-0.4em} & \hspace{-0.9em} 0.27 \hspace{-0.4em} & \hspace{-0.9em} 0.31 \hspace{-0.4em} & \hspace{-0.9em} 0.16 \hspace{-0.4em} & \hspace{-0.9em} 0.03 \hspace{-0.4em} & \hspace{-0.9em} 0.15 \hspace{-0.4em} & \hspace{-0.9em} 0.17 \hspace{-0.4em} & \hspace{-0.9em} 0.08 \hspace{-0.4em} & \hspace{-0.9em} 0.22 \hspace{-0.4em} & \hspace{-0.9em} 0.33 \hspace{-0.4em} & \hspace{-0.9em} 0.30 \hspace{-0.4em} & \hspace{-0.9em} 0.34 \hspace{-0.4em} & \hspace{-0.9em} 0.09 \hspace{-0.4em} & \hspace{-0.9em} 0.16 \hspace{-0.4em} & \hspace{-0.9em} 0.05 \hspace{-0.4em} & \hspace{-0.9em} 0.22 \hspace{-0.4em} & \hspace{-0.9em} 0.20 \hspace{-0.4em} & \hspace{-0.9em} 0.21 \hspace{-0.4em} & \hspace{-0.9em} 0.12 \hspace{-0.4em} &  \cellcolor[HTML]{FFFFED} \hspace{-0.9em} 0.183  \\

\hspace{-0.4em} & \hspace{-0.9em} G&
0.14 \hspace{-0.4em} & \hspace{-0.9em} 0.03 \hspace{-0.4em} & \hspace{-0.9em} 0.25 \hspace{-0.4em} & \hspace{-0.9em} 0.16 \hspace{-0.4em} & \hspace{-0.9em} 0.44 \hspace{-0.4em} & \hspace{-0.9em} 0.23 \hspace{-0.4em} & \hspace{-0.9em} 0.19 \hspace{-0.4em} & \hspace{-0.9em} 0.01 \hspace{-0.4em} & \hspace{-0.9em} 0.01 \hspace{-0.4em} & \hspace{-0.9em} 0.12 \hspace{-0.4em} & \hspace{-0.9em} 0.46 \hspace{-0.4em} & \hspace{-0.9em} 0.03 \hspace{-0.4em} & \hspace{-0.9em} 0.30 \hspace{-0.4em} & \hspace{-0.9em} 0.15 \hspace{-0.4em} & \hspace{-0.9em} 0.28 \hspace{-0.4em} & \hspace{-0.9em} 0.03 \hspace{-0.4em} & \hspace{-0.9em} 0.07 \hspace{-0.4em} & \hspace{-0.9em} 0.09 \hspace{-0.4em} & \hspace{-0.9em} 0.04 \hspace{-0.4em} & \hspace{-0.9em} 0.17 \hspace{-0.4em} & \hspace{-0.9em} 0.02 \hspace{-0.4em} & \hspace{-0.9em} 0.09 \hspace{-0.4em} & \hspace{-0.9em} 0.02 \hspace{-0.4em} &  \cellcolor[HTML]{FFFFED} \hspace{-0.9em} \textbf{0.145} \\ 

\hspace{-0.4em} & \hspace{-0.9em} C+G&
0.14 \hspace{-0.4em} & \hspace{-0.9em} 0.03 \hspace{-0.4em} & \hspace{-0.9em} 0.25 \hspace{-0.4em} & \hspace{-0.9em} 0.16 \hspace{-0.4em} & \hspace{-0.9em} 0.44 \hspace{-0.4em} & \hspace{-0.9em} 0.23 \hspace{-0.4em} & \hspace{-0.9em} 0.19 \hspace{-0.4em} & \hspace{-0.9em} 0.01 \hspace{-0.4em} & \hspace{-0.9em} 0.01 \hspace{-0.4em} & \hspace{-0.9em} 0.12 \hspace{-0.4em} & \hspace{-0.9em} 0.46 \hspace{-0.4em} & \hspace{-0.9em} 0.03 \hspace{-0.4em} & \hspace{-0.9em} 0.30 \hspace{-0.4em} & \hspace{-0.9em} 0.15 \hspace{-0.4em} & \hspace{-0.9em} 0.28 \hspace{-0.4em} & \hspace{-0.9em} 0.03 \hspace{-0.4em} & \hspace{-0.9em} 0.07 \hspace{-0.4em} & \hspace{-0.9em} 0.09 \hspace{-0.4em} & \hspace{-0.9em} 0.04 \hspace{-0.4em} & \hspace{-0.9em} 0.17 \hspace{-0.4em} & \hspace{-0.9em} 0.02 \hspace{-0.4em} & \hspace{-0.9em} 0.09 \hspace{-0.4em} & \hspace{-0.9em} 0.02 \hspace{-0.4em} &  \cellcolor[HTML]{FFFFED} \hspace{-0.9em} \textbf{0.145} \\

\hspace{-0.4em} & \hspace{-0.9em} INB+C&
0.07 \hspace{-0.4em} & \hspace{-0.9em} 0.09 \hspace{-0.4em} & \hspace{-0.9em} 0.22 \hspace{-0.4em} & \hspace{-0.9em} 0.14 \hspace{-0.4em} & \hspace{-0.9em} 0.26 \hspace{-0.4em} & \hspace{-0.9em} 0.27 \hspace{-0.4em} & \hspace{-0.9em} 0.31 \hspace{-0.4em} & \hspace{-0.9em} 0.16 \hspace{-0.4em} & \hspace{-0.9em} 0.03 \hspace{-0.4em} & \hspace{-0.9em} 0.15 \hspace{-0.4em} & \hspace{-0.9em} 0.17 \hspace{-0.4em} & \hspace{-0.9em} 0.08 \hspace{-0.4em} & \hspace{-0.9em} 0.22 \hspace{-0.4em} & \hspace{-0.9em} 0.33 \hspace{-0.4em} & \hspace{-0.9em} 0.30 \hspace{-0.4em} & \hspace{-0.9em} 0.34 \hspace{-0.4em} & \hspace{-0.9em} 0.09 \hspace{-0.4em} & \hspace{-0.9em} 0.16 \hspace{-0.4em} & \hspace{-0.9em} 0.05 \hspace{-0.4em} & \hspace{-0.9em} 0.22 \hspace{-0.4em} & \hspace{-0.9em} 0.20 \hspace{-0.4em} & \hspace{-0.9em} 0.21 \hspace{-0.4em} & \hspace{-0.9em} 0.12 \hspace{-0.4em} &  \cellcolor[HTML]{FFFFED} \hspace{-0.9em} 0.183  \\

\hspace{-0.4em} & \hspace{-0.9em} INB+G&
0.14 \hspace{-0.4em} & \hspace{-0.9em} 0.03 \hspace{-0.4em} & \hspace{-0.9em} 0.25 \hspace{-0.4em} & \hspace{-0.9em} 0.16 \hspace{-0.4em} & \hspace{-0.9em} 0.44 \hspace{-0.4em} & \hspace{-0.9em} 0.23 \hspace{-0.4em} & \hspace{-0.9em} 0.19 \hspace{-0.4em} & \hspace{-0.9em} 0.01 \hspace{-0.4em} & \hspace{-0.9em} 0.01 \hspace{-0.4em} & \hspace{-0.9em} 0.12 \hspace{-0.4em} & \hspace{-0.9em} 0.46 \hspace{-0.4em} & \hspace{-0.9em} 0.03 \hspace{-0.4em} & \hspace{-0.9em} 0.30 \hspace{-0.4em} & \hspace{-0.9em} 0.15 \hspace{-0.4em} & \hspace{-0.9em} 0.28 \hspace{-0.4em} & \hspace{-0.9em} 0.03 \hspace{-0.4em} & \hspace{-0.9em} 0.07 \hspace{-0.4em} & \hspace{-0.9em} 0.09 \hspace{-0.4em} & \hspace{-0.9em} 0.04 \hspace{-0.4em} & \hspace{-0.9em} 0.17 \hspace{-0.4em} & \hspace{-0.9em} 0.02 \hspace{-0.4em} & \hspace{-0.9em} 0.09 \hspace{-0.4em} & \hspace{-0.9em} 0.02 \hspace{-0.4em} &  \cellcolor[HTML]{FFFFED} \hspace{-0.9em} \textbf{0.145} \\ 

\hspace{-0.4em} & \hspace{-0.9em} INB+C+G&
0.14 \hspace{-0.4em} & \hspace{-0.9em} 0.03 \hspace{-0.4em} & \hspace{-0.9em} 0.25 \hspace{-0.4em} & \hspace{-0.9em} 0.16 \hspace{-0.4em} & \hspace{-0.9em} 0.44 \hspace{-0.4em} & \hspace{-0.9em} 0.23 \hspace{-0.4em} & \hspace{-0.9em} 0.19 \hspace{-0.4em} & \hspace{-0.9em} 0.01 \hspace{-0.4em} & \hspace{-0.9em} 0.01 \hspace{-0.4em} & \hspace{-0.9em} 0.12 \hspace{-0.4em} & \hspace{-0.9em} 0.46 \hspace{-0.4em} & \hspace{-0.9em} 0.03 \hspace{-0.4em} & \hspace{-0.9em} 0.30 \hspace{-0.4em} & \hspace{-0.9em} 0.15 \hspace{-0.4em} & \hspace{-0.9em} 0.28 \hspace{-0.4em} & \hspace{-0.9em} 0.03 \hspace{-0.4em} & \hspace{-0.9em} 0.07 \hspace{-0.4em} & \hspace{-0.9em} 0.09 \hspace{-0.4em} & \hspace{-0.9em} 0.04 \hspace{-0.4em} & \hspace{-0.9em} 0.17 \hspace{-0.4em} & \hspace{-0.9em} 0.02 \hspace{-0.4em} & \hspace{-0.9em} 0.09 \hspace{-0.4em} & \hspace{-0.9em} 0.02 \hspace{-0.4em} &  \cellcolor[HTML]{FFFFED} \hspace{-0.9em} \textbf{0.145} \\ 

    \bottomrule
\end{tabular}}

\vspace{0.01in}

\label{tab:main_table_mpcr}
\vspace{-0.1in}
\end{table*}

%% file: tables/ablation_acc1.tex
\begin{table}
\caption{\textbf{LOVM Model Selection Ablation.} Here, we ablate all the different scores used in our baselines for model ranking by top-1 accuracy. We separated the text classification (C) base scores, and the granularity-based scores that quantify inter- and intra-class similarity.
$a_{\text{IN}}$ - Imagenet Accuracy,  \intraClassCloseSymbl\ - \intraClassClose, text-f1 - caption dataset f1-score, text-acc1 - text top-1 accuracy, \superclassSymbl\ - \superclass, \intraSimSymbl\ - \intraSim, \interSimSymbl\ - \interSim.
\label{tab:ablation:rank_acc}}
\begin{multicols}{2}

\resizebox{1\textwidth}{!}{
\begin{tabular}{c|c|c|cc|cc|cc|cc@{}}
\toprule
&  & \multicolumn{7}{c|}{Scores} & \multicolumn{2}{c}{Metrics} \\
\midrule 
 & Row ID & $a_{\text{IN}}$ & text-f1 & text-acc1 &  \superclassSymbl  & \interSimSymbl & \intraSimSymbl & \intraClassCloseSymbl & $\tau$ ($\uparrow$) & $R_5$ ($\uparrow$) \\
\midrule
\multirow{64}{*}{\rotatebox[origin=c]{90}{\parbox[c]{4cm}{\centering \textbf{Model Ranking}}}}
& 1 & $\checkmark$ & $\times$ & $\times$ & $\times$ & $\times$ & $\times$ & $\times$ &  \cellcolor[HTML]{FFFFED} 0.177 &  \cellcolor[HTML]{FFFFED} 0.452  \\
& 2 & $\times$ & $\checkmark$ & $\times$ & $\times$ & $\times$ & $\times$ & $\times$ &  \cellcolor[HTML]{FFFFED} 0.058 &  \cellcolor[HTML]{FFFFED} 0.226  \\
& 3 & $\times$ & $\times$ & $\checkmark$ & $\times$ & $\times$ & $\times$ & $\times$ &  \cellcolor[HTML]{FFFFED} 0.029 &  \cellcolor[HTML]{FFFFED} 0.191  \\
& 4 & $\times$ & $\times$ & $\times$ & $\checkmark$ & $\times$ & $\times$ & $\times$ &  \cellcolor[HTML]{FFFFED} 0.000 &  \cellcolor[HTML]{FFFFED} 0.183  \\
& 5 & $\times$ & $\times$ & $\times$ & $\times$ & $\checkmark$ & $\times$ & $\times$ &  \cellcolor[HTML]{FFFFED} -0.014 &  \cellcolor[HTML]{FFFFED} 0.243  \\
&  6 & $\times$ & $\times$ & $\times$ & $\times$ & $\times$ & $\checkmark$ & $\times$ &  \cellcolor[HTML]{FFFFED} -0.014 &  \cellcolor[HTML]{FFFFED} 0.252  \\
& 7 & $\times$ & $\times$ & $\times$ & $\times$ & $\times$ & $\times$ & $\checkmark$ &  \cellcolor[HTML]{FFFFED} 0.000 &  \cellcolor[HTML]{FFFFED} 0.165  \\
&  8 & $\checkmark$ & $\checkmark$ & $\times$ & $\times$ & $\times$ & $\times$ & $\times$ &  \cellcolor[HTML]{FFFFED} 0.223 &  \cellcolor[HTML]{FFFFED} 0.461  \\
& 9 & $\checkmark$ & $\times$ & $\checkmark$ & $\times$ & $\times$ & $\times$ & $\times$ &  \cellcolor[HTML]{FFFFED} 0.200 &  \cellcolor[HTML]{FFFFED} 0.452  \\
& 10 & $\checkmark$ & $\times$ & $\times$ & $\checkmark$ & $\times$ & $\times$ & $\times$ &  \cellcolor[HTML]{FFFFED} 0.188 &  \cellcolor[HTML]{FFFFED} 0.426  \\
& 11 & $\checkmark$ & $\times$ & $\times$ & $\times$ & $\checkmark$ & $\times$ & $\times$ &  \cellcolor[HTML]{FFFFED} 0.096 &  \cellcolor[HTML]{FFFFED} 0.461  \\
& 12 & $\checkmark$ & $\times$ & $\times$ & $\times$ & $\times$ & $\checkmark$ & $\times$ &  \cellcolor[HTML]{FFFFED} 0.078 &  \cellcolor[HTML]{FFFFED} 0.417  \\
& 13 & $\checkmark$ & $\times$ & $\times$ & $\times$ & $\times$ & $\times$ & $\checkmark$ &  \cellcolor[HTML]{FFFFED} 0.119 &  \cellcolor[HTML]{FFFFED} 0.426  \\
& 14 & $\times$ & $\checkmark$ & $\checkmark$ & $\times$ & $\times$ & $\times$ & $\times$ &  \cellcolor[HTML]{FFFFED} 0.029 &  \cellcolor[HTML]{FFFFED} 0.226  \\
& 15 & $\times$ & $\checkmark$ & $\times$ & $\checkmark$ & $\times$ & $\times$ & $\times$ &  \cellcolor[HTML]{FFFFED} 0.043 &  \cellcolor[HTML]{FFFFED} 0.191  \\
& 16 & $\times$ & $\checkmark$ & $\times$ & $\times$ & $\checkmark$ & $\times$ & $\times$ &  \cellcolor[HTML]{FFFFED} 0.014 &  \cellcolor[HTML]{FFFFED} 0.209  \\
& 17 & $\times$ & $\checkmark$ & $\times$ & $\times$ & $\times$ & $\checkmark$ & $\times$ &  \cellcolor[HTML]{FFFFED} 0.014 &  \cellcolor[HTML]{FFFFED} 0.217  \\
& 18 & $\times$ & $\checkmark$ & $\times$ & $\times$ & $\times$ & $\times$ & $\checkmark$ &  \cellcolor[HTML]{FFFFED} 0.000 &  \cellcolor[HTML]{FFFFED} 0.183  \\
& 19 & $\times$ & $\times$ & $\checkmark$ & $\checkmark$ & $\times$ & $\times$ & $\times$ &  \cellcolor[HTML]{FFFFED} 0.043 &  \cellcolor[HTML]{FFFFED} 0.191  \\
& 20 & $\times$ & $\times$ & $\checkmark$ & $\times$ & $\checkmark$ & $\times$ & $\times$ &  \cellcolor[HTML]{FFFFED} 0.014 &  \cellcolor[HTML]{FFFFED} 0.209  \\
& 21 & $\times$ & $\times$ & $\checkmark$ & $\times$ & $\times$ & $\checkmark$ & $\times$ &  \cellcolor[HTML]{FFFFED} 0.014 &  \cellcolor[HTML]{FFFFED} 0.191  \\
& 22 & $\times$ & $\times$ & $\checkmark$ & $\times$ & $\times$ & $\times$ & $\checkmark$ &  \cellcolor[HTML]{FFFFED} 0.000 &  \cellcolor[HTML]{FFFFED} 0.174  \\
& 23 & $\times$ & $\times$ & $\times$ & $\checkmark$ & $\checkmark$ & $\times$ & $\times$ &  \cellcolor[HTML]{FFFFED} 0.000 &  \cellcolor[HTML]{FFFFED} 0.200  \\
& 24 & $\times$ & $\times$ & $\times$ & $\checkmark$ & $\times$ & $\checkmark$ & $\times$ &  \cellcolor[HTML]{FFFFED} 0.014 &  \cellcolor[HTML]{FFFFED} 0.235  \\
& 25 & $\times$ & $\times$ & $\times$ & $\checkmark$ & $\times$ & $\times$ & $\checkmark$ &  \cellcolor[HTML]{FFFFED} 0.000 &  \cellcolor[HTML]{FFFFED} 0.174  \\
& 26 & $\times$ & $\times$ & $\times$ & $\times$ & $\checkmark$ & $\checkmark$ & $\times$ &  \cellcolor[HTML]{FFFFED} 0.014 &  \cellcolor[HTML]{FFFFED} 0.209  \\
& 27 & $\times$ & $\times$ & $\times$ & $\times$ & $\checkmark$ & $\times$ & $\checkmark$ &  \cellcolor[HTML]{FFFFED} 0.000 &  \cellcolor[HTML]{FFFFED} 0.165  \\
& 28 & $\times$ & $\times$ & $\times$ & $\times$ & $\times$ & $\checkmark$ & $\checkmark$ &  \cellcolor[HTML]{FFFFED} 0.014 &  \cellcolor[HTML]{FFFFED} 0.200  \\
& 29 & $\checkmark$ & $\checkmark$ & $\checkmark$ & $\times$ & $\times$ & $\times$ & $\times$ &  \cellcolor[HTML]{FFFFED} 0.093 &  \cellcolor[HTML]{FFFFED} 0.443  \\
& 30 & $\checkmark$ & $\checkmark$ & $\times$ & $\checkmark$ & $\times$ & $\times$ & $\times$ &  \cellcolor[HTML]{FFFFED} 0.174 &  \cellcolor[HTML]{FFFFED} 0.443  \\
& 31 & $\checkmark$ & $\checkmark$ & $\times$ & $\times$ & $\checkmark$ & $\times$ & $\times$ &  \cellcolor[HTML]{FFFFED} 0.188 &  \cellcolor[HTML]{FFFFED} 0.435  \\
& 32 & $\checkmark$ & $\checkmark$ & $\times$ & $\times$ & $\times$ & $\checkmark$ & $\times$ &  \cellcolor[HTML]{FFFFED} 0.145 &  \cellcolor[HTML]{FFFFED} 0.443  \\
& 33 & $\checkmark$ & $\checkmark$ & $\times$ & $\times$ & $\times$ & $\times$ & $\checkmark$ &  \cellcolor[HTML]{FFFFED} 0.096 &  \cellcolor[HTML]{FFFFED} 0.417  \\
& 34 & $\checkmark$ & $\times$ & $\checkmark$ & $\checkmark$ & $\times$ & $\times$ & $\times$ &  \cellcolor[HTML]{FFFFED} 0.159 &  \cellcolor[HTML]{FFFFED} 0.435  \\
& 35 & $\checkmark$ & $\times$ & $\checkmark$ & $\times$ & $\checkmark$ & $\times$ & $\times$ &  \cellcolor[HTML]{FFFFED} 0.188 &  \cellcolor[HTML]{FFFFED} 0.443  \\
& 36 & $\checkmark$ & $\times$ & $\checkmark$ & $\times$ & $\times$ & $\checkmark$ & $\times$ &  \cellcolor[HTML]{FFFFED} 0.145 &  \cellcolor[HTML]{FFFFED} 0.452  \\
& 37 & $\checkmark$ & $\times$ & $\checkmark$ & $\times$ & $\times$ & $\times$ & $\checkmark$ &  \cellcolor[HTML]{FFFFED} 0.067 &  \cellcolor[HTML]{FFFFED} 0.417  \\
& 38 & $\checkmark$ & $\times$ & $\times$ & $\checkmark$ & $\checkmark$ & $\times$ & $\times$ &  \cellcolor[HTML]{FFFFED} 0.110 &  \cellcolor[HTML]{FFFFED} 0.426  \\
& 39 & $\checkmark$ & $\times$ & $\times$ & $\checkmark$ & $\times$ & $\checkmark$ & $\times$ &  \cellcolor[HTML]{FFFFED} 0.188 &  \cellcolor[HTML]{FFFFED} 0.426  \\
& 40 & $\checkmark$ & $\times$ & $\times$ & $\checkmark$ & $\times$ & $\times$ & $\checkmark$ &  \cellcolor[HTML]{FFFFED} 0.119 &  \cellcolor[HTML]{FFFFED} 0.417  \\
& 41 & $\checkmark$ & $\times$ & $\times$ & $\times$ & $\checkmark$ & $\checkmark$ & $\times$ &  \cellcolor[HTML]{FFFFED} 0.067 &  \cellcolor[HTML]{FFFFED} 0.435  \\
& 42 & $\checkmark$ & $\times$ & $\times$ & $\times$ & $\checkmark$ & $\times$ & $\checkmark$ &  \cellcolor[HTML]{FFFFED} 0.119 &  \cellcolor[HTML]{FFFFED} 0.417  \\
& 43 & $\checkmark$ & $\times$ & $\times$ & $\times$ & $\times$ & $\checkmark$ & $\checkmark$ &  \cellcolor[HTML]{FFFFED} 0.119 &  \cellcolor[HTML]{FFFFED} 0.426  \\
& 44 & $\times$ & $\checkmark$ & $\checkmark$ & $\checkmark$ & $\times$ & $\times$ & $\times$ &  \cellcolor[HTML]{FFFFED} 0.000 &  \cellcolor[HTML]{FFFFED} 0.209  \\
& 45 & $\times$ & $\checkmark$ & $\checkmark$ & $\times$ & $\checkmark$ & $\times$ & $\times$ &  \cellcolor[HTML]{FFFFED} 0.014 &  \cellcolor[HTML]{FFFFED} 0.200  \\
& 46 & $\times$ & $\checkmark$ & $\checkmark$ & $\times$ & $\times$ & $\checkmark$ & $\times$ &  \cellcolor[HTML]{FFFFED} 0.014 &  \cellcolor[HTML]{FFFFED} 0.209  \\
& 47 & $\times$ & $\checkmark$ & $\checkmark$ & $\times$ & $\times$ & $\times$ & $\checkmark$ &  \cellcolor[HTML]{FFFFED} 0.014 &  \cellcolor[HTML]{FFFFED} 0.209  \\
& 48 & $\times$ & $\checkmark$ & $\times$ & $\checkmark$ & $\checkmark$ & $\times$ & $\times$ &  \cellcolor[HTML]{FFFFED} 0.000 &  \cellcolor[HTML]{FFFFED} 0.217  \\
& 49 & $\times$ & $\checkmark$ & $\times$ & $\checkmark$ & $\times$ & $\checkmark$ & $\times$ &  \cellcolor[HTML]{FFFFED} 0.014 &  \cellcolor[HTML]{FFFFED} 0.217  \\
& 50 & $\times$ & $\checkmark$ & $\times$ & $\checkmark$ & $\times$ & $\times$ & $\checkmark$ &  \cellcolor[HTML]{FFFFED} 0.000 &  \cellcolor[HTML]{FFFFED} 0.191  \\
& 51 & $\times$ & $\checkmark$ & $\times$ & $\times$ & $\checkmark$ & $\checkmark$ & $\times$ &  \cellcolor[HTML]{FFFFED} 0.014 &  \cellcolor[HTML]{FFFFED} 0.209  \\
& 52 & $\times$ & $\checkmark$ & $\times$ & $\times$ & $\checkmark$ & $\times$ & $\checkmark$ &  \cellcolor[HTML]{FFFFED} 0.014 &  \cellcolor[HTML]{FFFFED} 0.217  \\
& 53 & $\times$ & $\checkmark$ & $\times$ & $\times$ & $\times$ & $\checkmark$ & $\checkmark$ &  \cellcolor[HTML]{FFFFED} 0.014 &  \cellcolor[HTML]{FFFFED} 0.209  \\
& 54 & $\times$ & $\times$ & $\checkmark$ & $\checkmark$ & $\checkmark$ & $\times$ & $\times$ &  \cellcolor[HTML]{FFFFED} 0.014 &  \cellcolor[HTML]{FFFFED} 0.217  \\
& 55 & $\times$ & $\times$ & $\checkmark$ & $\checkmark$ & $\times$ & $\checkmark$ & $\times$ &  \cellcolor[HTML]{FFFFED} 0.014 &  \cellcolor[HTML]{FFFFED} 0.209  \\
& 56 & $\times$ & $\times$ & $\checkmark$ & $\checkmark$ & $\times$ & $\times$ & $\checkmark$ &  \cellcolor[HTML]{FFFFED} 0.043 &  \cellcolor[HTML]{FFFFED} 0.191  \\
& 57 & $\times$ & $\times$ & $\checkmark$ & $\times$ & $\checkmark$ & $\checkmark$ & $\times$ &  \cellcolor[HTML]{FFFFED} 0.014 &  \cellcolor[HTML]{FFFFED} 0.191  \\
& 58 & $\times$ & $\times$ & $\checkmark$ & $\times$ & $\checkmark$ & $\times$ & $\checkmark$ &  \cellcolor[HTML]{FFFFED} 0.014 &  \cellcolor[HTML]{FFFFED} 0.200  \\
& 59 & $\times$ & $\times$ & $\checkmark$ & $\times$ & $\times$ & $\checkmark$ & $\checkmark$ &  \cellcolor[HTML]{FFFFED} 0.014 &  \cellcolor[HTML]{FFFFED} 0.200  \\
& 60 & $\times$ & $\times$ & $\times$ & $\checkmark$ & $\checkmark$ & $\checkmark$ & $\times$ &  \cellcolor[HTML]{FFFFED} 0.014 &  \cellcolor[HTML]{FFFFED} 0.200  \\
& 61 & $\times$ & $\times$ & $\times$ & $\checkmark$ & $\checkmark$ & $\times$ & $\checkmark$ &  \cellcolor[HTML]{FFFFED} 0.000 &  \cellcolor[HTML]{FFFFED} 0.165  \\
& 62 & $\times$ & $\times$ & $\times$ & $\checkmark$ & $\times$ & $\checkmark$ & $\checkmark$ &  \cellcolor[HTML]{FFFFED} 0.014 &  \cellcolor[HTML]{FFFFED} 0.191  \\
& 63 & $\times$ & $\times$ & $\times$ & $\times$ & $\checkmark$ & $\checkmark$ & $\checkmark$ &  \cellcolor[HTML]{FFFFED} 0.014 &  \cellcolor[HTML]{FFFFED} 0.191  \\
& 64 & $\checkmark$ & $\checkmark$ & $\checkmark$ & $\checkmark$ & $\times$ & $\times$ & $\times$ &  \cellcolor[HTML]{FFFFED} 0.151 &  \cellcolor[HTML]{FFFFED} 0.452  \\
\bottomrule
\end{tabular}

\hspace{0.2in}

\begin{tabular}{c|c|c|cc|cc|cc|cc@{}}
\toprule
&  & \multicolumn{7}{c|}{Scores} & \multicolumn{2}{c}{Metrics} \\
\midrule 
 & Row ID & $a_{\text{IN}}$ & text-f1 & text-acc1 &  \superclassSymbl  & \interSimSymbl & \intraSimSymbl & \intraClassCloseSymbl & $\tau$ ($\uparrow$) & $R_5$ ($\uparrow$) \\
 \midrule 

& 65 & $\checkmark$ & $\checkmark$ & $\checkmark$ & $\times$ & $\checkmark$ & $\times$ & $\times$ &  \cellcolor[HTML]{FFFFED} 0.107 &  \cellcolor[HTML]{FFFFED} 0.443  \\
& 66 & $\checkmark$ & $\checkmark$ & $\checkmark$ & $\times$ & $\times$ & $\checkmark$ & $\times$ &  \cellcolor[HTML]{FFFFED} 0.093 &  \cellcolor[HTML]{FFFFED} 0.435  \\
& 67 & $\checkmark$ & $\checkmark$ & $\checkmark$ & $\times$ & $\times$ & $\times$ & $\checkmark$ &  \cellcolor[HTML]{FFFFED} 0.133 &  \cellcolor[HTML]{FFFFED} 0.443  \\
& 68 & $\checkmark$ & $\checkmark$ & $\times$ & $\checkmark$ & $\checkmark$ & $\times$ & $\times$ &  \cellcolor[HTML]{FFFFED} 0.174 &  \cellcolor[HTML]{FFFFED} 0.443  \\
& 69 & $\checkmark$ & $\checkmark$ & $\times$ & $\checkmark$ & $\times$ & $\checkmark$ & $\times$ &  \cellcolor[HTML]{FFFFED} 0.157 &  \cellcolor[HTML]{FFFFED} 0.443  \\
& 70 & $\checkmark$ & $\checkmark$ & $\times$ & $\checkmark$ & $\times$ & $\times$ & $\checkmark$ &  \cellcolor[HTML]{FFFFED} 0.075 &  \cellcolor[HTML]{FFFFED} 0.417  \\
& 71 & $\checkmark$ & $\checkmark$ & $\times$ & $\times$ & $\checkmark$ & $\checkmark$ & $\times$ &  \cellcolor[HTML]{FFFFED} 0.113 &  \cellcolor[HTML]{FFFFED} 0.452  \\
& 72 & $\checkmark$ & $\checkmark$ & $\times$ & $\times$ & $\checkmark$ & $\times$ & $\checkmark$ &  \cellcolor[HTML]{FFFFED} 0.096 &  \cellcolor[HTML]{FFFFED} 0.417  \\
& 73 & $\checkmark$ & $\checkmark$ & $\times$ & $\times$ & $\times$ & $\checkmark$ & $\checkmark$ &  \cellcolor[HTML]{FFFFED} 0.096 &  \cellcolor[HTML]{FFFFED} 0.417  \\
& 74 & $\checkmark$ & $\times$ & $\checkmark$ & $\checkmark$ & $\checkmark$ & $\times$ & $\times$ &  \cellcolor[HTML]{FFFFED} 0.159 &  \cellcolor[HTML]{FFFFED} 0.435  \\
& 75 & $\checkmark$ & $\times$ & $\checkmark$ & $\checkmark$ & $\times$ & $\checkmark$ & $\times$ &  \cellcolor[HTML]{FFFFED} 0.128 &  \cellcolor[HTML]{FFFFED} 0.443  \\
& 76 & $\checkmark$ & $\times$ & $\checkmark$ & $\checkmark$ & $\times$ & $\times$ & $\checkmark$ &  \cellcolor[HTML]{FFFFED} 0.075 &  \cellcolor[HTML]{FFFFED} 0.417  \\
& 77 & $\checkmark$ & $\times$ & $\checkmark$ & $\times$ & $\checkmark$ & $\checkmark$ & $\times$ &  \cellcolor[HTML]{FFFFED} 0.113 &  \cellcolor[HTML]{FFFFED} 0.452  \\
& 78 & $\checkmark$ & $\times$ & $\checkmark$ & $\times$ & $\checkmark$ & $\times$ & $\checkmark$ &  \cellcolor[HTML]{FFFFED} 0.096 &  \cellcolor[HTML]{FFFFED} 0.417  \\
& 79 & $\checkmark$ & $\times$ & $\checkmark$ & $\times$ & $\times$ & $\checkmark$ & $\checkmark$ &  \cellcolor[HTML]{FFFFED} 0.096 &  \cellcolor[HTML]{FFFFED} 0.417  \\
& 80 & $\checkmark$ & $\times$ & $\times$ & $\checkmark$ & $\checkmark$ & $\checkmark$ & $\times$ &  \cellcolor[HTML]{FFFFED} 0.043 &  \cellcolor[HTML]{FFFFED} 0.417  \\
& 81 & $\checkmark$ & $\times$ & $\times$ & $\checkmark$ & $\checkmark$ & $\times$ & $\checkmark$ &  \cellcolor[HTML]{FFFFED} 0.119 &  \cellcolor[HTML]{FFFFED} 0.417  \\
& 82 & $\checkmark$ & $\times$ & $\times$ & $\checkmark$ & $\times$ & $\checkmark$ & $\checkmark$ &  \cellcolor[HTML]{FFFFED} 0.119 &  \cellcolor[HTML]{FFFFED} 0.417  \\
& 83 & $\checkmark$ & $\times$ & $\times$ & $\times$ & $\checkmark$ & $\checkmark$ & $\checkmark$ &  \cellcolor[HTML]{FFFFED} 0.110 &  \cellcolor[HTML]{FFFFED} 0.417  \\
& 84 & $\times$ & $\checkmark$ & $\checkmark$ & $\checkmark$ & $\checkmark$ & $\times$ & $\times$ &  \cellcolor[HTML]{FFFFED} -0.029 &  \cellcolor[HTML]{FFFFED} 0.209  \\
& 85 & $\times$ & $\checkmark$ & $\checkmark$ & $\checkmark$ & $\times$ & $\checkmark$ & $\times$ &  \cellcolor[HTML]{FFFFED} -0.029 &  \cellcolor[HTML]{FFFFED} 0.217  \\
& 86 & $\times$ & $\checkmark$ & $\checkmark$ & $\checkmark$ & $\times$ & $\times$ & $\checkmark$ &  \cellcolor[HTML]{FFFFED} -0.014 &  \cellcolor[HTML]{FFFFED} 0.217  \\
& 87 & $\times$ & $\checkmark$ & $\checkmark$ & $\times$ & $\checkmark$ & $\checkmark$ & $\times$ &  \cellcolor[HTML]{FFFFED} 0.014 &  \cellcolor[HTML]{FFFFED} 0.200  \\
& 88 & $\times$ & $\checkmark$ & $\checkmark$ & $\times$ & $\checkmark$ & $\times$ & $\checkmark$ &  \cellcolor[HTML]{FFFFED} 0.014 &  \cellcolor[HTML]{FFFFED} 0.226  \\
& 89 & $\times$ & $\checkmark$ & $\checkmark$ & $\times$ & $\times$ & $\checkmark$ & $\checkmark$ &  \cellcolor[HTML]{FFFFED} 0.014 &  \cellcolor[HTML]{FFFFED} 0.217  \\
& 90 & $\times$ & $\checkmark$ & $\times$ & $\checkmark$ & $\checkmark$ & $\checkmark$ & $\times$ &  \cellcolor[HTML]{FFFFED} 0.014 &  \cellcolor[HTML]{FFFFED} 0.217  \\
& 91 & $\times$ & $\checkmark$ & $\times$ & $\checkmark$ & $\checkmark$ & $\times$ & $\checkmark$ &  \cellcolor[HTML]{FFFFED} 0.000 &  \cellcolor[HTML]{FFFFED} 0.226  \\
& 92 & $\times$ & $\checkmark$ & $\times$ & $\checkmark$ & $\times$ & $\checkmark$ & $\checkmark$ &  \cellcolor[HTML]{FFFFED} 0.014 &  \cellcolor[HTML]{FFFFED} 0.209  \\
& 93 & $\times$ & $\checkmark$ & $\times$ & $\times$ & $\checkmark$ & $\checkmark$ & $\checkmark$ &  \cellcolor[HTML]{FFFFED} 0.014 &  \cellcolor[HTML]{FFFFED} 0.209  \\
& 94 & $\times$ & $\times$ & $\checkmark$ & $\checkmark$ & $\checkmark$ & $\checkmark$ & $\times$ &  \cellcolor[HTML]{FFFFED} 0.014 &  \cellcolor[HTML]{FFFFED} 0.209  \\
& 95 & $\times$ & $\times$ & $\checkmark$ & $\checkmark$ & $\checkmark$ & $\times$ & $\checkmark$ &  \cellcolor[HTML]{FFFFED} 0.014 &  \cellcolor[HTML]{FFFFED} 0.217  \\
& 96 & $\times$ & $\times$ & $\checkmark$ & $\checkmark$ & $\times$ & $\checkmark$ & $\checkmark$ &  \cellcolor[HTML]{FFFFED} 0.014 &  \cellcolor[HTML]{FFFFED} 0.209  \\
& 97 & $\times$ & $\times$ & $\checkmark$ & $\times$ & $\checkmark$ & $\checkmark$ & $\checkmark$ &  \cellcolor[HTML]{FFFFED} 0.014 &  \cellcolor[HTML]{FFFFED} 0.200  \\
& 98 & $\times$ & $\times$ & $\times$ & $\checkmark$ & $\checkmark$ & $\checkmark$ & $\checkmark$ &  \cellcolor[HTML]{FFFFED} 0.014 &  \cellcolor[HTML]{FFFFED} 0.226  \\
& 99 & $\checkmark$ & $\checkmark$ & $\checkmark$ & $\checkmark$ & $\checkmark$ & $\times$ & $\times$ &  \cellcolor[HTML]{FFFFED} 0.116 &  \cellcolor[HTML]{FFFFED} 0.435  \\
& 100 & $\checkmark$ & $\checkmark$ & $\checkmark$ & $\checkmark$ & $\times$ & $\checkmark$ & $\times$ &  \cellcolor[HTML]{FFFFED} 0.058 &  \cellcolor[HTML]{FFFFED} 0.435  \\
& 101 & $\checkmark$ & $\checkmark$ & $\checkmark$ & $\checkmark$ & $\times$ & $\times$ & $\checkmark$ &  \cellcolor[HTML]{FFFFED} 0.133 &  \cellcolor[HTML]{FFFFED} 0.443  \\
& 102 & $\checkmark$ & $\checkmark$ & $\checkmark$ & $\times$ & $\checkmark$ & $\checkmark$ & $\times$ &  \cellcolor[HTML]{FFFFED} 0.049 &  \cellcolor[HTML]{FFFFED} 0.443  \\
& 103 & $\checkmark$ & $\checkmark$ & $\checkmark$ & $\times$ & $\checkmark$ & $\times$ & $\checkmark$ &  \cellcolor[HTML]{FFFFED} 0.133 &  \cellcolor[HTML]{FFFFED} 0.443  \\
& 104 & $\checkmark$ & $\checkmark$ & $\checkmark$ & $\times$ & $\times$ & $\checkmark$ & $\checkmark$ &  \cellcolor[HTML]{FFFFED} 0.133 &  \cellcolor[HTML]{FFFFED} 0.443  \\
& 105 & $\checkmark$ & $\checkmark$ & $\times$ & $\checkmark$ & $\checkmark$ & $\checkmark$ & $\times$ &  \cellcolor[HTML]{FFFFED} 0.128 &  \cellcolor[HTML]{FFFFED} 0.443  \\
& 106 & $\checkmark$ & $\checkmark$ & $\times$ & $\checkmark$ & $\checkmark$ & $\times$ & $\checkmark$ &  \cellcolor[HTML]{FFFFED} 0.104 &  \cellcolor[HTML]{FFFFED} 0.417  \\
& 107 & $\checkmark$ & $\checkmark$ & $\times$ & $\checkmark$ & $\times$ & $\checkmark$ & $\checkmark$ &  \cellcolor[HTML]{FFFFED} 0.104 &  \cellcolor[HTML]{FFFFED} 0.417  \\
& 108 & $\checkmark$ & $\checkmark$ & $\times$ & $\times$ & $\checkmark$ & $\checkmark$ & $\checkmark$ &  \cellcolor[HTML]{FFFFED} 0.128 &  \cellcolor[HTML]{FFFFED} 0.435  \\
& 109 & $\checkmark$ & $\times$ & $\checkmark$ & $\checkmark$ & $\checkmark$ & $\checkmark$ & $\times$ &  \cellcolor[HTML]{FFFFED} 0.128 &  \cellcolor[HTML]{FFFFED} 0.443  \\
& 110 & $\checkmark$ & $\times$ & $\checkmark$ & $\checkmark$ & $\checkmark$ & $\times$ & $\checkmark$ &  \cellcolor[HTML]{FFFFED} 0.104 &  \cellcolor[HTML]{FFFFED} 0.417  \\
& 111 & $\checkmark$ & $\times$ & $\checkmark$ & $\checkmark$ & $\times$ & $\checkmark$ & $\checkmark$ &  \cellcolor[HTML]{FFFFED} 0.104 &  \cellcolor[HTML]{FFFFED} 0.417  \\
& 112 & $\checkmark$ & $\times$ & $\checkmark$ & $\times$ & $\checkmark$ & $\checkmark$ & $\checkmark$ &  \cellcolor[HTML]{FFFFED} 0.128 &  \cellcolor[HTML]{FFFFED} 0.435  \\
& 113 & $\checkmark$ & $\times$ & $\times$ & $\checkmark$ & $\checkmark$ & $\checkmark$ & $\checkmark$ &  \cellcolor[HTML]{FFFFED} 0.101 &  \cellcolor[HTML]{FFFFED} 0.400  \\
& 114 & $\times$ & $\checkmark$ & $\checkmark$ & $\checkmark$ & $\checkmark$ & $\checkmark$ & $\times$ &  \cellcolor[HTML]{FFFFED} -0.029 &  \cellcolor[HTML]{FFFFED} 0.217  \\
& 115 & $\times$ & $\checkmark$ & $\checkmark$ & $\checkmark$ & $\checkmark$ & $\times$ & $\checkmark$ &  \cellcolor[HTML]{FFFFED} -0.043 &  \cellcolor[HTML]{FFFFED} 0.209  \\
& 116 & $\times$ & $\checkmark$ & $\checkmark$ & $\checkmark$ & $\times$ & $\checkmark$ & $\checkmark$ &  \cellcolor[HTML]{FFFFED} -0.029 &  \cellcolor[HTML]{FFFFED} 0.217  \\
& 117 & $\times$ & $\checkmark$ & $\checkmark$ & $\times$ & $\checkmark$ & $\checkmark$ & $\checkmark$ &  \cellcolor[HTML]{FFFFED} 0.014 &  \cellcolor[HTML]{FFFFED} 0.217  \\
& 118 & $\times$ & $\checkmark$ & $\times$ & $\checkmark$ & $\checkmark$ & $\checkmark$ & $\checkmark$ &  \cellcolor[HTML]{FFFFED} 0.014 &  \cellcolor[HTML]{FFFFED} 0.209  \\
& 119 & $\times$ & $\times$ & $\checkmark$ & $\checkmark$ & $\checkmark$ & $\checkmark$ & $\checkmark$ &  \cellcolor[HTML]{FFFFED} 0.014 &  \cellcolor[HTML]{FFFFED} 0.209  \\
& 120 & $\checkmark$ & $\checkmark$ & $\checkmark$ & $\checkmark$ & $\checkmark$ & $\checkmark$ & $\times$ &  \cellcolor[HTML]{FFFFED} 0.055 &  \cellcolor[HTML]{FFFFED} 0.443  \\
& 121 & $\checkmark$ & $\checkmark$ & $\checkmark$ & $\checkmark$ & $\checkmark$ & $\times$ & $\checkmark$ &  \cellcolor[HTML]{FFFFED} 0.090 &  \cellcolor[HTML]{FFFFED} 0.435  \\
& 122 & $\checkmark$ & $\checkmark$ & $\checkmark$ & $\checkmark$ & $\times$ & $\checkmark$ & $\checkmark$ &  \cellcolor[HTML]{FFFFED} 0.090 &  \cellcolor[HTML]{FFFFED} 0.435  \\
& 123 & $\checkmark$ & $\checkmark$ & $\checkmark$ & $\times$ & $\checkmark$ & $\checkmark$ & $\checkmark$ &  \cellcolor[HTML]{FFFFED} 0.090 &  \cellcolor[HTML]{FFFFED} 0.443  \\
& 124 & $\checkmark$ & $\checkmark$ & $\times$ & $\checkmark$ & $\checkmark$ & $\checkmark$ & $\checkmark$ &  \cellcolor[HTML]{FFFFED} 0.128 &  \cellcolor[HTML]{FFFFED} 0.435  \\
& 125 & $\checkmark$ & $\times$ & $\checkmark$ & $\checkmark$ & $\checkmark$ & $\checkmark$ & $\checkmark$ &  \cellcolor[HTML]{FFFFED} 0.128 &  \cellcolor[HTML]{FFFFED} 0.435  \\
& 126 & $\times$ & $\checkmark$ & $\checkmark$ & $\checkmark$ & $\checkmark$ & $\checkmark$ & $\checkmark$ &  \cellcolor[HTML]{FFFFED} -0.029 &  \cellcolor[HTML]{FFFFED} 0.217  \\
& 127 & $\checkmark$ & $\checkmark$ & $\checkmark$ & $\checkmark$ & $\checkmark$ & $\checkmark$ & $\checkmark$ &  \cellcolor[HTML]{FFFFED} 0.046 &  \cellcolor[HTML]{FFFFED} 0.435  \\
\bottomrule
\end{tabular}
}

\end{multicols}

\end{table}

\begin{table}
\caption{\textbf{LOVM Model Prediction Ablation.} Here, we ablate all the different scores used in our baselines for predicting model top-1 accuracy. We separated the text classification (C) base scores, and the granularity-based scores that quantify inter- and intra-class similarity. \label{tab:ablation:pred_acc}
$a_{\text{IN}}$ - Imagenet Accuracy,  \intraClassCloseSymbl\ - \intraClassClose, text-f1 - caption dataset f1-score, text-acc1 - text top-1 accuracy, \superclassSymbl\ - \superclass, \intraSimSymbl\ - \intraSim, \interSimSymbl\ - \interSim.
}
\begin{multicols}{2}

\resizebox{1\textwidth}{!}{
\begin{tabular}{c|c|c|cc|cc|cc|c@{}}
\toprule
&  & \multicolumn{7}{c|}{Scores} & \multicolumn{1}{c}{Metrics} \\
\midrule 
 & Row ID & $a_{\text{IN}}$ & text-f1 & text-acc1 &  \superclassSymbl  & \interSimSymbl & \intraSimSymbl & \intraClassCloseSymbl & $L_1$ ($\downarrow$) ) \\
\midrule
\multirow{64}{*}{\rotatebox[origin=c]{90}{\parbox[c]{4cm}{\centering \textbf{Model Prediction}}}} 

& 1 & $\checkmark$ & $\times$ & $\times$ & $\times$ & $\times$ & $\times$ & $\times$ &  \cellcolor[HTML]{FFFFED} 0.220  \\
& 2 & $\times$ & $\checkmark$ & $\times$ & $\times$ & $\times$ & $\times$ & $\times$ &  \cellcolor[HTML]{FFFFED} 0.176  \\
& 3 & $\times$ & $\times$ & $\checkmark$ & $\times$ & $\times$ & $\times$ & $\times$ &  \cellcolor[HTML]{FFFFED} 0.177  \\
& 4 & $\times$ & $\times$ & $\times$ & $\checkmark$ & $\times$ & $\times$ & $\times$ &  \cellcolor[HTML]{FFFFED} 0.188  \\
& 5 & $\times$ & $\times$ & $\times$ & $\times$ & $\checkmark$ & $\times$ & $\times$ &  \cellcolor[HTML]{FFFFED} 0.200  \\
& 6 & $\times$ & $\times$ & $\times$ & $\times$ & $\times$ & $\checkmark$ & $\times$ &  \cellcolor[HTML]{FFFFED} 0.223  \\
& 7 & $\times$ & $\times$ & $\times$ & $\times$ & $\times$ & $\times$ & $\checkmark$ &  \cellcolor[HTML]{FFFFED} 0.170  \\
& 8 & $\checkmark$ & $\checkmark$ & $\times$ & $\times$ & $\times$ & $\times$ & $\times$ &  \cellcolor[HTML]{FFFFED} 0.189  \\
& 9 & $\checkmark$ & $\times$ & $\checkmark$ & $\times$ & $\times$ & $\times$ & $\times$ &  \cellcolor[HTML]{FFFFED} 0.184  \\
& 10 & $\checkmark$ & $\times$ & $\times$ & $\checkmark$ & $\times$ & $\times$ & $\times$ &  \cellcolor[HTML]{FFFFED} 0.215  \\
& 11 & $\checkmark$ & $\times$ & $\times$ & $\times$ & $\checkmark$ & $\times$ & $\times$ &  \cellcolor[HTML]{FFFFED} 0.205  \\
& 12 & $\checkmark$ & $\times$ & $\times$ & $\times$ & $\times$ & $\checkmark$ & $\times$ &  \cellcolor[HTML]{FFFFED} 0.215  \\
& 13 & $\checkmark$ & $\times$ & $\times$ & $\times$ & $\times$ & $\times$ & $\checkmark$ &  \cellcolor[HTML]{FFFFED} 0.167  \\
& 14 & $\times$ & $\checkmark$ & $\checkmark$ & $\times$ & $\times$ & $\times$ & $\times$ &  \cellcolor[HTML]{FFFFED} 0.180  \\
& 15 & $\times$ & $\checkmark$ & $\times$ & $\checkmark$ & $\times$ & $\times$ & $\times$ &  \cellcolor[HTML]{FFFFED} 0.190  \\
& 16 & $\times$ & $\checkmark$ & $\times$ & $\times$ & $\checkmark$ & $\times$ & $\times$ &  \cellcolor[HTML]{FFFFED} 0.184  \\
& 17 & $\times$ & $\checkmark$ & $\times$ & $\times$ & $\times$ & $\checkmark$ & $\times$ &  \cellcolor[HTML]{FFFFED} 0.180  \\
& 18 & $\times$ & $\checkmark$ & $\times$ & $\times$ & $\times$ & $\times$ & $\checkmark$ &  \cellcolor[HTML]{FFFFED} 0.160  \\
& 19 & $\times$ & $\times$ & $\checkmark$ & $\checkmark$ & $\times$ & $\times$ & $\times$ &  \cellcolor[HTML]{FFFFED} 0.183  \\
& 20 & $\times$ & $\times$ & $\checkmark$ & $\times$ & $\checkmark$ & $\times$ & $\times$ &  \cellcolor[HTML]{FFFFED} 0.176  \\
& 21 & $\times$ & $\times$ & $\checkmark$ & $\times$ & $\times$ & $\checkmark$ & $\times$ &  \cellcolor[HTML]{FFFFED} 0.179  \\
& 22 & $\times$ & $\times$ & $\checkmark$ & $\times$ & $\times$ & $\times$ & $\checkmark$ &  \cellcolor[HTML]{FFFFED} 0.159  \\
& 23 & $\times$ & $\times$ & $\times$ & $\checkmark$ & $\checkmark$ & $\times$ & $\times$ &  \cellcolor[HTML]{FFFFED} 0.199  \\
& 24 & $\times$ & $\times$ & $\times$ & $\checkmark$ & $\times$ & $\checkmark$ & $\times$ &  \cellcolor[HTML]{FFFFED} 0.188  \\
& 25 & $\times$ & $\times$ & $\times$ & $\checkmark$ & $\times$ & $\times$ & $\checkmark$ &  \cellcolor[HTML]{FFFFED} 0.154  \\
& 26 & $\times$ & $\times$ & $\times$ & $\times$ & $\checkmark$ & $\checkmark$ & $\times$ &  \cellcolor[HTML]{FFFFED} 0.143  \\
& 27 & $\times$ & $\times$ & $\times$ & $\times$ & $\checkmark$ & $\times$ & $\checkmark$ &  \cellcolor[HTML]{FFFFED} 0.163  \\
& 28 & $\times$ & $\times$ & $\times$ & $\times$ & $\times$ & $\checkmark$ & $\checkmark$ &  \cellcolor[HTML]{FFFFED} 0.166  \\
& 29 & $\checkmark$ & $\checkmark$ & $\checkmark$ & $\times$ & $\times$ & $\times$ & $\times$ &  \cellcolor[HTML]{FFFFED} 0.191  \\
& 30 & $\checkmark$ & $\checkmark$ & $\times$ & $\checkmark$ & $\times$ & $\times$ & $\times$ &  \cellcolor[HTML]{FFFFED} 0.197  \\
& 31 & $\checkmark$ & $\checkmark$ & $\times$ & $\times$ & $\checkmark$ & $\times$ & $\times$ &  \cellcolor[HTML]{FFFFED} 0.189  \\
& 32 & $\checkmark$ & $\checkmark$ & $\times$ & $\times$ & $\times$ & $\checkmark$ & $\times$ &  \cellcolor[HTML]{FFFFED} 0.187  \\
& 33 & $\checkmark$ & $\checkmark$ & $\times$ & $\times$ & $\times$ & $\times$ & $\checkmark$ &  \cellcolor[HTML]{FFFFED} 0.159  \\
& 34 & $\checkmark$ & $\times$ & $\checkmark$ & $\checkmark$ & $\times$ & $\times$ & $\times$ &  \cellcolor[HTML]{FFFFED} 0.196  \\
& 35 & $\checkmark$ & $\times$ & $\checkmark$ & $\times$ & $\checkmark$ & $\times$ & $\times$ &  \cellcolor[HTML]{FFFFED} 0.189  \\
& 36 & $\checkmark$ & $\times$ & $\checkmark$ & $\times$ & $\times$ & $\checkmark$ & $\times$ &  \cellcolor[HTML]{FFFFED} 0.185  \\
& 37 & $\checkmark$ & $\times$ & $\checkmark$ & $\times$ & $\times$ & $\times$ & $\checkmark$ &  \cellcolor[HTML]{FFFFED} 0.155  \\
& 38 & $\checkmark$ & $\times$ & $\times$ & $\checkmark$ & $\checkmark$ & $\times$ & $\times$ &  \cellcolor[HTML]{FFFFED} 0.212  \\
& 39 & $\checkmark$ & $\times$ & $\times$ & $\checkmark$ & $\times$ & $\checkmark$ & $\times$ &  \cellcolor[HTML]{FFFFED} 0.219  \\
& 40 & $\checkmark$ & $\times$ & $\times$ & $\checkmark$ & $\times$ & $\times$ & $\checkmark$ &  \cellcolor[HTML]{FFFFED} 0.156  \\
& 41 & $\checkmark$ & $\times$ & $\times$ & $\times$ & $\checkmark$ & $\checkmark$ & $\times$ &  \cellcolor[HTML]{FFFFED} 0.141  \\
& 42 & $\checkmark$ & $\times$ & $\times$ & $\times$ & $\checkmark$ & $\times$ & $\checkmark$ &  \cellcolor[HTML]{FFFFED} 0.160  \\
& 43 & $\checkmark$ & $\times$ & $\times$ & $\times$ & $\times$ & $\checkmark$ & $\checkmark$ &  \cellcolor[HTML]{FFFFED} 0.167  \\
& 44 & $\times$ & $\checkmark$ & $\checkmark$ & $\checkmark$ & $\times$ & $\times$ & $\times$ &  \cellcolor[HTML]{FFFFED} 0.184  \\
& 45 & $\times$ & $\checkmark$ & $\checkmark$ & $\times$ & $\checkmark$ & $\times$ & $\times$ &  \cellcolor[HTML]{FFFFED} 0.187  \\
& 46 & $\times$ & $\checkmark$ & $\checkmark$ & $\times$ & $\times$ & $\checkmark$ & $\times$ &  \cellcolor[HTML]{FFFFED} 0.186  \\
& 47 & $\times$ & $\checkmark$ & $\checkmark$ & $\times$ & $\times$ & $\times$ & $\checkmark$ &  \cellcolor[HTML]{FFFFED} 0.170  \\
& 48 & $\times$ & $\checkmark$ & $\times$ & $\checkmark$ & $\checkmark$ & $\times$ & $\times$ &  \cellcolor[HTML]{FFFFED} 0.190  \\
& 49 & $\times$ & $\checkmark$ & $\times$ & $\checkmark$ & $\times$ & $\checkmark$ & $\times$ &  \cellcolor[HTML]{FFFFED} 0.189  \\
& 50 & $\times$ & $\checkmark$ & $\times$ & $\checkmark$ & $\times$ & $\times$ & $\checkmark$ &  \cellcolor[HTML]{FFFFED} 0.160  \\
& 51 & $\times$ & $\checkmark$ & $\times$ & $\times$ & $\checkmark$ & $\checkmark$ & $\times$ &  \cellcolor[HTML]{FFFFED} 0.148  \\
& 52 & $\times$ & $\checkmark$ & $\times$ & $\times$ & $\checkmark$ & $\times$ & $\checkmark$ &  \cellcolor[HTML]{FFFFED} 0.163  \\
& 53 & $\times$ & $\checkmark$ & $\times$ & $\times$ & $\times$ & $\checkmark$ & $\checkmark$ &  \cellcolor[HTML]{FFFFED} 0.167  \\
& 54 & $\times$ & $\times$ & $\checkmark$ & $\checkmark$ & $\checkmark$ & $\times$ & $\times$ &  \cellcolor[HTML]{FFFFED} 0.183  \\
& 55 & $\times$ & $\times$ & $\checkmark$ & $\checkmark$ & $\times$ & $\checkmark$ & $\times$ &  \cellcolor[HTML]{FFFFED} 0.187  \\
& 56 & $\times$ & $\times$ & $\checkmark$ & $\checkmark$ & $\times$ & $\times$ & $\checkmark$ &  \cellcolor[HTML]{FFFFED} 0.159  \\
& 57 & $\times$ & $\times$ & $\checkmark$ & $\times$ & $\checkmark$ & $\checkmark$ & $\times$ &  \cellcolor[HTML]{FFFFED} 0.147  \\
& 58 & $\times$ & $\times$ & $\checkmark$ & $\times$ & $\checkmark$ & $\times$ & $\checkmark$ &  \cellcolor[HTML]{FFFFED} 0.159  \\
& 59 & $\times$ & $\times$ & $\checkmark$ & $\times$ & $\times$ & $\checkmark$ & $\checkmark$ &  \cellcolor[HTML]{FFFFED} 0.167  \\
& 60 & $\times$ & $\times$ & $\times$ & $\checkmark$ & $\checkmark$ & $\checkmark$ & $\times$ &  \cellcolor[HTML]{FFFFED} 0.148  \\
& 61 & $\times$ & $\times$ & $\times$ & $\checkmark$ & $\checkmark$ & $\times$ & $\checkmark$ &  \cellcolor[HTML]{FFFFED} 0.154  \\
& 62 & $\times$ & $\times$ & $\times$ & $\checkmark$ & $\times$ & $\checkmark$ & $\checkmark$ &  \cellcolor[HTML]{FFFFED} 0.161  \\
& 63 & $\times$ & $\times$ & $\times$ & $\times$ & $\checkmark$ & $\checkmark$ & $\checkmark$ &  \cellcolor[HTML]{FFFFED} 0.153  \\
& 64 & $\checkmark$ & $\checkmark$ & $\checkmark$ & $\checkmark$ & $\times$ & $\times$ & $\times$ &  \cellcolor[HTML]{FFFFED} 0.200  \\
\bottomrule
\end{tabular}

\hspace{0.2in}

\begin{tabular}{c|c|c|cc|cc|cc|c@{}}
\toprule
&  & \multicolumn{7}{c|}{Scores} & \multicolumn{1}{c}{Metrics} \\
\midrule 
 & Row ID & $a_{\text{IN}}$ & text-f1 & text-acc1 &  \superclassSymbl  & \interSimSymbl & \intraSimSymbl & \intraClassCloseSymbl & $L_1$ ($\downarrow$) ) \\
 \midrule 
& 65 & $\checkmark$ & $\checkmark$ & $\checkmark$ & $\times$ & $\checkmark$ & $\times$ & $\times$ &  \cellcolor[HTML]{FFFFED} 0.191  \\
& 66 & $\checkmark$ & $\checkmark$ & $\checkmark$ & $\times$ & $\times$ & $\checkmark$ & $\times$ &  \cellcolor[HTML]{FFFFED} 0.191  \\
& 67 & $\checkmark$ & $\checkmark$ & $\checkmark$ & $\times$ & $\times$ & $\times$ & $\checkmark$ &  \cellcolor[HTML]{FFFFED} 0.167  \\
& 68 & $\checkmark$ & $\checkmark$ & $\times$ & $\checkmark$ & $\checkmark$ & $\times$ & $\times$ &  \cellcolor[HTML]{FFFFED} 0.198  \\
& 69 & $\checkmark$ & $\checkmark$ & $\times$ & $\checkmark$ & $\times$ & $\checkmark$ & $\times$ &  \cellcolor[HTML]{FFFFED} 0.200  \\
& 70 & $\checkmark$ & $\checkmark$ & $\times$ & $\checkmark$ & $\times$ & $\times$ & $\checkmark$ &  \cellcolor[HTML]{FFFFED} 0.160  \\
& 71 & $\checkmark$ & $\checkmark$ & $\times$ & $\times$ & $\checkmark$ & $\checkmark$ & $\times$ &  \cellcolor[HTML]{FFFFED} 0.151  \\
& 72 & $\checkmark$ & $\checkmark$ & $\times$ & $\times$ & $\checkmark$ & $\times$ & $\checkmark$ &  \cellcolor[HTML]{FFFFED} 0.158  \\
& 73 & $\checkmark$ & $\checkmark$ & $\times$ & $\times$ & $\times$ & $\checkmark$ & $\checkmark$ &  \cellcolor[HTML]{FFFFED} 0.163  \\
& 74 & $\checkmark$ & $\times$ & $\checkmark$ & $\checkmark$ & $\checkmark$ & $\times$ & $\times$ &  \cellcolor[HTML]{FFFFED} 0.198  \\
& 75 & $\checkmark$ & $\times$ & $\checkmark$ & $\checkmark$ & $\times$ & $\checkmark$ & $\times$ &  \cellcolor[HTML]{FFFFED} 0.198  \\
& 76 & $\checkmark$ & $\times$ & $\checkmark$ & $\checkmark$ & $\times$ & $\times$ & $\checkmark$ &  \cellcolor[HTML]{FFFFED} 0.158  \\
& 77 & $\checkmark$ & $\times$ & $\checkmark$ & $\times$ & $\checkmark$ & $\checkmark$ & $\times$ &  \cellcolor[HTML]{FFFFED} 0.156  \\
& 78 & $\checkmark$ & $\times$ & $\checkmark$ & $\times$ & $\checkmark$ & $\times$ & $\checkmark$ &  \cellcolor[HTML]{FFFFED} 0.154  \\
& 79 & $\checkmark$ & $\times$ & $\checkmark$ & $\times$ & $\times$ & $\checkmark$ & $\checkmark$ &  \cellcolor[HTML]{FFFFED} 0.162  \\
& 80 & $\checkmark$ & $\times$ & $\times$ & $\checkmark$ & $\checkmark$ & $\checkmark$ & $\times$ &  \cellcolor[HTML]{FFFFED} 0.144  \\
& 81 & $\checkmark$ & $\times$ & $\times$ & $\checkmark$ & $\checkmark$ & $\times$ & $\checkmark$ &  \cellcolor[HTML]{FFFFED} 0.158  \\
& 82 & $\checkmark$ & $\times$ & $\times$ & $\checkmark$ & $\times$ & $\checkmark$ & $\checkmark$ &  \cellcolor[HTML]{FFFFED} 0.164  \\
& 83 & $\checkmark$ & $\times$ & $\times$ & $\times$ & $\checkmark$ & $\checkmark$ & $\checkmark$ &  \cellcolor[HTML]{FFFFED} 0.149  \\
& 84 & $\times$ & $\checkmark$ & $\checkmark$ & $\checkmark$ & $\checkmark$ & $\times$ & $\times$ &  \cellcolor[HTML]{FFFFED} 0.186  \\
& 85 & $\times$ & $\checkmark$ & $\checkmark$ & $\checkmark$ & $\times$ & $\checkmark$ & $\times$ &  \cellcolor[HTML]{FFFFED} 0.194  \\
& 86 & $\times$ & $\checkmark$ & $\checkmark$ & $\checkmark$ & $\times$ & $\times$ & $\checkmark$ &  \cellcolor[HTML]{FFFFED} 0.168  \\
& 87 & $\times$ & $\checkmark$ & $\checkmark$ & $\times$ & $\checkmark$ & $\checkmark$ & $\times$ &  \cellcolor[HTML]{FFFFED} 0.152  \\
& 88 & $\times$ & $\checkmark$ & $\checkmark$ & $\times$ & $\checkmark$ & $\times$ & $\checkmark$ &  \cellcolor[HTML]{FFFFED} 0.171  \\
& 89 & $\times$ & $\checkmark$ & $\checkmark$ & $\times$ & $\times$ & $\checkmark$ & $\checkmark$ &  \cellcolor[HTML]{FFFFED} 0.177  \\
& 90 & $\times$ & $\checkmark$ & $\times$ & $\checkmark$ & $\checkmark$ & $\checkmark$ & $\times$ &  \cellcolor[HTML]{FFFFED} 0.156  \\
& 91 & $\times$ & $\checkmark$ & $\times$ & $\checkmark$ & $\checkmark$ & $\times$ & $\checkmark$ &  \cellcolor[HTML]{FFFFED} 0.168  \\
& 92 & $\times$ & $\checkmark$ & $\times$ & $\checkmark$ & $\times$ & $\checkmark$ & $\checkmark$ &  \cellcolor[HTML]{FFFFED} 0.173  \\
& 93 & $\times$ & $\checkmark$ & $\times$ & $\times$ & $\checkmark$ & $\checkmark$ & $\checkmark$ &  \cellcolor[HTML]{FFFFED} 0.155  \\
& 94 & $\times$ & $\times$ & $\checkmark$ & $\checkmark$ & $\checkmark$ & $\checkmark$ & $\times$ &  \cellcolor[HTML]{FFFFED} 0.156  \\
& 95 & $\times$ & $\times$ & $\checkmark$ & $\checkmark$ & $\checkmark$ & $\times$ & $\checkmark$ &  \cellcolor[HTML]{FFFFED} 0.166  \\
& 96 & $\times$ & $\times$ & $\checkmark$ & $\checkmark$ & $\times$ & $\checkmark$ & $\checkmark$ &  \cellcolor[HTML]{FFFFED} 0.168  \\
& 97 & $\times$ & $\times$ & $\checkmark$ & $\times$ & $\checkmark$ & $\checkmark$ & $\checkmark$ &  \cellcolor[HTML]{FFFFED} 0.158  \\
& 98 & $\times$ & $\times$ & $\times$ & $\checkmark$ & $\checkmark$ & $\checkmark$ & $\checkmark$ &  \cellcolor[HTML]{FFFFED} 0.150  \\
& 99 & $\checkmark$ & $\checkmark$ & $\checkmark$ & $\checkmark$ & $\checkmark$ & $\times$ & $\times$ &  \cellcolor[HTML]{FFFFED} 0.202  \\
& 100 & $\checkmark$ & $\checkmark$ & $\checkmark$ & $\checkmark$ & $\times$ & $\checkmark$ & $\times$ &  \cellcolor[HTML]{FFFFED} 0.198  \\
& 101 & $\checkmark$ & $\checkmark$ & $\checkmark$ & $\checkmark$ & $\times$ & $\times$ & $\checkmark$ &  \cellcolor[HTML]{FFFFED} 0.164  \\
& 102 & $\checkmark$ & $\checkmark$ & $\checkmark$ & $\times$ & $\checkmark$ & $\checkmark$ & $\times$ &  \cellcolor[HTML]{FFFFED} 0.154  \\
& 103 & $\checkmark$ & $\checkmark$ & $\checkmark$ & $\times$ & $\checkmark$ & $\times$ & $\checkmark$ &  \cellcolor[HTML]{FFFFED} 0.164  \\
& 104 & $\checkmark$ & $\checkmark$ & $\checkmark$ & $\times$ & $\times$ & $\checkmark$ & $\checkmark$ &  \cellcolor[HTML]{FFFFED} 0.165  \\
& 105 & $\checkmark$ & $\checkmark$ & $\times$ & $\checkmark$ & $\checkmark$ & $\checkmark$ & $\times$ &  \cellcolor[HTML]{FFFFED} 0.161  \\
& 106 & $\checkmark$ & $\checkmark$ & $\times$ & $\checkmark$ & $\checkmark$ & $\times$ & $\checkmark$ &  \cellcolor[HTML]{FFFFED} 0.165  \\
& 107 & $\checkmark$ & $\checkmark$ & $\times$ & $\checkmark$ & $\times$ & $\checkmark$ & $\checkmark$ &  \cellcolor[HTML]{FFFFED} 0.171  \\
& 108 & $\checkmark$ & $\checkmark$ & $\times$ & $\times$ & $\checkmark$ & $\checkmark$ & $\checkmark$ &  \cellcolor[HTML]{FFFFED} 0.165  \\
& 109 & $\checkmark$ & $\times$ & $\checkmark$ & $\checkmark$ & $\checkmark$ & $\checkmark$ & $\times$ &  \cellcolor[HTML]{FFFFED} 0.159  \\
& 110 & $\checkmark$ & $\times$ & $\checkmark$ & $\checkmark$ & $\checkmark$ & $\times$ & $\checkmark$ &  \cellcolor[HTML]{FFFFED} 0.165  \\
& 111 & $\checkmark$ & $\times$ & $\checkmark$ & $\checkmark$ & $\times$ & $\checkmark$ & $\checkmark$ &  \cellcolor[HTML]{FFFFED} 0.168  \\
& 112 & $\checkmark$ & $\times$ & $\checkmark$ & $\times$ & $\checkmark$ & $\checkmark$ & $\checkmark$ &  \cellcolor[HTML]{FFFFED} 0.166  \\
& 113 & $\checkmark$ & $\times$ & $\times$ & $\checkmark$ & $\checkmark$ & $\checkmark$ & $\checkmark$ &  \cellcolor[HTML]{FFFFED} 0.153  \\
& 114 & $\times$ & $\checkmark$ & $\checkmark$ & $\checkmark$ & $\checkmark$ & $\checkmark$ & $\times$ &  \cellcolor[HTML]{FFFFED} 0.162  \\
& 115 & $\times$ & $\checkmark$ & $\checkmark$ & $\checkmark$ & $\checkmark$ & $\times$ & $\checkmark$ &  \cellcolor[HTML]{FFFFED} 0.175  \\
& 116 & $\times$ & $\checkmark$ & $\checkmark$ & $\checkmark$ & $\times$ & $\checkmark$ & $\checkmark$ &  \cellcolor[HTML]{FFFFED} 0.178  \\
& 117 & $\times$ & $\checkmark$ & $\checkmark$ & $\times$ & $\checkmark$ & $\checkmark$ & $\checkmark$ &  \cellcolor[HTML]{FFFFED} 0.159  \\
& 118 & $\times$ & $\checkmark$ & $\times$ & $\checkmark$ & $\checkmark$ & $\checkmark$ & $\checkmark$ &  \cellcolor[HTML]{FFFFED} 0.163  \\
& 119 & $\times$ & $\times$ & $\checkmark$ & $\checkmark$ & $\checkmark$ & $\checkmark$ & $\checkmark$ &  \cellcolor[HTML]{FFFFED} 0.161  \\
& 120 & $\checkmark$ & $\checkmark$ & $\checkmark$ & $\checkmark$ & $\checkmark$ & $\checkmark$ & $\times$ &  \cellcolor[HTML]{FFFFED} 0.161  \\
& 121 & $\checkmark$ & $\checkmark$ & $\checkmark$ & $\checkmark$ & $\checkmark$ & $\times$ & $\checkmark$ &  \cellcolor[HTML]{FFFFED} 0.171  \\
& 122 & $\checkmark$ & $\checkmark$ & $\checkmark$ & $\checkmark$ & $\times$ & $\checkmark$ & $\checkmark$ &  \cellcolor[HTML]{FFFFED} 0.173  \\
& 123 & $\checkmark$ & $\checkmark$ & $\checkmark$ & $\times$ & $\checkmark$ & $\checkmark$ & $\checkmark$ &  \cellcolor[HTML]{FFFFED} 0.162  \\
& 124 & $\checkmark$ & $\checkmark$ & $\times$ & $\checkmark$ & $\checkmark$ & $\checkmark$ & $\checkmark$ &  \cellcolor[HTML]{FFFFED} 0.167  \\
& 125 & $\checkmark$ & $\times$ & $\checkmark$ & $\checkmark$ & $\checkmark$ & $\checkmark$ & $\checkmark$ &  \cellcolor[HTML]{FFFFED} 0.164  \\
& 126 & $\times$ & $\checkmark$ & $\checkmark$ & $\checkmark$ & $\checkmark$ & $\checkmark$ & $\checkmark$ &  \cellcolor[HTML]{FFFFED} 0.165  \\
& 127 & $\checkmark$ & $\checkmark$ & $\checkmark$ & $\checkmark$ & $\checkmark$ & $\checkmark$ & $\checkmark$ &  \cellcolor[HTML]{FFFFED} 0.171  \\
\bottomrule
\end{tabular}
}

\end{multicols}

\end{table}

%% file: tables/ablation_mpcr.tex
\begin{table}[t]
\caption{\textbf{LOVM Model Selection Ablation.} Here, we ablate all the different scores used in our baselines for model ranking by mean per-class recall. We separated the text classification (C) base scores, and the granularity-based scores that quantify inter- and intra-class similarity.  
$a_{\text{IN}}$ - Imagenet Accuracy,  \intraClassCloseSymbl\ - \intraClassClose, text-f1 - caption dataset f1-score, text-acc1 - text top-1 accuracy, \superclassSymbl\ - \superclass, \intraSimSymbl\ - \intraSim, \interSimSymbl\ - \interSim.
\label{tab:ablation:rank_mpcr}}
\begin{multicols}{2}
\resizebox{1\textwidth}{!}{
\begin{tabular}{c|c|c|cc|cc|cc|cc@{}}
\toprule
&  & \multicolumn{7}{c|}{Scores} & \multicolumn{2}{c}{Metrics} \\
\midrule 
 & Row ID & $a_{\text{IN}}$ & text-f1 & text-acc1 &  \superclassSymbl  & \interSimSymbl & \intraSimSymbl & \intraClassCloseSymbl & $\tau$ ($\uparrow$) & $R_5$ ($\uparrow$) \\
\midrule
\multirow{63}{*}{\rotatebox[origin=c]{90}{\parbox[c]{4cm}{\centering \textbf{Model Ranking}}}}
&  1 & $\checkmark$ & $\times$ & $\times$ & $\times$ & $\times$ & $\times$ & $\times$ &  \cellcolor[HTML]{FFFFED} 0.186 &  \cellcolor[HTML]{FFFFED} 0.504  \\
&  2 & $\times$ & $\checkmark$ & $\times$ & $\times$ & $\times$ & $\times$ & $\times$ &  \cellcolor[HTML]{FFFFED} 0.058 &  \cellcolor[HTML]{FFFFED} 0.252  \\
& 3 & $\times$ & $\times$ & $\checkmark$ & $\times$ & $\times$ & $\times$ & $\times$ &  \cellcolor[HTML]{FFFFED} 0.029 &  \cellcolor[HTML]{FFFFED} 0.191  \\
& 4 & $\times$ & $\times$ & $\times$ & $\checkmark$ & $\times$ & $\times$ & $\times$ &  \cellcolor[HTML]{FFFFED} -0.014 &  \cellcolor[HTML]{FFFFED} 0.217  \\
& 5 & $\times$ & $\times$ & $\times$ & $\times$ & $\checkmark$ & $\times$ & $\times$ &  \cellcolor[HTML]{FFFFED} 0.014 &  \cellcolor[HTML]{FFFFED} 0.261  \\
&  6 & $\times$ & $\times$ & $\times$ & $\times$ & $\times$ & $\checkmark$ & $\times$ &  \cellcolor[HTML]{FFFFED} -0.014 &  \cellcolor[HTML]{FFFFED} 0.270  \\
& 7 & $\times$ & $\times$ & $\times$ & $\times$ & $\times$ & $\times$ & $\checkmark$ &  \cellcolor[HTML]{FFFFED} 0.000 &  \cellcolor[HTML]{FFFFED} 0.157  \\
&  8 & $\checkmark$ & $\checkmark$ & $\times$ & $\times$ & $\times$ & $\times$ & $\times$ &  \cellcolor[HTML]{FFFFED} 0.214 &  \cellcolor[HTML]{FFFFED} 0.522  \\
& 9 & $\checkmark$ & $\times$ & $\checkmark$ & $\times$ & $\times$ & $\times$ & $\times$ &  \cellcolor[HTML]{FFFFED} 0.209 &  \cellcolor[HTML]{FFFFED} 0.513  \\
& 10 & $\checkmark$ & $\times$ & $\times$ & $\checkmark$ & $\times$ & $\times$ & $\times$ &  \cellcolor[HTML]{FFFFED} 0.116 &  \cellcolor[HTML]{FFFFED} 0.487  \\
& 11 & $\checkmark$ & $\times$ & $\times$ & $\times$ & $\checkmark$ & $\times$ & $\times$ &  \cellcolor[HTML]{FFFFED} 0.177 &  \cellcolor[HTML]{FFFFED} 0.530  \\
& 12 & $\checkmark$ & $\times$ & $\times$ & $\times$ & $\times$ & $\checkmark$ & $\times$ &  \cellcolor[HTML]{FFFFED} 0.186 &  \cellcolor[HTML]{FFFFED} 0.504  \\
& 13 & $\checkmark$ & $\times$ & $\times$ & $\times$ & $\times$ & $\times$ & $\checkmark$ &  \cellcolor[HTML]{FFFFED} 0.072 &  \cellcolor[HTML]{FFFFED} 0.478  \\
& 14 & $\times$ & $\checkmark$ & $\checkmark$ & $\times$ & $\times$ & $\times$ & $\times$ &  \cellcolor[HTML]{FFFFED} 0.029 &  \cellcolor[HTML]{FFFFED} 0.235  \\
& 15 & $\times$ & $\checkmark$ & $\times$ & $\checkmark$ & $\times$ & $\times$ & $\times$ &  \cellcolor[HTML]{FFFFED} 0.072 &  \cellcolor[HTML]{FFFFED} 0.235  \\
& 16 & $\times$ & $\checkmark$ & $\times$ & $\times$ & $\checkmark$ & $\times$ & $\times$ &  \cellcolor[HTML]{FFFFED} 0.014 &  \cellcolor[HTML]{FFFFED} 0.243  \\
& 17 & $\times$ & $\checkmark$ & $\times$ & $\times$ & $\times$ & $\checkmark$ & $\times$ &  \cellcolor[HTML]{FFFFED} 0.014 &  \cellcolor[HTML]{FFFFED} 0.252  \\
& 18 & $\times$ & $\checkmark$ & $\times$ & $\times$ & $\times$ & $\times$ & $\checkmark$ &  \cellcolor[HTML]{FFFFED} 0.014 &  \cellcolor[HTML]{FFFFED} 0.235  \\
& 19 & $\times$ & $\times$ & $\checkmark$ & $\checkmark$ & $\times$ & $\times$ & $\times$ &  \cellcolor[HTML]{FFFFED} 0.072 &  \cellcolor[HTML]{FFFFED} 0.252  \\
& 20 & $\times$ & $\times$ & $\checkmark$ & $\times$ & $\checkmark$ & $\times$ & $\times$ &  \cellcolor[HTML]{FFFFED} 0.043 &  \cellcolor[HTML]{FFFFED} 0.243  \\
& 21 & $\times$ & $\times$ & $\checkmark$ & $\times$ & $\times$ & $\checkmark$ & $\times$ &  \cellcolor[HTML]{FFFFED} 0.043 &  \cellcolor[HTML]{FFFFED} 0.226  \\
& 22 & $\times$ & $\times$ & $\checkmark$ & $\times$ & $\times$ & $\times$ & $\checkmark$ &  \cellcolor[HTML]{FFFFED} 0.043 &  \cellcolor[HTML]{FFFFED} 0.226  \\
& 23 & $\times$ & $\times$ & $\times$ & $\checkmark$ & $\checkmark$ & $\times$ & $\times$ &  \cellcolor[HTML]{FFFFED} -0.014 &  \cellcolor[HTML]{FFFFED} 0.217  \\
& 24 & $\times$ & $\times$ & $\times$ & $\checkmark$ & $\times$ & $\checkmark$ & $\times$ &  \cellcolor[HTML]{FFFFED} 0.000 &  \cellcolor[HTML]{FFFFED} 0.252  \\
& 25 & $\times$ & $\times$ & $\times$ & $\checkmark$ & $\times$ & $\times$ & $\checkmark$ &  \cellcolor[HTML]{FFFFED} 0.000 &  \cellcolor[HTML]{FFFFED} 0.174  \\
& 26 & $\times$ & $\times$ & $\times$ & $\times$ & $\checkmark$ & $\checkmark$ & $\times$ &  \cellcolor[HTML]{FFFFED} 0.000 &  \cellcolor[HTML]{FFFFED} 0.226  \\
& 27 & $\times$ & $\times$ & $\times$ & $\times$ & $\checkmark$ & $\times$ & $\checkmark$ &  \cellcolor[HTML]{FFFFED} 0.000 &  \cellcolor[HTML]{FFFFED} 0.165  \\
& 28 & $\times$ & $\times$ & $\times$ & $\times$ & $\times$ & $\checkmark$ & $\checkmark$ &  \cellcolor[HTML]{FFFFED} 0.000 &  \cellcolor[HTML]{FFFFED} 0.217  \\
& 29 & $\checkmark$ & $\checkmark$ & $\checkmark$ & $\times$ & $\times$ & $\times$ & $\times$ &  \cellcolor[HTML]{FFFFED} 0.145 &  \cellcolor[HTML]{FFFFED} 0.487  \\
& 30 & $\checkmark$ & $\checkmark$ & $\times$ & $\checkmark$ & $\times$ & $\times$ & $\times$ &  \cellcolor[HTML]{FFFFED} 0.214 &  \cellcolor[HTML]{FFFFED} 0.504  \\
& 31 & $\checkmark$ & $\checkmark$ & $\times$ & $\times$ & $\checkmark$ & $\times$ & $\times$ &  \cellcolor[HTML]{FFFFED} 0.200 &  \cellcolor[HTML]{FFFFED} 0.504  \\
& 32 & $\checkmark$ & $\checkmark$ & $\times$ & $\times$ & $\times$ & $\checkmark$ & $\times$ &  \cellcolor[HTML]{FFFFED} 0.180 &  \cellcolor[HTML]{FFFFED} 0.496  \\
& 33 & $\checkmark$ & $\checkmark$ & $\times$ & $\times$ & $\times$ & $\times$ & $\checkmark$ &  \cellcolor[HTML]{FFFFED} 0.122 &  \cellcolor[HTML]{FFFFED} 0.496  \\
& 34 & $\checkmark$ & $\times$ & $\checkmark$ & $\checkmark$ & $\times$ & $\times$ & $\times$ &  \cellcolor[HTML]{FFFFED} 0.194 &  \cellcolor[HTML]{FFFFED} 0.496  \\
& 35 & $\checkmark$ & $\times$ & $\checkmark$ & $\times$ & $\checkmark$ & $\times$ & $\times$ &  \cellcolor[HTML]{FFFFED} 0.180 &  \cellcolor[HTML]{FFFFED} 0.504  \\
& 36 & $\checkmark$ & $\times$ & $\checkmark$ & $\times$ & $\times$ & $\checkmark$ & $\times$ &  \cellcolor[HTML]{FFFFED} 0.180 &  \cellcolor[HTML]{FFFFED} 0.504  \\
& 37 & $\checkmark$ & $\times$ & $\checkmark$ & $\times$ & $\times$ & $\times$ & $\checkmark$ &  \cellcolor[HTML]{FFFFED} 0.136 &  \cellcolor[HTML]{FFFFED} 0.504  \\
&  38 & $\checkmark$ & $\times$ & $\times$ & $\checkmark$ & $\checkmark$ & $\times$ & $\times$ &  \cellcolor[HTML]{FFFFED} 0.197 &  \cellcolor[HTML]{FFFFED} 0.539  \\
&  39 & $\checkmark$ & $\times$ & $\times$ & $\checkmark$ & $\times$ & $\checkmark$ & $\times$ &  \cellcolor[HTML]{FFFFED} 0.145 &  \cellcolor[HTML]{FFFFED} 0.496  \\
& 40 & $\checkmark$ & $\times$ & $\times$ & $\checkmark$ & $\times$ & $\times$ & $\checkmark$ &  \cellcolor[HTML]{FFFFED} 0.072 &  \cellcolor[HTML]{FFFFED} 0.470  \\
& 41 & $\checkmark$ & $\times$ & $\times$ & $\times$ & $\checkmark$ & $\checkmark$ & $\times$ &  \cellcolor[HTML]{FFFFED} 0.174 &  \cellcolor[HTML]{FFFFED} 0.487  \\
& 42 & $\checkmark$ & $\times$ & $\times$ & $\times$ & $\checkmark$ & $\times$ & $\checkmark$ &  \cellcolor[HTML]{FFFFED} 0.058 &  \cellcolor[HTML]{FFFFED} 0.487  \\
& 43 & $\checkmark$ & $\times$ & $\times$ & $\times$ & $\times$ & $\checkmark$ & $\checkmark$ &  \cellcolor[HTML]{FFFFED} 0.072 &  \cellcolor[HTML]{FFFFED} 0.487  \\
& 44 & $\times$ & $\checkmark$ & $\checkmark$ & $\checkmark$ & $\times$ & $\times$ & $\times$ &  \cellcolor[HTML]{FFFFED} 0.000 &  \cellcolor[HTML]{FFFFED} 0.243  \\
& 45 & $\times$ & $\checkmark$ & $\checkmark$ & $\times$ & $\checkmark$ & $\times$ & $\times$ &  \cellcolor[HTML]{FFFFED} 0.043 &  \cellcolor[HTML]{FFFFED} 0.235  \\
& 46 & $\times$ & $\checkmark$ & $\checkmark$ & $\times$ & $\times$ & $\checkmark$ & $\times$ &  \cellcolor[HTML]{FFFFED} 0.043 &  \cellcolor[HTML]{FFFFED} 0.243  \\
& 47 & $\times$ & $\checkmark$ & $\checkmark$ & $\times$ & $\times$ & $\times$ & $\checkmark$ &  \cellcolor[HTML]{FFFFED} 0.000 &  \cellcolor[HTML]{FFFFED} 0.226  \\
& 48 & $\times$ & $\checkmark$ & $\times$ & $\checkmark$ & $\checkmark$ & $\times$ & $\times$ &  \cellcolor[HTML]{FFFFED} 0.029 &  \cellcolor[HTML]{FFFFED} 0.252  \\
& 49 & $\times$ & $\checkmark$ & $\times$ & $\checkmark$ & $\times$ & $\checkmark$ & $\times$ &  \cellcolor[HTML]{FFFFED} 0.043 &  \cellcolor[HTML]{FFFFED} 0.243  \\
& 50 & $\times$ & $\checkmark$ & $\times$ & $\checkmark$ & $\times$ & $\times$ & $\checkmark$ &  \cellcolor[HTML]{FFFFED} 0.101 &  \cellcolor[HTML]{FFFFED} 0.252  \\
& 51 & $\times$ & $\checkmark$ & $\times$ & $\times$ & $\checkmark$ & $\checkmark$ & $\times$ &  \cellcolor[HTML]{FFFFED} 0.043 &  \cellcolor[HTML]{FFFFED} 0.252  \\
& 52 & $\times$ & $\checkmark$ & $\times$ & $\times$ & $\checkmark$ & $\times$ & $\checkmark$ &  \cellcolor[HTML]{FFFFED} 0.043 &  \cellcolor[HTML]{FFFFED} 0.243  \\
& 53 & $\times$ & $\checkmark$ & $\times$ & $\times$ & $\times$ & $\checkmark$ & $\checkmark$ &  \cellcolor[HTML]{FFFFED} 0.043 &  \cellcolor[HTML]{FFFFED} 0.243  \\
& 54 & $\times$ & $\times$ & $\checkmark$ & $\checkmark$ & $\checkmark$ & $\times$ & $\times$ &  \cellcolor[HTML]{FFFFED} 0.043 &  \cellcolor[HTML]{FFFFED} 0.243  \\
& 55 & $\times$ & $\times$ & $\checkmark$ & $\checkmark$ & $\times$ & $\checkmark$ & $\times$ &  \cellcolor[HTML]{FFFFED} 0.043 &  \cellcolor[HTML]{FFFFED} 0.252  \\
&  56 & $\times$ & $\times$ & $\checkmark$ & $\checkmark$ & $\times$ & $\times$ & $\checkmark$ &  \cellcolor[HTML]{FFFFED} 0.101 &  \cellcolor[HTML]{FFFFED} 0.261  \\
& 57 & $\times$ & $\times$ & $\checkmark$ & $\times$ & $\checkmark$ & $\checkmark$ & $\times$ &  \cellcolor[HTML]{FFFFED} 0.043 &  \cellcolor[HTML]{FFFFED} 0.226  \\
& 58 & $\times$ & $\times$ & $\checkmark$ & $\times$ & $\checkmark$ & $\times$ & $\checkmark$ &  \cellcolor[HTML]{FFFFED} 0.000 &  \cellcolor[HTML]{FFFFED} 0.226  \\
& 59 & $\times$ & $\times$ & $\checkmark$ & $\times$ & $\times$ & $\checkmark$ & $\checkmark$ &  \cellcolor[HTML]{FFFFED} 0.000 &  \cellcolor[HTML]{FFFFED} 0.217  \\
& 60 & $\times$ & $\times$ & $\times$ & $\checkmark$ & $\checkmark$ & $\checkmark$ & $\times$ &  \cellcolor[HTML]{FFFFED} 0.000 &  \cellcolor[HTML]{FFFFED} 0.226  \\
& 61 & $\times$ & $\times$ & $\times$ & $\checkmark$ & $\checkmark$ & $\times$ & $\checkmark$ &  \cellcolor[HTML]{FFFFED} 0.000 &  \cellcolor[HTML]{FFFFED} 0.165  \\
& 62 & $\times$ & $\times$ & $\times$ & $\checkmark$ & $\times$ & $\checkmark$ & $\checkmark$ &  \cellcolor[HTML]{FFFFED} 0.000 &  \cellcolor[HTML]{FFFFED} 0.209  \\
& 63 & $\times$ & $\times$ & $\times$ & $\times$ & $\checkmark$ & $\checkmark$ & $\checkmark$ &  \cellcolor[HTML]{FFFFED} 0.014 &  \cellcolor[HTML]{FFFFED} 0.226  \\
& 64 & $\checkmark$ & $\checkmark$ & $\checkmark$ & $\checkmark$ & $\times$ & $\times$ & $\times$ &  \cellcolor[HTML]{FFFFED} 0.130 &  \cellcolor[HTML]{FFFFED} 0.478  \\

\bottomrule

\end{tabular}

\hspace{0.2in}

\begin{tabular}{c|c|c|cc|cc|cc|cc@{}}
\toprule
&  & \multicolumn{7}{c|}{Scores} & \multicolumn{2}{c}{Metrics} \\
\midrule 
 & Row ID & $a_{\text{IN}}$ & text-f1 & text-acc1 &  \superclassSymbl  & \interSimSymbl & \intraSimSymbl & \intraClassCloseSymbl & $\tau$ ($\uparrow$) & $R_5$ ($\uparrow$) \\
\midrule

& 65 & $\checkmark$ & $\checkmark$ & $\checkmark$ & $\times$ & $\checkmark$ & $\times$ & $\times$ &  \cellcolor[HTML]{FFFFED} 0.130 &  \cellcolor[HTML]{FFFFED} 0.487  \\
& 66 & $\checkmark$ & $\checkmark$ & $\checkmark$ & $\times$ & $\times$ & $\checkmark$ & $\times$ &  \cellcolor[HTML]{FFFFED} 0.130 &  \cellcolor[HTML]{FFFFED} 0.478  \\
& 67 & $\checkmark$ & $\checkmark$ & $\checkmark$ & $\times$ & $\times$ & $\times$ & $\checkmark$ &  \cellcolor[HTML]{FFFFED} 0.145 &  \cellcolor[HTML]{FFFFED} 0.478  \\
& 68 & $\checkmark$ & $\checkmark$ & $\times$ & $\checkmark$ & $\checkmark$ & $\times$ & $\times$ &  \cellcolor[HTML]{FFFFED} 0.194 &  \cellcolor[HTML]{FFFFED} 0.496  \\
& 69 & $\checkmark$ & $\checkmark$ & $\times$ & $\checkmark$ & $\times$ & $\checkmark$ & $\times$ &  \cellcolor[HTML]{FFFFED} 0.186 &  \cellcolor[HTML]{FFFFED} 0.504  \\
& 70 & $\checkmark$ & $\checkmark$ & $\times$ & $\checkmark$ & $\times$ & $\times$ & $\checkmark$ &  \cellcolor[HTML]{FFFFED} 0.136 &  \cellcolor[HTML]{FFFFED} 0.496  \\
& 71 & $\checkmark$ & $\checkmark$ & $\times$ & $\times$ & $\checkmark$ & $\checkmark$ & $\times$ &  \cellcolor[HTML]{FFFFED} 0.171 &  \cellcolor[HTML]{FFFFED} 0.504  \\
& 72 & $\checkmark$ & $\checkmark$ & $\times$ & $\times$ & $\checkmark$ & $\times$ & $\checkmark$ &  \cellcolor[HTML]{FFFFED} 0.122 &  \cellcolor[HTML]{FFFFED} 0.496  \\
& 73 & $\checkmark$ & $\checkmark$ & $\times$ & $\times$ & $\times$ & $\checkmark$ & $\checkmark$ &  \cellcolor[HTML]{FFFFED} 0.122 &  \cellcolor[HTML]{FFFFED} 0.496  \\
& 74 & $\checkmark$ & $\times$ & $\checkmark$ & $\checkmark$ & $\checkmark$ & $\times$ & $\times$ &  \cellcolor[HTML]{FFFFED} 0.194 &  \cellcolor[HTML]{FFFFED} 0.496  \\
& 75 & $\checkmark$ & $\times$ & $\checkmark$ & $\checkmark$ & $\times$ & $\checkmark$ & $\times$ &  \cellcolor[HTML]{FFFFED} 0.180 &  \cellcolor[HTML]{FFFFED} 0.496  \\
& 76 & $\checkmark$ & $\times$ & $\checkmark$ & $\checkmark$ & $\times$ & $\times$ & $\checkmark$ &  \cellcolor[HTML]{FFFFED} 0.151 &  \cellcolor[HTML]{FFFFED} 0.504  \\
& 77 & $\checkmark$ & $\times$ & $\checkmark$ & $\times$ & $\checkmark$ & $\checkmark$ & $\times$ &  \cellcolor[HTML]{FFFFED} 0.171 &  \cellcolor[HTML]{FFFFED} 0.513  \\
& 78 & $\checkmark$ & $\times$ & $\checkmark$ & $\times$ & $\checkmark$ & $\times$ & $\checkmark$ &  \cellcolor[HTML]{FFFFED} 0.151 &  \cellcolor[HTML]{FFFFED} 0.504  \\
& 79 & $\checkmark$ & $\times$ & $\checkmark$ & $\times$ & $\times$ & $\checkmark$ & $\checkmark$ &  \cellcolor[HTML]{FFFFED} 0.151 &  \cellcolor[HTML]{FFFFED} 0.504  \\
& 80 & $\checkmark$ & $\times$ & $\times$ & $\checkmark$ & $\checkmark$ & $\checkmark$ & $\times$ &  \cellcolor[HTML]{FFFFED} 0.130 &  \cellcolor[HTML]{FFFFED} 0.487  \\
& 81 & $\checkmark$ & $\times$ & $\times$ & $\checkmark$ & $\checkmark$ & $\times$ & $\checkmark$ &  \cellcolor[HTML]{FFFFED} 0.058 &  \cellcolor[HTML]{FFFFED} 0.478  \\
& 82 & $\checkmark$ & $\times$ & $\times$ & $\checkmark$ & $\times$ & $\checkmark$ & $\checkmark$ &  \cellcolor[HTML]{FFFFED} 0.058 &  \cellcolor[HTML]{FFFFED} 0.478  \\
& 83 & $\checkmark$ & $\times$ & $\times$ & $\times$ & $\checkmark$ & $\checkmark$ & $\checkmark$ &  \cellcolor[HTML]{FFFFED} 0.087 &  \cellcolor[HTML]{FFFFED} 0.487  \\
& 84 & $\times$ & $\checkmark$ & $\checkmark$ & $\checkmark$ & $\checkmark$ & $\times$ & $\times$ &  \cellcolor[HTML]{FFFFED} 0.000 &  \cellcolor[HTML]{FFFFED} 0.235  \\
& 85 & $\times$ & $\checkmark$ & $\checkmark$ & $\checkmark$ & $\times$ & $\checkmark$ & $\times$ &  \cellcolor[HTML]{FFFFED} 0.000 &  \cellcolor[HTML]{FFFFED} 0.252  \\
& 86 & $\times$ & $\checkmark$ & $\checkmark$ & $\checkmark$ & $\times$ & $\times$ & $\checkmark$ &  \cellcolor[HTML]{FFFFED} 0.043 &  \cellcolor[HTML]{FFFFED} 0.261  \\
& 87 & $\times$ & $\checkmark$ & $\checkmark$ & $\times$ & $\checkmark$ & $\checkmark$ & $\times$ &  \cellcolor[HTML]{FFFFED} 0.014 &  \cellcolor[HTML]{FFFFED} 0.243  \\
& 88 & $\times$ & $\checkmark$ & $\checkmark$ & $\times$ & $\checkmark$ & $\times$ & $\checkmark$ &  \cellcolor[HTML]{FFFFED} 0.043 &  \cellcolor[HTML]{FFFFED} 0.235  \\
& 89 & $\times$ & $\checkmark$ & $\checkmark$ & $\times$ & $\times$ & $\checkmark$ & $\checkmark$ &  \cellcolor[HTML]{FFFFED} 0.043 &  \cellcolor[HTML]{FFFFED} 0.235  \\
& 90 & $\times$ & $\checkmark$ & $\times$ & $\checkmark$ & $\checkmark$ & $\checkmark$ & $\times$ &  \cellcolor[HTML]{FFFFED} 0.043 &  \cellcolor[HTML]{FFFFED} 0.243  \\
& 91 & $\times$ & $\checkmark$ & $\times$ & $\checkmark$ & $\checkmark$ & $\times$ & $\checkmark$ &  \cellcolor[HTML]{FFFFED} 0.029 &  \cellcolor[HTML]{FFFFED} 0.261  \\
& 92 & $\times$ & $\checkmark$ & $\times$ & $\checkmark$ & $\times$ & $\checkmark$ & $\checkmark$ &  \cellcolor[HTML]{FFFFED} 0.043 &  \cellcolor[HTML]{FFFFED} 0.243  \\
& 93 & $\times$ & $\checkmark$ & $\times$ & $\times$ & $\checkmark$ & $\checkmark$ & $\checkmark$ &  \cellcolor[HTML]{FFFFED} 0.014 &  \cellcolor[HTML]{FFFFED} 0.243  \\
& 94 & $\times$ & $\times$ & $\checkmark$ & $\checkmark$ & $\checkmark$ & $\checkmark$ & $\times$ &  \cellcolor[HTML]{FFFFED} 0.043 &  \cellcolor[HTML]{FFFFED} 0.252  \\
& 95 & $\times$ & $\times$ & $\checkmark$ & $\checkmark$ & $\checkmark$ & $\times$ & $\checkmark$ &  \cellcolor[HTML]{FFFFED} 0.043 &  \cellcolor[HTML]{FFFFED} 0.252  \\
& 96 & $\times$ & $\times$ & $\checkmark$ & $\checkmark$ & $\times$ & $\checkmark$ & $\checkmark$ &  \cellcolor[HTML]{FFFFED} 0.043 &  \cellcolor[HTML]{FFFFED} 0.252  \\
& 97 & $\times$ & $\times$ & $\checkmark$ & $\times$ & $\checkmark$ & $\checkmark$ & $\checkmark$ &  \cellcolor[HTML]{FFFFED} 0.043 &  \cellcolor[HTML]{FFFFED} 0.226  \\
& 98 & $\times$ & $\times$ & $\times$ & $\checkmark$ & $\checkmark$ & $\checkmark$ & $\checkmark$ &  \cellcolor[HTML]{FFFFED} 0.014 &  \cellcolor[HTML]{FFFFED} 0.261  \\
& 99 & $\checkmark$ & $\checkmark$ & $\checkmark$ & $\checkmark$ & $\checkmark$ & $\times$ & $\times$ &  \cellcolor[HTML]{FFFFED} 0.130 &  \cellcolor[HTML]{FFFFED} 0.487  \\
& 100 & $\checkmark$ & $\checkmark$ & $\checkmark$ & $\checkmark$ & $\times$ & $\checkmark$ & $\times$ &  \cellcolor[HTML]{FFFFED} 0.116 &  \cellcolor[HTML]{FFFFED} 0.487  \\
& 101 & $\checkmark$ & $\checkmark$ & $\checkmark$ & $\checkmark$ & $\times$ & $\times$ & $\checkmark$ &  \cellcolor[HTML]{FFFFED} 0.145 &  \cellcolor[HTML]{FFFFED} 0.487  \\
& 102 & $\checkmark$ & $\checkmark$ & $\checkmark$ & $\times$ & $\checkmark$ & $\checkmark$ & $\times$ &  \cellcolor[HTML]{FFFFED} 0.130 &  \cellcolor[HTML]{FFFFED} 0.478  \\
& 103 & $\checkmark$ & $\checkmark$ & $\checkmark$ & $\times$ & $\checkmark$ & $\times$ & $\checkmark$ &  \cellcolor[HTML]{FFFFED} 0.130 &  \cellcolor[HTML]{FFFFED} 0.487  \\
& 104 & $\checkmark$ & $\checkmark$ & $\checkmark$ & $\times$ & $\times$ & $\checkmark$ & $\checkmark$ &  \cellcolor[HTML]{FFFFED} 0.130 &  \cellcolor[HTML]{FFFFED} 0.487  \\
& 105 & $\checkmark$ & $\checkmark$ & $\times$ & $\checkmark$ & $\checkmark$ & $\checkmark$ & $\times$ &  \cellcolor[HTML]{FFFFED} 0.180 &  \cellcolor[HTML]{FFFFED} 0.496  \\
& 106 & $\checkmark$ & $\checkmark$ & $\times$ & $\checkmark$ & $\checkmark$ & $\times$ & $\checkmark$ &  \cellcolor[HTML]{FFFFED} 0.136 &  \cellcolor[HTML]{FFFFED} 0.496  \\
& 107 & $\checkmark$ & $\checkmark$ & $\times$ & $\checkmark$ & $\times$ & $\checkmark$ & $\checkmark$ &  \cellcolor[HTML]{FFFFED} 0.136 &  \cellcolor[HTML]{FFFFED} 0.496  \\
& 108 & $\checkmark$ & $\checkmark$ & $\times$ & $\times$ & $\checkmark$ & $\checkmark$ & $\checkmark$ &  \cellcolor[HTML]{FFFFED} 0.136 &  \cellcolor[HTML]{FFFFED} 0.496  \\
& 109 & $\checkmark$ & $\times$ & $\checkmark$ & $\checkmark$ & $\checkmark$ & $\checkmark$ & $\times$ &  \cellcolor[HTML]{FFFFED} 0.186 &  \cellcolor[HTML]{FFFFED} 0.504  \\
& 110 & $\checkmark$ & $\times$ & $\checkmark$ & $\checkmark$ & $\checkmark$ & $\times$ & $\checkmark$ &  \cellcolor[HTML]{FFFFED} 0.151 &  \cellcolor[HTML]{FFFFED} 0.504  \\
& 111 & $\checkmark$ & $\times$ & $\checkmark$ & $\checkmark$ & $\times$ & $\checkmark$ & $\checkmark$ &  \cellcolor[HTML]{FFFFED} 0.151 &  \cellcolor[HTML]{FFFFED} 0.504  \\
& 112 & $\checkmark$ & $\times$ & $\checkmark$ & $\times$ & $\checkmark$ & $\checkmark$ & $\checkmark$ &  \cellcolor[HTML]{FFFFED} 0.136 &  \cellcolor[HTML]{FFFFED} 0.496  \\
& 113 & $\checkmark$ & $\times$ & $\times$ & $\checkmark$ & $\checkmark$ & $\checkmark$ & $\checkmark$ &  \cellcolor[HTML]{FFFFED} 0.087 &  \cellcolor[HTML]{FFFFED} 0.487  \\
& 114 & $\times$ & $\checkmark$ & $\checkmark$ & $\checkmark$ & $\checkmark$ & $\checkmark$ & $\times$ &  \cellcolor[HTML]{FFFFED} 0.000 &  \cellcolor[HTML]{FFFFED} 0.252  \\
& 115 & $\times$ & $\checkmark$ & $\checkmark$ & $\checkmark$ & $\checkmark$ & $\times$ & $\checkmark$ &  \cellcolor[HTML]{FFFFED} 0.000 &  \cellcolor[HTML]{FFFFED} 0.235  \\
& 116 & $\times$ & $\checkmark$ & $\checkmark$ & $\checkmark$ & $\times$ & $\checkmark$ & $\checkmark$ &  \cellcolor[HTML]{FFFFED} 0.029 &  \cellcolor[HTML]{FFFFED} 0.252  \\
& 117 & $\times$ & $\checkmark$ & $\checkmark$ & $\times$ & $\checkmark$ & $\checkmark$ & $\checkmark$ &  \cellcolor[HTML]{FFFFED} 0.014 &  \cellcolor[HTML]{FFFFED} 0.235  \\
& 118 & $\times$ & $\checkmark$ & $\times$ & $\checkmark$ & $\checkmark$ & $\checkmark$ & $\checkmark$ &  \cellcolor[HTML]{FFFFED} 0.043 &  \cellcolor[HTML]{FFFFED} 0.252  \\
& 119 & $\times$ & $\times$ & $\checkmark$ & $\checkmark$ & $\checkmark$ & $\checkmark$ & $\checkmark$ &  \cellcolor[HTML]{FFFFED} 0.043 &  \cellcolor[HTML]{FFFFED} 0.261  \\
& 120 & $\checkmark$ & $\checkmark$ & $\checkmark$ & $\checkmark$ & $\checkmark$ & $\checkmark$ & $\times$ &  \cellcolor[HTML]{FFFFED} 0.130 &  \cellcolor[HTML]{FFFFED} 0.487  \\
& 121 & $\checkmark$ & $\checkmark$ & $\checkmark$ & $\checkmark$ & $\checkmark$ & $\times$ & $\checkmark$ &  \cellcolor[HTML]{FFFFED} 0.130 &  \cellcolor[HTML]{FFFFED} 0.487  \\
& 122 & $\checkmark$ & $\checkmark$ & $\checkmark$ & $\checkmark$ & $\times$ & $\checkmark$ & $\checkmark$ &  \cellcolor[HTML]{FFFFED} 0.130 &  \cellcolor[HTML]{FFFFED} 0.487  \\
& 123 & $\checkmark$ & $\checkmark$ & $\checkmark$ & $\times$ & $\checkmark$ & $\checkmark$ & $\checkmark$ &  \cellcolor[HTML]{FFFFED} 0.116 &  \cellcolor[HTML]{FFFFED} 0.496  \\
& 124 & $\checkmark$ & $\checkmark$ & $\times$ & $\checkmark$ & $\checkmark$ & $\checkmark$ & $\checkmark$ &  \cellcolor[HTML]{FFFFED} 0.136 &  \cellcolor[HTML]{FFFFED} 0.496  \\
& 125 & $\checkmark$ & $\times$ & $\checkmark$ & $\checkmark$ & $\checkmark$ & $\checkmark$ & $\checkmark$ &  \cellcolor[HTML]{FFFFED} 0.136 &  \cellcolor[HTML]{FFFFED} 0.496  \\
& 126 & $\times$ & $\checkmark$ & $\checkmark$ & $\checkmark$ & $\checkmark$ & $\checkmark$ & $\checkmark$ &  \cellcolor[HTML]{FFFFED} 0.043 &  \cellcolor[HTML]{FFFFED} 0.252  \\
& 127 & $\checkmark$ & $\checkmark$ & $\checkmark$ & $\checkmark$ & $\checkmark$ & $\checkmark$ & $\checkmark$ &  \cellcolor[HTML]{FFFFED} 0.130 &  \cellcolor[HTML]{FFFFED} 0.478  \\
\bottomrule
\end{tabular}}

\end{multicols}

\end{table}

\begin{table}[t]
\caption{\textbf{LOVM Model Prediction Ablation.} Here, we ablate all the different scores used in our baselines for mean per-class recall prediction.  We separated the text classification (C) base scores, and the granularity-based scores that quantify inter- and intra-class similarity. 
$a_{\text{IN}}$ - Imagenet Accuracy,  \intraClassCloseSymbl\ - \intraClassClose, text-f1 - caption dataset f1-score, text-acc1 - text top-1 accuracy, \superclassSymbl\ - \superclass, \intraSimSymbl\ - \intraSim, \interSimSymbl\ - \interSim.
\label{tab:ablation:pred_mpcr}}

\begin{multicols}{2}
\resizebox{1\textwidth}{!}{
\begin{tabular}{c|c|c|cc|cc|cc|c@{}}
\toprule
&  & \multicolumn{7}{c|}{Scores} & \multicolumn{1}{c}{Metrics} \\
\midrule 
 & Row ID & $a_{\text{IN}}$ & text-f1 & text-acc1 &  \superclassSymbl  & \interSimSymbl & \intraSimSymbl & \intraClassCloseSymbl & $L_1$ ($\downarrow$) ) \\
\midrule
\multirow{63}{*}{\rotatebox[origin=c]{90}{\parbox[c]{4cm}{\centering \textbf{Metric Prediction}}}}
&  1 & $\checkmark$ & $\times$ & $\times$ & $\times$ & $\times$ & $\times$ & $\times$ &  \cellcolor[HTML]{FFFFED} 0.228  \\
&  2 & $\times$ & $\checkmark$ & $\times$ & $\times$ & $\times$ & $\times$ & $\times$ &  \cellcolor[HTML]{FFFFED} 0.182  \\
& 3 & $\times$ & $\times$ & $\checkmark$ & $\times$ & $\times$ & $\times$ & $\times$ &  \cellcolor[HTML]{FFFFED} 0.183  \\
& 4 & $\times$ & $\times$ & $\times$ & $\checkmark$ & $\times$ & $\times$ & $\times$ &  \cellcolor[HTML]{FFFFED} 0.179  \\
& 5 & $\times$ & $\times$ & $\times$ & $\times$ & $\checkmark$ & $\times$ & $\times$ &  \cellcolor[HTML]{FFFFED} 0.205  \\
& 6 & $\times$ & $\times$ & $\times$ & $\times$ & $\times$ & $\checkmark$ & $\times$ &  \cellcolor[HTML]{FFFFED} 0.232  \\
& 7 & $\times$ & $\times$ & $\times$ & $\times$ & $\times$ & $\times$ & $\checkmark$ &  \cellcolor[HTML]{FFFFED} 0.175  \\
& 8 & $\checkmark$ & $\checkmark$ & $\times$ & $\times$ & $\times$ & $\times$ & $\times$ &  \cellcolor[HTML]{FFFFED} 0.196  \\
& 9 & $\checkmark$ & $\times$ & $\checkmark$ & $\times$ & $\times$ & $\times$ & $\times$ &  \cellcolor[HTML]{FFFFED} 0.190  \\
& 10 & $\checkmark$ & $\times$ & $\times$ & $\checkmark$ & $\times$ & $\times$ & $\times$ &  \cellcolor[HTML]{FFFFED} 0.225  \\
& 11 & $\checkmark$ & $\times$ & $\times$ & $\times$ & $\checkmark$ & $\times$ & $\times$ &  \cellcolor[HTML]{FFFFED} 0.218  \\
& 12 & $\checkmark$ & $\times$ & $\times$ & $\times$ & $\times$ & $\checkmark$ & $\times$ &  \cellcolor[HTML]{FFFFED} 0.228  \\
& 13 & $\checkmark$ & $\times$ & $\times$ & $\times$ & $\times$ & $\times$ & $\checkmark$ &  \cellcolor[HTML]{FFFFED} 0.177  \\
& 14 & $\times$ & $\checkmark$ & $\checkmark$ & $\times$ & $\times$ & $\times$ & $\times$ &  \cellcolor[HTML]{FFFFED} 0.182  \\
& 15 & $\times$ & $\checkmark$ & $\times$ & $\checkmark$ & $\times$ & $\times$ & $\times$ &  \cellcolor[HTML]{FFFFED} 0.192  \\
& 16 & $\times$ & $\checkmark$ & $\times$ & $\times$ & $\checkmark$ & $\times$ & $\times$ &  \cellcolor[HTML]{FFFFED} 0.191  \\
& 17 & $\times$ & $\checkmark$ & $\times$ & $\times$ & $\times$ & $\checkmark$ & $\times$ &  \cellcolor[HTML]{FFFFED} 0.188  \\
& 18 & $\times$ & $\checkmark$ & $\times$ & $\times$ & $\times$ & $\times$ & $\checkmark$ &  \cellcolor[HTML]{FFFFED} 0.165  \\
& 19 & $\times$ & $\times$ & $\checkmark$ & $\checkmark$ & $\times$ & $\times$ & $\times$ &  \cellcolor[HTML]{FFFFED} 0.190  \\
& 20 & $\times$ & $\times$ & $\checkmark$ & $\times$ & $\checkmark$ & $\times$ & $\times$ &  \cellcolor[HTML]{FFFFED} 0.190  \\
& 21 & $\times$ & $\times$ & $\checkmark$ & $\times$ & $\times$ & $\checkmark$ & $\times$ &  \cellcolor[HTML]{FFFFED} 0.187  \\
& 22 & $\times$ & $\times$ & $\checkmark$ & $\times$ & $\times$ & $\times$ & $\checkmark$ &  \cellcolor[HTML]{FFFFED} 0.162  \\
& 23 & $\times$ & $\times$ & $\times$ & $\checkmark$ & $\checkmark$ & $\times$ & $\times$ &  \cellcolor[HTML]{FFFFED} 0.214  \\
& 24 & $\times$ & $\times$ & $\times$ & $\checkmark$ & $\times$ & $\checkmark$ & $\times$ &  \cellcolor[HTML]{FFFFED} 0.222  \\
& 25 & $\times$ & $\times$ & $\times$ & $\checkmark$ & $\times$ & $\times$ & $\checkmark$ &  \cellcolor[HTML]{FFFFED} 0.159  \\
& 26 & $\times$ & $\times$ & $\times$ & $\times$ & $\checkmark$ & $\checkmark$ & $\times$ &  \cellcolor[HTML]{FFFFED} 0.151  \\
& 27 & $\times$ & $\times$ & $\times$ & $\times$ & $\checkmark$ & $\times$ & $\checkmark$ &  \cellcolor[HTML]{FFFFED} 0.177  \\
& 28 & $\times$ & $\times$ & $\times$ & $\times$ & $\times$ & $\checkmark$ & $\checkmark$ &  \cellcolor[HTML]{FFFFED} 0.181  \\
& 29 & $\checkmark$ & $\checkmark$ & $\checkmark$ & $\times$ & $\times$ & $\times$ & $\times$ &  \cellcolor[HTML]{FFFFED} 0.197  \\
& 30 & $\checkmark$ & $\checkmark$ & $\times$ & $\checkmark$ & $\times$ & $\times$ & $\times$ &  \cellcolor[HTML]{FFFFED} 0.203  \\
& 31 & $\checkmark$ & $\checkmark$ & $\times$ & $\times$ & $\checkmark$ & $\times$ & $\times$ &  \cellcolor[HTML]{FFFFED} 0.196  \\
& 32 & $\checkmark$ & $\checkmark$ & $\times$ & $\times$ & $\times$ & $\checkmark$ & $\times$ &  \cellcolor[HTML]{FFFFED} 0.195  \\
& 33 & $\checkmark$ & $\checkmark$ & $\times$ & $\times$ & $\times$ & $\times$ & $\checkmark$ &  \cellcolor[HTML]{FFFFED} 0.164  \\
& 34 & $\checkmark$ & $\times$ & $\checkmark$ & $\checkmark$ & $\times$ & $\times$ & $\times$ &  \cellcolor[HTML]{FFFFED} 0.203  \\
& 35 & $\checkmark$ & $\times$ & $\checkmark$ & $\times$ & $\checkmark$ & $\times$ & $\times$ &  \cellcolor[HTML]{FFFFED} 0.196  \\
& 36 & $\checkmark$ & $\times$ & $\checkmark$ & $\times$ & $\times$ & $\checkmark$ & $\times$ &  \cellcolor[HTML]{FFFFED} 0.194  \\
& 37 & $\checkmark$ & $\times$ & $\checkmark$ & $\times$ & $\times$ & $\times$ & $\checkmark$ &  \cellcolor[HTML]{FFFFED} 0.163  \\
& 38 & $\checkmark$ & $\times$ & $\times$ & $\checkmark$ & $\checkmark$ & $\times$ & $\times$ &  \cellcolor[HTML]{FFFFED} 0.227  \\
& 39 & $\checkmark$ & $\times$ & $\times$ & $\checkmark$ & $\times$ & $\checkmark$ & $\times$ &  \cellcolor[HTML]{FFFFED} 0.232  \\
& 40 & $\checkmark$ & $\times$ & $\times$ & $\checkmark$ & $\times$ & $\times$ & $\checkmark$ &  \cellcolor[HTML]{FFFFED} 0.164  \\
& 41 & $\checkmark$ & $\times$ & $\times$ & $\times$ & $\checkmark$ & $\checkmark$ & $\times$ &  \cellcolor[HTML]{FFFFED} 0.156  \\
& 42 & $\checkmark$ & $\times$ & $\times$ & $\times$ & $\checkmark$ & $\times$ & $\checkmark$ &  \cellcolor[HTML]{FFFFED} 0.177  \\
& 43 & $\checkmark$ & $\times$ & $\times$ & $\times$ & $\times$ & $\checkmark$ & $\checkmark$ &  \cellcolor[HTML]{FFFFED} 0.179  \\
& 44 & $\times$ & $\checkmark$ & $\checkmark$ & $\checkmark$ & $\times$ & $\times$ & $\times$ &  \cellcolor[HTML]{FFFFED} 0.191  \\
& 45 & $\times$ & $\checkmark$ & $\checkmark$ & $\times$ & $\checkmark$ & $\times$ & $\times$ &  \cellcolor[HTML]{FFFFED} 0.192  \\
& 46 & $\times$ & $\checkmark$ & $\checkmark$ & $\times$ & $\times$ & $\checkmark$ & $\times$ &  \cellcolor[HTML]{FFFFED} 0.191  \\
& 47 & $\times$ & $\checkmark$ & $\checkmark$ & $\times$ & $\times$ & $\times$ & $\checkmark$ &  \cellcolor[HTML]{FFFFED} 0.172  \\
& 48 & $\times$ & $\checkmark$ & $\times$ & $\checkmark$ & $\checkmark$ & $\times$ & $\times$ &  \cellcolor[HTML]{FFFFED} 0.197  \\
& 49 & $\times$ & $\checkmark$ & $\times$ & $\checkmark$ & $\times$ & $\checkmark$ & $\times$ &  \cellcolor[HTML]{FFFFED} 0.197  \\
& 50 & $\times$ & $\checkmark$ & $\times$ & $\checkmark$ & $\times$ & $\times$ & $\checkmark$ &  \cellcolor[HTML]{FFFFED} 0.162  \\
& 51 & $\times$ & $\checkmark$ & $\times$ & $\times$ & $\checkmark$ & $\checkmark$ & $\times$ &  \cellcolor[HTML]{FFFFED} 0.155  \\
& 52 & $\times$ & $\checkmark$ & $\times$ & $\times$ & $\checkmark$ & $\times$ & $\checkmark$ &  \cellcolor[HTML]{FFFFED} 0.171  \\
& 53 & $\times$ & $\checkmark$ & $\times$ & $\times$ & $\times$ & $\checkmark$ & $\checkmark$ &  \cellcolor[HTML]{FFFFED} 0.176  \\
& 54 & $\times$ & $\times$ & $\checkmark$ & $\checkmark$ & $\checkmark$ & $\times$ & $\times$ &  \cellcolor[HTML]{FFFFED} 0.194  \\
& 55 & $\times$ & $\times$ & $\checkmark$ & $\checkmark$ & $\times$ & $\checkmark$ & $\times$ &  \cellcolor[HTML]{FFFFED} 0.192  \\
& 56 & $\times$ & $\times$ & $\checkmark$ & $\checkmark$ & $\times$ & $\times$ & $\checkmark$ &  \cellcolor[HTML]{FFFFED} 0.159  \\
& 57 & $\times$ & $\times$ & $\checkmark$ & $\times$ & $\checkmark$ & $\checkmark$ & $\times$ &  \cellcolor[HTML]{FFFFED} 0.153  \\
& 58 & $\times$ & $\times$ & $\checkmark$ & $\times$ & $\checkmark$ & $\times$ & $\checkmark$ &  \cellcolor[HTML]{FFFFED} 0.169  \\
& 59 & $\times$ & $\times$ & $\checkmark$ & $\times$ & $\times$ & $\checkmark$ & $\checkmark$ &  \cellcolor[HTML]{FFFFED} 0.173  \\
& 60 & $\times$ & $\times$ & $\times$ & $\checkmark$ & $\checkmark$ & $\checkmark$ & $\times$ &  \cellcolor[HTML]{FFFFED} 0.145  \\
& 61 & $\times$ & $\times$ & $\times$ & $\checkmark$ & $\checkmark$ & $\times$ & $\checkmark$ &  \cellcolor[HTML]{FFFFED} 0.164  \\
& 62 & $\times$ & $\times$ & $\times$ & $\checkmark$ & $\times$ & $\checkmark$ & $\checkmark$ &  \cellcolor[HTML]{FFFFED} 0.164  \\
& 63 & $\times$ & $\times$ & $\times$ & $\times$ & $\checkmark$ & $\checkmark$ & $\checkmark$ &  \cellcolor[HTML]{FFFFED} 0.159  \\
& 64 & $\checkmark$ & $\checkmark$ & $\checkmark$ & $\checkmark$ & $\times$ & $\times$ & $\times$ &  \cellcolor[HTML]{FFFFED} 0.206  \\

\bottomrule

\end{tabular}

\hspace{0.2in}

\begin{tabular}{c|c|c|cc|cc|cc|cc@{}}
\toprule
&  & \multicolumn{7}{c|}{Scores} & \multicolumn{1}{c}{Metrics} \\
\midrule 
 & Row ID & $a_{\text{IN}}$ & text-f1 & text-acc1 &  \superclassSymbl  & \interSimSymbl & \intraSimSymbl & \intraClassCloseSymbl & $L_1$ ($\downarrow$) ) \\
\midrule

& 65 & $\checkmark$ & $\checkmark$ & $\checkmark$ & $\times$ & $\checkmark$ & $\times$ & $\times$ &  \cellcolor[HTML]{FFFFED} 0.200  \\
& 66 & $\checkmark$ & $\checkmark$ & $\checkmark$ & $\times$ & $\times$ & $\checkmark$ & $\times$ &  \cellcolor[HTML]{FFFFED} 0.199  \\
& 67 & $\checkmark$ & $\checkmark$ & $\checkmark$ & $\times$ & $\times$ & $\times$ & $\checkmark$ &  \cellcolor[HTML]{FFFFED} 0.170  \\
& 68 & $\checkmark$ & $\checkmark$ & $\times$ & $\checkmark$ & $\checkmark$ & $\times$ & $\times$ &  \cellcolor[HTML]{FFFFED} 0.209  \\
& 69 & $\checkmark$ & $\checkmark$ & $\times$ & $\checkmark$ & $\times$ & $\checkmark$ & $\times$ &  \cellcolor[HTML]{FFFFED} 0.206  \\
& 70 & $\checkmark$ & $\checkmark$ & $\times$ & $\checkmark$ & $\times$ & $\times$ & $\checkmark$ &  \cellcolor[HTML]{FFFFED} 0.168  \\
& 71 & $\checkmark$ & $\checkmark$ & $\times$ & $\times$ & $\checkmark$ & $\checkmark$ & $\times$ &  \cellcolor[HTML]{FFFFED} 0.160  \\
& 72 & $\checkmark$ & $\checkmark$ & $\times$ & $\times$ & $\checkmark$ & $\times$ & $\checkmark$ &  \cellcolor[HTML]{FFFFED} 0.169  \\
& 73 & $\checkmark$ & $\checkmark$ & $\times$ & $\times$ & $\times$ & $\checkmark$ & $\checkmark$ &  \cellcolor[HTML]{FFFFED} 0.171  \\
& 74 & $\checkmark$ & $\times$ & $\checkmark$ & $\checkmark$ & $\checkmark$ & $\times$ & $\times$ &  \cellcolor[HTML]{FFFFED} 0.207  \\
& 75 & $\checkmark$ & $\times$ & $\checkmark$ & $\checkmark$ & $\times$ & $\checkmark$ & $\times$ &  \cellcolor[HTML]{FFFFED} 0.205  \\
& 76 & $\checkmark$ & $\times$ & $\checkmark$ & $\checkmark$ & $\times$ & $\times$ & $\checkmark$ &  \cellcolor[HTML]{FFFFED} 0.166  \\
& 77 & $\checkmark$ & $\times$ & $\checkmark$ & $\times$ & $\checkmark$ & $\checkmark$ & $\times$ &  \cellcolor[HTML]{FFFFED} 0.161  \\
& 78 & $\checkmark$ & $\times$ & $\checkmark$ & $\times$ & $\checkmark$ & $\times$ & $\checkmark$ &  \cellcolor[HTML]{FFFFED} 0.170  \\
& 79 & $\checkmark$ & $\times$ & $\checkmark$ & $\times$ & $\times$ & $\checkmark$ & $\checkmark$ &  \cellcolor[HTML]{FFFFED} 0.171  \\
& 80 & $\checkmark$ & $\times$ & $\times$ & $\checkmark$ & $\checkmark$ & $\checkmark$ & $\times$ &  \cellcolor[HTML]{FFFFED} 0.152  \\
& 81 & $\checkmark$ & $\times$ & $\times$ & $\checkmark$ & $\checkmark$ & $\times$ & $\checkmark$ &  \cellcolor[HTML]{FFFFED} 0.166  \\
& 82 & $\checkmark$ & $\times$ & $\times$ & $\checkmark$ & $\times$ & $\checkmark$ & $\checkmark$ &  \cellcolor[HTML]{FFFFED} 0.173  \\
& 83 & $\checkmark$ & $\times$ & $\times$ & $\times$ & $\checkmark$ & $\checkmark$ & $\checkmark$ &  \cellcolor[HTML]{FFFFED} 0.163  \\
& 84 & $\times$ & $\checkmark$ & $\checkmark$ & $\checkmark$ & $\checkmark$ & $\times$ & $\times$ &  \cellcolor[HTML]{FFFFED} 0.193  \\
& 85 & $\times$ & $\checkmark$ & $\checkmark$ & $\checkmark$ & $\times$ & $\checkmark$ & $\times$ &  \cellcolor[HTML]{FFFFED} 0.197  \\
& 86 & $\times$ & $\checkmark$ & $\checkmark$ & $\checkmark$ & $\times$ & $\times$ & $\checkmark$ &  \cellcolor[HTML]{FFFFED} 0.170  \\
& 87 & $\times$ & $\checkmark$ & $\checkmark$ & $\times$ & $\checkmark$ & $\checkmark$ & $\times$ &  \cellcolor[HTML]{FFFFED} 0.160  \\
& 88 & $\times$ & $\checkmark$ & $\checkmark$ & $\times$ & $\checkmark$ & $\times$ & $\checkmark$ &  \cellcolor[HTML]{FFFFED} 0.180  \\
& 89 & $\times$ & $\checkmark$ & $\checkmark$ & $\times$ & $\times$ & $\checkmark$ & $\checkmark$ &  \cellcolor[HTML]{FFFFED} 0.182  \\
& 90 & $\times$ & $\checkmark$ & $\times$ & $\checkmark$ & $\checkmark$ & $\checkmark$ & $\times$ &  \cellcolor[HTML]{FFFFED} 0.156  \\
& 91 & $\times$ & $\checkmark$ & $\times$ & $\checkmark$ & $\checkmark$ & $\times$ & $\checkmark$ &  \cellcolor[HTML]{FFFFED} 0.172  \\
& 92 & $\times$ & $\checkmark$ & $\times$ & $\checkmark$ & $\times$ & $\checkmark$ & $\checkmark$ &  \cellcolor[HTML]{FFFFED} 0.176  \\
& 93 & $\times$ & $\checkmark$ & $\times$ & $\times$ & $\checkmark$ & $\checkmark$ & $\checkmark$ &  \cellcolor[HTML]{FFFFED} 0.164  \\
& 94 & $\times$ & $\times$ & $\checkmark$ & $\checkmark$ & $\checkmark$ & $\checkmark$ & $\times$ &  \cellcolor[HTML]{FFFFED} 0.155  \\
& 95 & $\times$ & $\times$ & $\checkmark$ & $\checkmark$ & $\checkmark$ & $\times$ & $\checkmark$ &  \cellcolor[HTML]{FFFFED} 0.171  \\
& 96 & $\times$ & $\times$ & $\checkmark$ & $\checkmark$ & $\times$ & $\checkmark$ & $\checkmark$ &  \cellcolor[HTML]{FFFFED} 0.172  \\
& 97 & $\times$ & $\times$ & $\checkmark$ & $\times$ & $\checkmark$ & $\checkmark$ & $\checkmark$ &  \cellcolor[HTML]{FFFFED} 0.162  \\
& 98 & $\times$ & $\times$ & $\times$ & $\checkmark$ & $\checkmark$ & $\checkmark$ & $\checkmark$ &  \cellcolor[HTML]{FFFFED} 0.151  \\
& 99 & $\checkmark$ & $\checkmark$ & $\checkmark$ & $\checkmark$ & $\checkmark$ & $\times$ & $\times$ &  \cellcolor[HTML]{FFFFED} 0.211  \\
& 100 & $\checkmark$ & $\checkmark$ & $\checkmark$ & $\checkmark$ & $\times$ & $\checkmark$ & $\times$ &  \cellcolor[HTML]{FFFFED} 0.203  \\
& 101 & $\checkmark$ & $\checkmark$ & $\checkmark$ & $\checkmark$ & $\times$ & $\times$ & $\checkmark$ &  \cellcolor[HTML]{FFFFED} 0.171  \\
& 102 & $\checkmark$ & $\checkmark$ & $\checkmark$ & $\times$ & $\checkmark$ & $\checkmark$ & $\times$ &  \cellcolor[HTML]{FFFFED} 0.163  \\
& 103 & $\checkmark$ & $\checkmark$ & $\checkmark$ & $\times$ & $\checkmark$ & $\times$ & $\checkmark$ &  \cellcolor[HTML]{FFFFED} 0.176  \\
& 104 & $\checkmark$ & $\checkmark$ & $\checkmark$ & $\times$ & $\times$ & $\checkmark$ & $\checkmark$ &  \cellcolor[HTML]{FFFFED} 0.178  \\
& 105 & $\checkmark$ & $\checkmark$ & $\times$ & $\checkmark$ & $\checkmark$ & $\checkmark$ & $\times$ &  \cellcolor[HTML]{FFFFED} 0.162  \\
& 106 & $\checkmark$ & $\checkmark$ & $\times$ & $\checkmark$ & $\checkmark$ & $\times$ & $\checkmark$ &  \cellcolor[HTML]{FFFFED} 0.176  \\
& 107 & $\checkmark$ & $\checkmark$ & $\times$ & $\checkmark$ & $\times$ & $\checkmark$ & $\checkmark$ &  \cellcolor[HTML]{FFFFED} 0.177  \\
& 108 & $\checkmark$ & $\checkmark$ & $\times$ & $\times$ & $\checkmark$ & $\checkmark$ & $\checkmark$ &  \cellcolor[HTML]{FFFFED} 0.171  \\
& 109 & $\checkmark$ & $\times$ & $\checkmark$ & $\checkmark$ & $\checkmark$ & $\checkmark$ & $\times$ &  \cellcolor[HTML]{FFFFED} 0.162  \\
& 110 & $\checkmark$ & $\times$ & $\checkmark$ & $\checkmark$ & $\checkmark$ & $\times$ & $\checkmark$ &  \cellcolor[HTML]{FFFFED} 0.174  \\
& 111 & $\checkmark$ & $\times$ & $\checkmark$ & $\checkmark$ & $\times$ & $\checkmark$ & $\checkmark$ &  \cellcolor[HTML]{FFFFED} 0.177  \\
& 112 & $\checkmark$ & $\times$ & $\checkmark$ & $\times$ & $\checkmark$ & $\checkmark$ & $\checkmark$ &  \cellcolor[HTML]{FFFFED} 0.171  \\
& 113 & $\checkmark$ & $\times$ & $\times$ & $\checkmark$ & $\checkmark$ & $\checkmark$ & $\checkmark$ &  \cellcolor[HTML]{FFFFED} 0.165  \\
& 114 & $\times$ & $\checkmark$ & $\checkmark$ & $\checkmark$ & $\checkmark$ & $\checkmark$ & $\times$ &  \cellcolor[HTML]{FFFFED} 0.159  \\
& 115 & $\times$ & $\checkmark$ & $\checkmark$ & $\checkmark$ & $\checkmark$ & $\times$ & $\checkmark$ &  \cellcolor[HTML]{FFFFED} 0.178  \\
& 116 & $\times$ & $\checkmark$ & $\checkmark$ & $\checkmark$ & $\times$ & $\checkmark$ & $\checkmark$ &  \cellcolor[HTML]{FFFFED} 0.181  \\
& 117 & $\times$ & $\checkmark$ & $\checkmark$ & $\times$ & $\checkmark$ & $\checkmark$ & $\checkmark$ &  \cellcolor[HTML]{FFFFED} 0.168  \\
& 118 & $\times$ & $\checkmark$ & $\times$ & $\checkmark$ & $\checkmark$ & $\checkmark$ & $\checkmark$ &  \cellcolor[HTML]{FFFFED} 0.164  \\
& 119 & $\times$ & $\times$ & $\checkmark$ & $\checkmark$ & $\checkmark$ & $\checkmark$ & $\checkmark$ &  \cellcolor[HTML]{FFFFED} 0.162  \\
& 120 & $\checkmark$ & $\checkmark$ & $\checkmark$ & $\checkmark$ & $\checkmark$ & $\checkmark$ & $\times$ &  \cellcolor[HTML]{FFFFED} 0.163  \\
& 121 & $\checkmark$ & $\checkmark$ & $\checkmark$ & $\checkmark$ & $\checkmark$ & $\times$ & $\checkmark$ &  \cellcolor[HTML]{FFFFED} 0.180  \\
& 122 & $\checkmark$ & $\checkmark$ & $\checkmark$ & $\checkmark$ & $\times$ & $\checkmark$ & $\checkmark$ &  \cellcolor[HTML]{FFFFED} 0.181  \\
& 123 & $\checkmark$ & $\checkmark$ & $\checkmark$ & $\times$ & $\checkmark$ & $\checkmark$ & $\checkmark$ &  \cellcolor[HTML]{FFFFED} 0.172  \\
& 124 & $\checkmark$ & $\checkmark$ & $\times$ & $\checkmark$ & $\checkmark$ & $\checkmark$ & $\checkmark$ &  \cellcolor[HTML]{FFFFED} 0.171  \\
& 125 & $\checkmark$ & $\times$ & $\checkmark$ & $\checkmark$ & $\checkmark$ & $\checkmark$ & $\checkmark$ &  \cellcolor[HTML]{FFFFED} 0.171  \\
& 126 & $\times$ & $\checkmark$ & $\checkmark$ & $\checkmark$ & $\checkmark$ & $\checkmark$ & $\checkmark$ &  \cellcolor[HTML]{FFFFED} 0.167  \\
& 127 & $\checkmark$ & $\checkmark$ & $\checkmark$ & $\checkmark$ & $\checkmark$ & $\checkmark$ & $\checkmark$ &  \cellcolor[HTML]{FFFFED} 0.174  \\
\bottomrule
\end{tabular}}
\end{multicols}

\end{table}

%% file: tables/templates.tex
\begin{figure}
\begin{multicols}{2}
\begin{python}
# imagenet prompts
imagenet = [
f'a bad photo of a {c}.',
f'a photo of many {c}.',
f'a bright photo of a {c}.',
f'a photo of a clean {c}.',
f'a photo of a dirty {c}.',
f'a photo of my {c}.',
f'a photo of the cool {c}.',
f'a close-up photo of a {c}.',
f'a bright photo of the {c}.',
f'a photo of the dirty {c}.',
f'a photo of the {c}.',
f'a good photo of the {c}.',
f'a photo of one {c}.',
f'a close-up photo of the {c}.',
f'a photo of a {c}.',
f'a photo of the clean {c}.',
f'a photo of a large {c}.',
f'a photo of a nice {c}.',
f'a good photo of a {c}.',
f'a photo of the nice {c}.',
f'a photo of the small {c}.',
f'a photo of the weird {c}.',
f'a photo of the large {c}.',
f'a photo of a cool {c}.',
f'a photo of a small {c}.',
]

# imagenet v2 prompts
imagenet_v2 = [
 f'a bad photo of a {c}.',
 f'a photo of many {c}.',
 f'a bright photo of a {c}.',
 f'a photo of a clean {c}.',
 f'a photo of a dirty {c}.',
 f'a photo of my {c}.',
 f'a photo of the cool {c}.',
 f'a close-up photo of a {c}.',
 f'a bright photo of the {c}.',
 f'a photo of the dirty {c}.',
 f'a photo of the {c}.',
 f'a good photo of the {c}.',
 f'a photo of one {c}.',
 f'a close-up photo of the {c}.',
 f'a photo of a {c}.',
 f'a photo of the clean {c}.',
 f'a photo of a large {c}.',
 f'a photo of a nice {c}.',
 f'a good photo of a {c}.',
 f'a photo of the nice {c}.',
 f'a photo of the small {c}.',
 f'a photo of the weird {c}.',
 f'a photo of the large {c}.',
 f'a photo of a cool {c}.',
 f'a photo of a small {c}.',
]
\end{python}

\begin{python}
# imagenet-a prompts
imagenet-a = [
 f'a bad photo of a {c}.',
 f'a bad photo of the {c}.',
 f'a cropped photo of the {c}.',
 f'a photo of a hard to see {c}.',
 f'a photo of a dirty {c}.',
 f'a dark photo of the {c}.',
 f'a pixelated photo of the {c}.',
 f'a cropped photo of a {c}.',
 f'a photo of the dirty {c}.',
 f'a blurry photo of the {c}.',
 f'a photo of a weird {c}.',
 f'a blurry photo of a {c}.',
 f'a pixelated photo of a {c}.',
 f'a photo of the weird {c}.',
]

# imagenet-s prompts
imagenet-s = [
 f'a drawing of a {c}.',
 f'a doodle of a {c}.',
 f'a sketch of a {c}.',
 f'a doodle of the {c}.',
 f'a sketch of the {c}.',
]

# imagenet-r prompts
imagenet-r = [
 f'a sculpture of a {c}.',
 f'a rendering of a {c}.',
 f'graffiti of a {c}.',
 f'a tattoo of a {c}.',
 f'the embroidered {c}.',
 f'a drawing of a {c}.',
 f'the plastic {c}.',
 f'a painting of the {c}.',
 f'a painting of a {c}.',
 f'a sculpture of the {c}.',
 f'a plastic {c}.',
 f'a rendering of the {c}.',
 f'a {c} in a video game.',
 f'the origami {c}.',
 f'the {c} in a video game.',
 f'a origami {c}.',
 f'the toy {c}.',
 f'a rendition of a {c}.',
 f'a cartoon {c}.',
 f'art of a {c}.',
 f'a sketch of the {c}.',
 f'a embroidered {c}.',
 f'a plushie {c}.',
 f'the cartoon {c}.',
 f'the plushie {c}.',
 f'graffiti of the {c}.',
 f'a toy {c}.',
 f'a tattoo of the {c}.'
]
\end{python}
\end{multicols}\vspace{-0.2in}
\caption{\textbf{Prompting Templates Examples.} Above can be examples of different prompting templates used in the study. When using a prompting template, the `\{c\}' character is replaced by the class name. \label{sup:fig:templates}}
\end{figure}

%% file: tables/domain_shift_rank.tex
\begin{table*}[t]
\caption{
    \textbf{Dataset difficulty prediction.} Here we evaluate different method's ability to rank variations of dataset based on their difficulty. Ground truth is based on CLIP's zero-shot performance on each dataset. We evaluate each method based on Kendall's rank correlation ($\tau$). IN - ImageNet, V2 - ImageNet-v2, A - ImageNet Adversarial, R - Imagenet Rendition, S - ImageNet Sketch. }
    \label{tab:sup_dataset_difficulty}
\vspace{0.8em}
\centering
\adjustbox{width=\textwidth}{

\begin{tabular}{c|c|ccccc|ccccc|c}
\toprule
 Method & Metric & \multicolumn{5}{c|}{Raw Value} & \multicolumn{5}{c|}{Rank} & $\tau$ ($\uparrow$) \\
 \cline{3-12}
  &                                            & IN    & V2    & A     & R     & S     & IN & V2 & A & R & S &  \\
 \midrule
 \cellcolor[HTML]{EDF6FF} True Performance &  \cellcolor[HTML]{EDF6FF} acc ($\uparrow$)                   & \cellcolor[HTML]{EDF6FF} 0.762 & \cellcolor[HTML]{EDF6FF} 0.701 &\cellcolor[HTML]{EDF6FF} 0.771 &\cellcolor[HTML]{EDF6FF} 0.889 & \cellcolor[HTML]{EDF6FF} 0.602 & \cellcolor[HTML]{EDF6FF} 3  & \cellcolor[HTML]{EDF6FF} 4  & \cellcolor[HTML]{EDF6FF} 2 & \cellcolor[HTML]{EDF6FF} 1 & \cellcolor[HTML]{EDF6FF} 5 &\cellcolor[HTML]{FFFFED} 1.000 \\
 
 Text  Sim. & cosine ($\uparrow$)     & 0.807 & 0.808 & 0.781 & 0.832 & 0.786 & 3  & 2  & 5 & 1 & 4 & \cellcolor[HTML]{FFFFED}0.200 \\
 
 Prompt Embedding Sim. & cosine ($\uparrow$)     & 0.820 & 0.820 & 0.801 & 0.808 & 0.774 & 1  & 1  & 4 & 3 & 5 & \cellcolor[HTML]{FFFFED}0.105 \\
 
 Image-Text Embedding & entropy ($\uparrow$)                & 10.819 $\pm$ 1.6e-4 & 9.210 $\pm$ 1.5e-4 & 8.922 $\pm$ 1.8e-4 & 10.309 $\pm$ 1.4e-4 & 10.837 $\pm$ 1.4e-4 & 2  & 5  & 4 & 3 & 1 & \cellcolor[HTML]{FFFFED}-0.200 \\
 
 Image-Text Embedding & max logits ($\uparrow$)             & 0.273 $\pm$ 0.032 & 0.264 $\pm$ 0.035 & 0.252 $\pm$ 0.027 & 0.260 $\pm$ 0.028 & 0.271 $\pm$ 0.029 & 1  & 3  & 5 & 4 & 2 & \cellcolor[HTML]{FFFFED}-0.400 \\
 
 Image Embedding Dist.  & min $L_{2}$ ($\downarrow$) & 0.768 $\pm$ 0.084 & 0.790 $\pm$ 0.090 & 0.850 $\pm$ 0.068 & 0.796 $\pm$ 0.079 & 0.753 $\pm$ 0.093& 2  & 3  & 5 & 4 & 1 & \cellcolor[HTML]{FFFFED}-0.600\\
 
 Image Embedding Dist. & mean $L_{2}$ ($\downarrow$)& 1.422 $\pm$ 0.013 & 1.414 $\pm$0.014 & 1.412 $\pm$ 0.013 & 0.413 $\pm$ 0.014 & 1.411 $\pm$ 0.012 & 5  & 4  & 2 & 3 & 1 & \cellcolor[HTML]{FFFFED}-0.200 \\
 Image Embedding Dist.  & max $L_{2}$ ($\downarrow$)& 1.099 $\pm$ 0.036 & 1.103 $\pm$ 0.041 & 1.131 $\pm$ 0.033 & 1.089 $\pm$ 0.038 & 1.077 $\pm$ 0.033 & 3  & 4  & 5 & 2 & 1 & \cellcolor[HTML]{FFFFED}-0.200 \\
\bottomrule
\end{tabular}}

\end{table*}